\documentclass[11pt]{article}

\usepackage[final]{acl}
\usepackage{enumitem}

\usepackage{times}
\usepackage{latexsym}
\usepackage{amsmath} 
\usepackage[T1]{fontenc}
\usepackage{amsfonts}

\usepackage[utf8]{inputenc}
\usepackage{microtype}
\usepackage{multirow}
\usepackage{inconsolata}
\usepackage{graphicx}
\usepackage{tcolorbox}
\usepackage{booktabs} 
\usepackage{hyperref}
\usepackage{subcaption}
\usepackage{array}
\usepackage{tabularx}
\usepackage{float}
\newcolumntype{L}[1]{>{\raggedleft\let\newline\\\arraybackslash\hspace{0pt}}m{#1}}
\tcbuselibrary{listings,breakable}
\tcbuselibrary{skins}

\title{PAKTON: A Multi-Agent Framework for Question Answering in Long Legal Agreements}

\author{
    Petros Raptopoulos$^{1}$, Giorgos Filandrianos$^{1,2}$, Maria Lymperaiou$^{1}$, Giorgos Stamou$^{1}$ \\
    $^{1}$School of Electrical and Computer Engineering, AILS Laboratory, \\
    National Technical University of Athens, Greece \\
    $^{2}$Instituto de Telecomunicações, Portugal \\
    \texttt{\href{mailto:petrosrapto@gmail.com}{petrosrapto@gmail.com}}, \{\texttt{\href{mailto:geofila@islab.ntua.gr}{geofila}}, \href{mailto:marialymp@islab.ntua.gr}{marialymp}\}@islab.ntua.gr, \texttt{\href{mailto:gstam@cs.ntua.gr}{gstam@cs.ntua.gr}}
}

\begin{document}
\maketitle
\begin{abstract}
% Reviewing contract documents is a time-consuming process that often requires expert legal knowledge, making it inaccessible to the general public.
% We introduce PAKTON, a multi-agent solution aimed at overcoming the challenges of contract analysis through teamwork among agents and effective use of external knowledge.
% It consists of three agents: (1) the Archivist, who interacts with the user and manages knowledge storage (2) the Researcher, who retrieves relevant information using RAG techniques and (3) the Interrogator, who iteratively questions the Researcher to refine the answer and ensure confident responses.
% In the legal domain, answers are rarely binary, ambiguity is common, and interpretation frequently involves subjective judgment. \\
% PAKTON addresses this by grounding its responses in the contract and relevant external sources, while offering justifications for its conclusions. When the available information is insufficient to reach a confident decision,  the knowledge gap is explicitly acknowledged, promoting informed uncertainty rather than unsupported answers.
% Experiments across multiple contract analysis tasks show that our framework outperforms general-purpose models not only in accuracy, but also in explainability and reasoning, both of which are critical for decision-making in the presence of legal uncertainty.

Contract review is a complex and time-intensive task that typically demands specialized legal expertise, rendering it largely inaccessible to non-experts. Moreover, legal interpretation is rarely straightforward—ambiguity is pervasive, and judgments often hinge on subjective assessments. Compounding these challenges, contracts are usually confidential, restricting their use with proprietary models and necessitating reliance on open-source alternatives.
To address these challenges, we introduce PAKTON: a fully open-source, end-to-end, multi-agent framework with plug-and-play capabilities. PAKTON is designed to handle the complexities of contract analysis through collaborative agent workflows and a novel multi-stage retrieval-augmented generation (RAG) component, enabling automated legal document review that is more accessible, adaptable, and privacy-preserving.
Experiments demonstrate that PAKTON outperforms both general-purpose and pretrained models in predictive accuracy, retrieval performance, explainability, completeness, and grounded justifications as evaluated through a human study and validated with automated metrics.\footnote{Code can be found at \href{https://github.com/petrosrapto/PAKTON}{github.com/petrosrapto/PAKTON}.}

% Experiments across diverse contract analysis tasks demonstrate that our framework outperforms general-purpose models in accuracy, retrieval performance, explainability, completeness, and grounded justifications—all essential for decision-making under legal uncertainty.\footnote{The code will be available upon publication.}

% \footnote{Our resources and code are released at \url{https://github.com/petrosrapto/___} (to be made public).}
\end{abstract}

\section{Introduction}

Contracts are among the most foundational legal documents, governing a wide range of personal, commercial, and governmental relationships. Yet, analyzing contracts remains a complex and time-consuming task that typically requires legal expertise. As a result, contract review is often inaccessible to the general public and remains demanding even for professionals.
Research by World Commerce \& Contracting (WorldCC) reveals that organizations lose an average of 9.2\% of their annual revenue due to contract mismanagement, with that figure rising to 15\% for larger enterprises~\citep{worldcc2020value}. In parallel, the Institute for Supply Management (ISM) has reported that a typical Fortune 1000 company manages between 20{,}000 and 40{,}000 active contracts at any given time~\citep{ism2024homepage}, while even simple agreements can take over a week to approve. These challenges highlight the growing need for better support in understanding contracts and reducing the time required for manual review.

% \begin{figure}[h]  % Try to place the figure here
%   \centering
%   \includegraphics[width=0.50\textwidth,clip,trim=6cm 40cm 6cm 20cm]{static/pakton_overview_2.jpg}
%   \caption{User interaction flow within the PAKTON framework: submission of a legal query followed by the generation of a comprehensive report grounded in the provided contract and augmented by external knowledge sources.}
%   \label{fig:pakton_overview}
% \end{figure}

\begin{figure}[t!] 
\vskip -0.08in
  \centering
\includegraphics[width=0.45\textwidth,clip,trim=6cm 40cm 9cm 20cm]{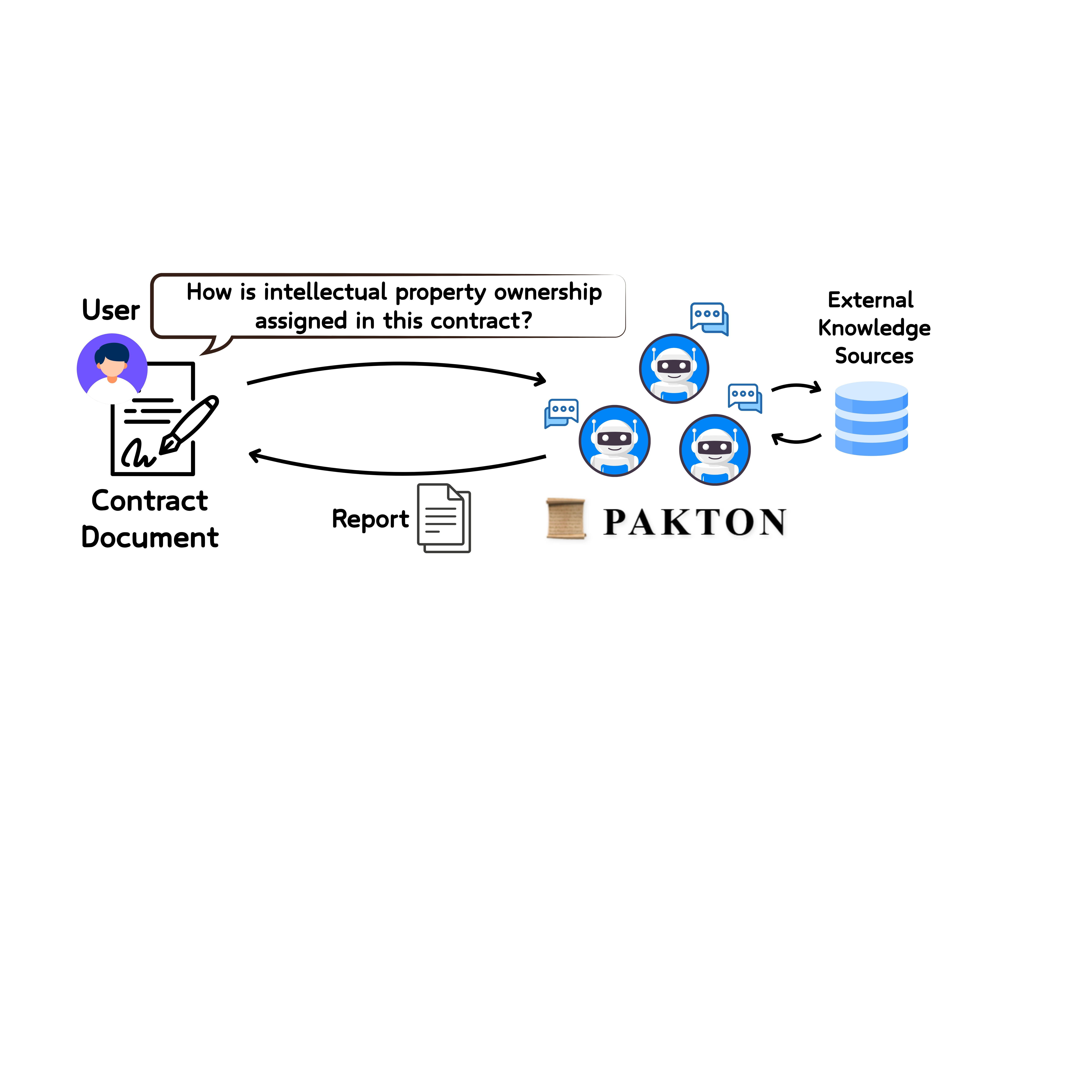}
  \caption{PAKTON user flow: legal query submission followed by comprehensive report generation.}
  \label{fig:pakton_overview}
\end{figure}

Recent advancements in large language models (LLMs) show promise in natural language understanding, question answering, and document summarization ~\citep{Achiam2023GPT4TR, Claude3S}. While LLMs are powerful, their application in specialized domains such as legal contract analysis presents unique challenges \cite{Frei2016AnAI, Eisenberg2022LegalR, kant2025towards}, including justifying decisions with clear reasoning and referencing both the contract and relevant external sources \cite{zhang2025syler}. Contract documents also exhibit several peculiarities that require specialized handling. They often contain complex legal terminology, calling for domain-specific language understanding. Overlapping or even contradictory clauses appear frequently \cite{marques2024contract, ichida2021detecting, aires2019concon}, requiring robust clause retrieval and conflict resolution \cite{aires2019concon, zhou2024intelligent}. Similarly, exceptions and references to different document parts also request robust retrieval mechanisms. Ambiguous phrasing and multiple interpretations are common, making careful contextual analysis essential.  Additionally, legal differences across jurisdictions necessitate consultation with external legal databases to maintain precision and relevance.

% A natural way to address these limitations is through Retrieval-Augmented Generation (RAG)~\citep{Gao2023RetrievalAugmentedGF, Fan2024ASO}. RAG offers a principled framework for augmenting LLMs with targeted retrieval of both internal and external documents, enabling models to ground their responses in verifiable evidence and domain-specific context. By explicitly incorporating retrieved passages during generation, RAG improves not only the factual accuracy but also the transparency and explainability of model outputs—qualities that are particularly crucial in legal contract analysis. Nevertheless, there is currently no comprehensive and open source RAG-based framework tailored for English-language contract analysis, with thorough evaluation across diverse contract-related tasks.

Retrieval-Augmented Generation (RAG)~\citep{Gao2023RetrievalAugmentedGF, Fan2024ASO, Wang2025ChainofRetrievalAG, gao2024retrievalaugmentedgenerationlargelanguage} directly addresses these limitations by integrating targeted retrieval of internal and external documents into the LLM generation process, grounding outputs in domain-specific, verifiable evidence. This approach enhances not only factual accuracy but also transparency and explainability, which constitute critical attributes for legal contract analysis. Moreover, RAG plays a pivotal role in the legal domain, where the inability to retrieve relevant spans can compromise reasoning and produce unsupported conclusions \cite{pipitone2024legalbenchragbenchmarkretrievalaugmentedgeneration}. Furthermore, given that legal contracts frequently contain sensitive or confidential information, proprietary models are often unsuitable, necessitating open-source alternatives capable of operating under limited computational resources.  Consequently, the development of efficient mechanisms for contract analysis becomes imperative.
% However, no comprehensive open-source RAG framework currently exists for English-language contract analysis with rigorous evaluation across diverse contract-related tasks.

Within this context, we introduce PAKTON\footnote{\textit{PAKTON} comes from the ancient Greek word that means agreement or contract, related to the Latin "pactum".} (Figure \ref{fig:pakton_overview}), a multi-agent framework designed to analyze contract documents and provide explainable, legally grounded answers to user queries along with a comprehensive report. Inspired by ~\citet{Shao2024AssistingIW}, PAKTON is composed of three specialized collaborative agents: (1) the \textit{Archivist}, which interacts with the user and manages structured document input; (2) the \textit{Researcher}, which retrieves relevant internal and external information using hybrid and graph-aware retrieval; (3) the \textit{Interrogator}, which engages in multi-step reasoning to iteratively refine the report. Each agent is dedicated to separate core legal aspects, ensuring optimal attribution of tasks. At the same time, the tri-agent structure facilitates implementation using different backbone LLMs, thus ultimately  offering a plug-and-play, highly customizable  solution.
% Unlike end-to-end black-box models, PAKTON emphasizes transparency, progressive refinement, and grounded justifications. It produces a structured legal report containing topic summaries, legal reasoning, key findings, and citations to both contract clauses and external sources. In cases where information is insufficient, the system explicitly acknowledges knowledge gaps, promoting informed decision-making rather than overconfident guesses.
% % PAKTON additionally, is designed with plug-and-play modularity, and all of its components are open source and lightweight, enabling on-premise deployment.
% We evaluate PAKTON across five contract analysis tasks, assessing its performance using ten criteria, including both automated metrics and human evaluations. Our experiments demonstrate that PAKTON significantly outperforms general-purpose LLMs in terms of retrieval accuracy and explainability. These results indicate that PAKTON moves us closer to achieving practical, interpretable AI systems capable of effectively supporting contract analysis.
PAKTON departs from black-box models by prioritizing transparency, progressive refinement, and grounded justifications. It generates structured legal reports with topic summaries, legal reasoning, key findings, and precise citations to contract clauses and external sources, while explicitly flagging knowledge gaps to avoid unsupported claims. All components are open source, lightweight, and support on-premise deployment. We benchmark PAKTON on five contract analysis tasks using ten criteria combining automated and human metrics, and find that it substantially surpasses general-purpose LLMs in accuracy and explainability. %These advances bring us closer to practical, interpretable AI for contract analysis.

\section{Related Work}

% Our approach draws inspiration from recent advances in multi-agent systems, retrieval-augmented generation, and legal domain applications of LLMss. We highlight key works that influenced the design of our system, while also identifying important distinctions in their scope and methodology.

% Recent progress in applying LLMs and RAG to legal tasks has led to notable improvements in contract analysis, legal question answering, and document review \cite{shu2024lawllm, lai2024large}. Domain-specific RAG frameworks, such as Legal Query RAG, have demonstrated significant gains in accuracy and relevance by integrating fine-tuned legal LLMs, custom evaluation agents, and recursive feedback mechanisms, effectively reducing hallucinations and improving the quality of responses to complex legal queries \cite{wahidur2025legal}. Literature reviews confirm a surge in research activity, with LLMs increasingly used for contract review, legal research, and regulatory compliance, and highlight the growing sophistication of methods and the expansion to multilingual and cross-jurisdictional applications \cite{siino2025exploring}.

Recent advances in applying LLMs and RAG to legal tasks significantly boost contract analysis, legal QA, and document review~\cite{shu2024lawllm, lai2024large}. Domain-specific RAG frameworks, such as Legal Query RAG, improve accuracy and relevance by combining fine-tuned legal LLMs, evaluation agents, and recursive feedback, reducing hallucinations and enhancing responses to complex queries~\cite{wahidur2025legal}. Literature surveys highlight a sharp rise in research on LLM-driven contract review, legal research, and regulatory compliance, alongside increasing methodological sophistication and expansion into multilingual, cross-jurisdictional contexts~\cite{siino2025exploring}.

Despite these advances, the community faces persistent challenges. A major issue is the tendency of LLMs to generate hallucinated or misleading responses, especially when lacking deep domain knowledge or when retrieval mechanisms fail to surface the most relevant legal context \cite{wahidur2025legal, zhao2024chat2data, saha2024advancing}. One strategy that has been investigated to mitigate this issue is fine-tuning language models on dedicated legal corpora, with the aim of deepening their understanding of legal terminology, reasoning styles, and contextual subtleties \cite{Colombo2024SaulLM54BS, Colombo2024SaulLM7BAP, huang2023lawyer}.  The evaluation of generated content remains difficult, as automated metrics often do not align with expert legal judgment, and human evaluation is costly and time-consuming \cite{wahidur2025legal, ryu2023retrieval, veturi2024rag}. Data availability and quality are also significant hurdles, particularly for specialized legal domains or languages with limited resources \cite{ryu2023retrieval, akarajaradwong2025nitibench}. Furthermore, the complexity of legal language and reasoning, as well as the need for transparency and explainability in AI-generated outputs, present ongoing obstacles \cite{wahidur2025legal, akarajaradwong2025nitibench, bianchini2024enhancing}.
Approaches, such as integrating case-based reasoning with RAG \cite{wiratungacbr}, leveraging knowledge graphs \cite{bianchini2024enhancing}, and developing new retrieval and evaluation strategies  \cite{akarajaradwong2025nitibench, saha2024advancing, ryu2023retrieval}, are being explored to address these limitations.

Additionally, multi-agent frameworks have recently gained attention, enhancing the reasoning capabilities and reliability of legal AI systems \cite{shengbinyue2025multi, sun2024lawluo, zhao2024longagent, liu2025hm}. Inspired by prior work such as STORM~\citep{Shao2024AssistingIW}, which introduces collaborative multi-turn interactions among diverse agents for content generation, and ChatLaw~\citep{Cui2023ChatlawAM}, which employs a knowledge graph-enhanced multi-agent design for legal assistance, we adopt a similar approach tailored specifically for contract analysis. PAKTON is designed to address the complex reasoning and rigorous transparency requirements inherent in legal tasks by orchestrating agents that iteratively interrogate, retrieve, and refine multi-source information.

\section{PAKTON Framework}

\begin{figure*}[t!]
\vskip -0.05in
    \centering
    % First image
    \includegraphics[width=0.68\textwidth]{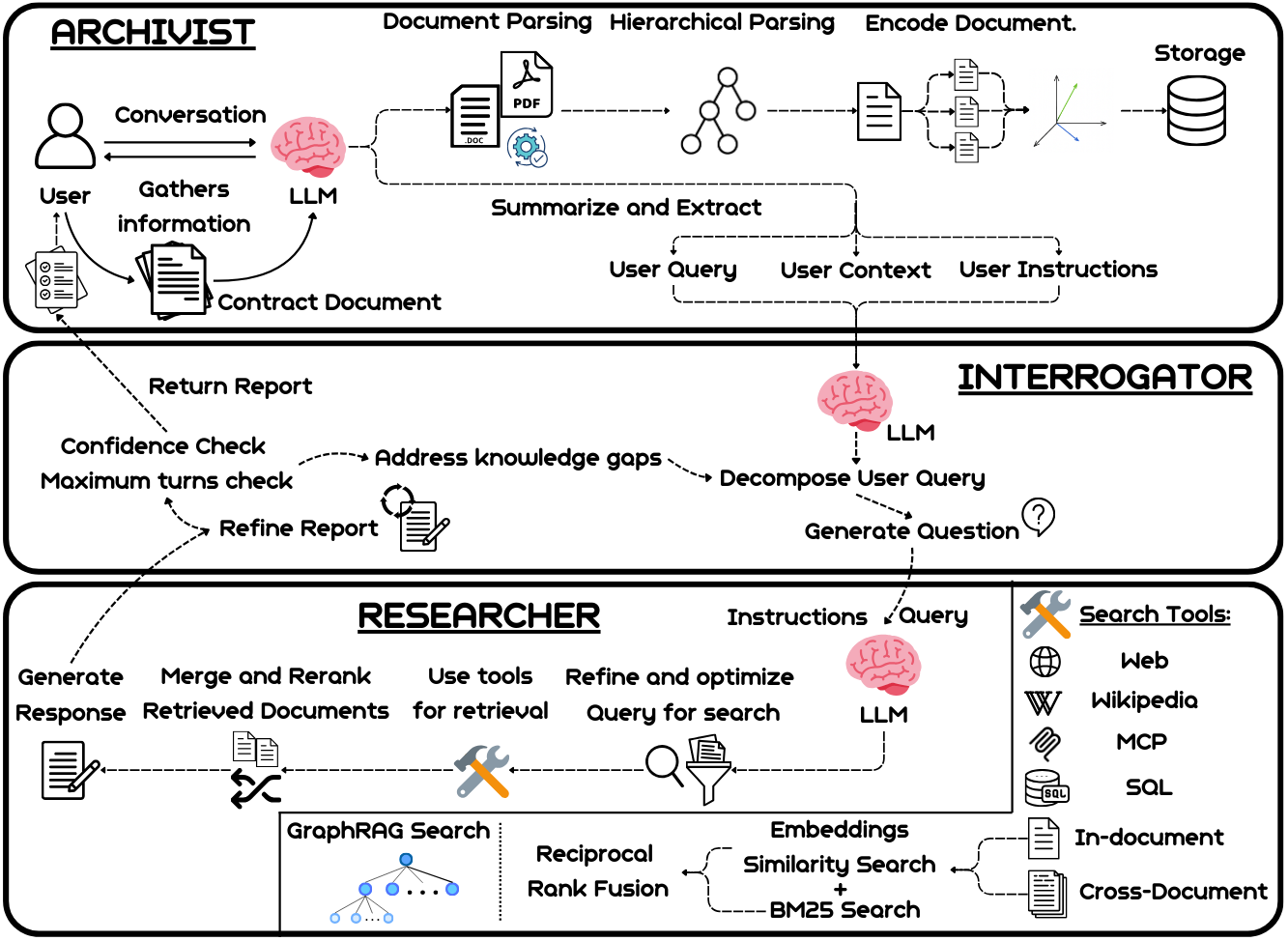}
\vskip -0.04in
    \caption{An overview of the proposed PAKTON framework and its internal components.}
    \label{fig:framework_overview}
\end{figure*}

% The main objective of the proposed framework is to analyze a contract document provided by the user and respond to user queries grounded in that document, while also incorporating relevant insights from external sources. A central focus of the system is to justify its responses by explicitly referencing spans from the contract and providing transparent reasoning behind the final answer. This final output is presented in the form of a structured legal report.

% To achieve this, the framework is built around three specialized agents that collaborate with one another: the Archivist, the Interrogator and the Researcher. Each agent plays a distinct role in the reasoning and information retrieval process, contributing to a more accurate, explainable, and well-supported response. An illustration of the framework is presented in Fig.\ref{fig:framework_overview}.

PAKTON aims to analyze user-provided contracts and generate query responses grounded in the contract’s content while integrating pertinent external knowledge. A key design goal is to ensure transparent and traceable reasoning by referencing evidence spans from the contract and articulating the rationale behind each conclusion. The final output is delivered as a structured legal report.

To operationalize this functionality, we employ a tri-agent, model-independent architecture comprising the \textit{Archivist}, \textit{Interrogator}, and \textit{Researcher}, each fulfilling a clear separation of concerns by assigning  distinct roles regarding the retrieval, reasoning, and synthesis pipeline. This agentic structure mirrors how legal professionals work in practice, aligning the system’s workflow with real-world roles for greater trust and interpretability, improving contract analysis organization.
Each agent is implemented as a reasoning loop following the ReAct paradigm~\citep{Yao2022ReActSR}, combining reasoning and action steps via a backbone LLM. This design enables agents to reason over their current state, generate tool-specific actions and iteratively refine their decisions based on intermediate outcomes. 
% This tri-agent modular collaboration enhances the accuracy, explainability, and evidentiary grounding of the system’s outputs. 
An overview of PAKTON is illustrated in Figure ~\ref{fig:framework_overview}. Further  details are provided in App. ~\ref{app:framework_implementation_details}.

\subsection{Archivist}
\label{subsection:archivist}

The \textit{Archivist} is responsible for gathering and organizing relevant user information. It plays a central role in the initial stages of the system by collecting and summarizing the user's query, any accompanying instructions, and contextual background. This information is then structured and passed to the \textit{Interrogator} agent. The \textit{Archivist} also manages storage and embedding of the user's contract document and actively engages in dialogue to resolve ambiguities or request missing user information, ultimately implementing three core functionalities:

% \subsubsection{Information Extraction}
% The Archivist interacts with the user to extract three attributes: the user query, any specific instructions or goals associated with that query, and the broader context relevant to the task. These components are identified, summarized and passed as input to the Interrogator. When necessary, the Archivist prompts the user for clarification or additional details, ensuring complete input.

\paragraph{Document Parsing.} 
To enable practical deployment, the \textit{Archivist} ingests diverse document formats within an integrated information pipeline. It natively parses text-based files, preserving structure such as headings, paragraphs, and numbered clauses. For PDFs, it applies OCR methods \citep{Smith2007AnOO, Smith2013HistoryOT} with visual layout analysis to recover document hierarchies, including sections, tables, and multi-column formats. All extracted content is then filtered and normalized to ensure consistent, structured input for downstream processing.
% To support real-world usage, the Archivist is equipped to process a variety of document formats as part of the broader information pipeline. 
% It handles attached text-based files directly, preserving structural elements such as headings, paragraphs, and numbered clauses.
% For PDF documents, the Archivist employs Optical Character Recognition (OCR) techniques ~\citep{Smith2007AnOO, Smith2013HistoryOT}, which leverage visual layout analysis to recover the document's structure accurately. This includes identifying sections, tables, and multi-column layouts when present. Following extraction, all documents undergo a filtering and normalization step to ensure clean, structured input for downstream processing.

\paragraph{Hierarchical Parsing.} 
Legal documents typically feature complex structures, cross-references, and nuanced semantic variations that render flat or linear representations inadequate for accurate analysis. To overcome this, we generate a hierarchical tree that mirrors the contract’s internal organization, introducing one core design novelty of PAKTON. The contract text is parsed into this tree structure by leveraging structural cues in conjunction with either semantic similarity measures based on BERT embeddings \cite{wang2020minilm} or from LLMs. 
% Long legal documents often exhibit intricate structure, interdependencies across distant sections, and subtle semantic shifts that make flat or linear representations insufficient for accurate analysis. To address this, we construct a hierarchical tree representation of the contract that reflects its internal organization.
% The parsed contract content is transformed into a tree that captures the hierarchical organization of the document. This is accomplished by combining structural cues extracted from OCR (in the case of PDFs) with either semantic similarity computations (using BERT-based embeddings \cite{wang2020minilm}) or LLM calls (using structured output).

The structural parsing process involves two main steps as shown in Figure~\ref{fig:graph_representation}: (1) identifying the distinct document sections, and (2) determining the hierarchical relationships between them. Examples of sections include titles, clauses, paragraphs, and enumerated list items.
A section is considered the child of another if, structurally, it appears nested within the parent section based on indentation, numbering patterns, visual layout (in OCR-extracted documents), or semantic proximity. 

\paragraph{Encode Document.} The tree representation of the contract is used to create contextualized chunks that are then embedded for retrieval. Context-aware chunking has been shown to reduce failed retrievals by preserving semantic coherence across sections~\citep{anthropic2024contextual}. In particular, both overlapping~\citep{Wang2024SearchingFB} and dynamic chunking~\citep{duarte-etal-2024-lumberchunker} have demonstrated gains in retrieval performance by maintaining relevant context around each chunk boundary.
To capture different contextual information levels, we generate three chunk types for each tree node, which are later concatenated into a unified chunk set. Duplicate or highly similar chunks are filtered out to improve embedding diversity and avoid redundancy. The chunking strategies are as follows:

\begin{itemize}[left=0.1em, itemsep=2pt, parsep=0pt, topsep=1pt]
    % \item \textbf{Node-level chunking:} One chunk is created per tree node, using only its own content. This provides fine-grained, localized embeddings and ensures that each atomic section is represented individually and can be matched precisely to user queries targeting specific clauses.
    
    % \item \textbf{Ancestor-aware chunking:} One chunk is created per node by concatenating its content with all of its ancestor nodes, in hierarchical order. This allows each section to be interpreted within the broader context of the contract structure, inheriting meaning and framing from parent sections such as titles or article headers.
     \item \textbf{Node-level:} Encodes each node in isolation to preserve fine-grained semantics. This supports precise alignment with user queries targeting specific clauses or legal provisions and ensures that each atomic unit is independently retrievable.
    
    \item \textbf{Ancestor-aware:} Concatenates a node with its hierarchical ancestors, capturing inherited context from section headers, articles, and titles. This contextualization aids disambiguation and enhances understanding of a section’s role within the broader contractual structure.
    
    \item \textbf{Descendant-aware:} Aggregates a node with its descendants to embed cohesive semantic units, such as clauses with their subclauses or enumerations. This is particularly beneficial for reasoning over compound or multi-part provisions, where meaning is distributed across nested content.

    % \item \textbf{Descendant-aware chunking:} One chunk is created per node by concatenating its content with all of its descendant nodes. This ensures that semantically related content (e.g., subclauses and list items) are embedded together, allowing the model to interpret sections in a more complete and context-rich manner.
\end{itemize}

Each chunk is further enriched with metadata, including its structural location within the tree, document position, filename, and a contract-level summary. This information facilitates both cross-document retrieval by guiding vector store routing, as well as intra-document relevance ranking.
By integrating multiple structural perspectives, our chunking framework improves the likelihood of retrieving relevant content across a spectrum of query granularities, from clause-specific lookups to section-level reasoning, advancing the encoding process in comparison to existing works.

% In addition to textual content, each chunk is enriched with metadata including its structural origin within the tree, its position within the document, and high-level document information such as the filename and a contract summary. This metadata, together with the content, supports cross-document retrieval by guiding the selection of the most appropriate vector store during indexing and enabling the identification of the most relevant chunks during retrieval.”

% The aforementioned multi-perspective chunking strategy increases the likelihood of retrieving relevant content across a variety of query granularities, from specific clause-level inquiries to broader contextual reasoning over full sections.

% \subsubsection{Safety and Guardrailing}
% The Archivist serves as the sole interface between the user and the system, bearing the critical responsibility of safeguarding the system against prompt injection attacks and other forms of malicious input. It acts as the first line of defense, ensuring robust guardrails are in place to maintain the integrity and security of the overall pipeline. INCOMPLETE, MUST SEE IT AGAIN

% \begin{figure*}[ht]
%     \centering

%     % First image
%     \includegraphics[width=0.8\textwidth]{static/pakton_cropped_v2.png}
        
%     \caption{PAKTON Framework Overview}
%     \label{fig:framework_overview}
% \end{figure*}

\subsection{Interrogator}

The \textit{Interrogator} is responsible for generating the final report presented to the user as the system’s response, orchestrating a multi-step reasoning process that addresses queries with  accuracy and confidence. It receives the user's query, contextual information, and instructions, and initiates an iterative interrogation of the \textit{Researcher}. This process involves decomposing the original query and generating a series of questions aimed at refining and deepening the system’s understanding.
Once a response is received from the \textit{Researcher} for the first question, the \textit{Interrogator} generates a preliminary report, attempting to directly answer the user’s query. This initial response also serves to identify potential knowledge gaps that require further research.
Based on this initial exchange, the \textit{Interrogator} formulates follow-up questions aimed at clarifying missing information in the preliminary answer. 
When generating each subsequent question, it considers the user query, context, instructions, a summary of previously asked questions, and the current draft of the report. Questions that are most likely to significantly improve the final response are prioritized.
With each new response from the \textit{Researcher}, the \textit{Interrogator} incrementally refines the draft report. The process continues until one of two stopping conditions is met: 1) the \textit{Interrogator} determines that the answer is sufficiently confident and complete, or 2) the maximum number of allowed interrogation turns (simulating retrieval depth), set by the user, has been reached.

The \textit{Interrogator} ensures the structural integrity and completeness of the  report, which must include:
1) Title and topic summary, 2) Legal reasoning and key findings, 3) Preliminary answer and suggested research directions, 4) Knowledge gaps and follow-up questions, 5) Cited sources and evidentiary support.
% \begin{enumerate}[left=0.1em, itemsep=2pt, parsep=0pt, topsep=1pt]
%     \item Title and topic summary
%     \item Legal reasoning and key findings
%     \item Preliminary answer and suggested research directions
%     \item Knowledge gaps and follow-up questions
%     \item Cited sources and evidentiary support
% \end{enumerate}
% Incorporating a title and topic summary at the outset of each refinement round is essential. Prompting strategies such as question rephrasing and repetition have been shown to improve in-context learning and output accuracy~\citep{mekala-etal-2024-echoprompt,xu-etal-2024-reading}. Incorporating a title and summary a
% nd by conducnting the concretely implements these strategies by priming the model with focused topical cues and constraining generative scope. This not only promotes alignment with user intent but also enhances interpretability by enabling early detection of semantic drift. Their inclusion thus serves both functional and communicative purposes—improving model performance while clarifying the system’s understanding for the user.
Incorporating query-specific information e.g. title and summary into the final report not only enhances user readability but also improves model performance by encouraging query rephrasing at each refinement step, thus deepening task comprehension~\citep{mekala-etal-2024-echoprompt}. The \textit{Interrogator}'s iterative architecture  further facilitates  knowledge gap identification, targeted information retrieval, and progressive query disambiguation. The inherent repetition in this process has been shown to enhance in-context learning and output accuracy~\citep{xu-etal-2024-reading}. These design choices are integral to the overall effectiveness of  PAKTON.

% We also explored an alternative approach in which the entire interrogation history was provided to the LLM to generate the final report in a single step. However, this method proved less effective in practice, as relevant information was often lost or diluted in the middle of the context~\citep{Liu2023LostIT}. Moreover, smaller models struggled to manage the extended input, leading to reduced performance and difficulty in accurately locating supporting evidence.

\subsection{Researcher}

The \textit{Researcher} is responsible for retrieving relevant information to support the \textit{Interrogator} in answering the user's query, equipped with multiple retrieval methods. Depending on the nature of the query and the accompanying instructions, the \textit{Researcher} autonomously selects the most suitable retrieval method, or combination of methods. This selection process is driven by prompting an LLM to choose the optimal set of tools based on the query content and tool descriptions.
Retrieval methods are categorized into the following:
\begin{itemize}[left=0.1em, itemsep=2pt, parsep=0pt, topsep=1pt]
    \item \textbf{In-document retrieval}: Leveraging the document chunks and embeddings provided by the \textit{Archivist}, the \textit{Researcher} retrieves spans to address the query. A hybrid retrieval approach is employed, combining BM25~\citep{Robertson2009ThePR}, dense embeddings~\citep{Lewis2020RetrievalAugmentedGF}, and Reciprocal Rank Fusion (RRF)~\citep{Cormack2009ReciprocalRF}. This is further enhanced by LightRAG~\citep{Guo2024LightRAGSA} (a lightweight version of GraphRAG ~\citep{Edge2024FromLT}), improving entity- and relation-level matching within and across documents, and is particularly effective for answering global queries that require reasoning over multiple interrelated documents.
    
    \item \textbf{Cross-document retrieval}: This method retrieves relevant spans from other documents, which the \textit{Interrogator} can leverage as exemplars in few-shot prompting. 

    \item \textbf{Retrieval of external knowledge}: \textsc{PAKTON} provides retrieval tools including web search, Wikipedia, SQL databases, and external legal sources via the Model Context Protocol (MCP)~\citep{anthropic2024modelcontext} to supplement model knowledge or provide real-time information. For fair evaluation, these retrieval capabilities were not used in our experiments.
    % \textsc{PAKTON} supports additional retrieval capabilities, including web search, Wikipedia access, SQL database querying, and integration with external legal databases via the Model Context Protocol (MCP)~\citep{anthropic2024modelcontext}. These tools are designed to supplement the model’s knowledge in cases where it lacks sufficient information or requires real-time updates. However, to ensure a fair and controlled evaluation, this category of retrieval was not employed in our experiments.
    
\end{itemize}

% The available retrieval tools include:
% \begin{itemize}[left=0.1em, itemsep=2pt, parsep=0pt, topsep=1pt]
%     \item \textbf{Web, Wikipedia, SQL-DB search}
%     \item \textbf{In-document and cross-document search}, leveraging a hybrid retrieval approach that combines BM25 ~\citep{Robertson2009ThePR} with dense embeddings ~\citep{Lewis2020RetrievalAugmentedGF} and Reciprocal Rank Fusion ~\citep{Cormack2009ReciprocalRF}. This is further enhanced with GraphRAG ~\citep{Edge2024FromLT} or LightRAG ~\citep{Guo2024LightRAGSA} techniques, which help accurately identify information related to entities and their relationships within and across documents.
    
%     \item \textbf{External legal database integration}, allowing access to third-party legal information systems or APIs using Model Context Protocol (MCP) ~\citep{anthropic2024modelcontext}.
% \end{itemize}
To ensure both high recall and precision, the \textit{Researcher} employs a two-step retrieval and reranking process. Initially, high-recall retrievers, such as the aforementioned ones, are used to collect a broad set of potentially relevant passages maximizing coverage. Subsequently, a cross-encoder model reranks the retrieved results by jointly encoding the query and each passage, allowing for more accurate semantic relevance estimation~\citep{Karpukhin2020DensePR}. This reranking stage serves as a high-precision filter that refines the initial candidate set. Relevance scores are normalized using a sigmoid function, and passages exceeding a predefined similarity threshold are passed to the response generation module, which synthesizes the final answer.

\section{Experiments and Results}

% inside the table
\setlength\tabcolsep{1.8pt}   % a bit tighter
\renewcommand{\arraystretch}{0.95}
\begin{table}[h!]
\vskip -0.02in
  \small
  \centering
  \begin{tabular}{L{1.1cm}|p{1.4cm}|cc|ccc}
    \toprule
    \textbf{Model} & \textbf{Method} & \textbf{Acc.} & \textbf{F1 [W]} &
    \textbf{F1 [E]} & \textbf{F1 [C]} & \textbf{F1 [N]}\\ \midrule
    
    Saul7B & Inst. Tun. & 0.4196 & 0.2900 & 0.0589 & 0.0680 & 0.5920 \\
    Saul54B & Inst. Tun.  & 0.7020 & 0.6792 & 0.7727 & 0.1729 & 0.7024 \\
    \midrule
    \multirow{5}{1cm}{Mistral 7B} & ZS & 0.5364
       & 0.5042 & 0.5279 & 0.0248 & 0.5951 \\
                              & FS & 0.5065 & 0.4702 & 0.6053 & 0.0082 & 0.4379 \\
                              & FS+Spans & 0.4940 & 0.4576 & 0.6085 & 0.0076 & 0.4053 \\
                              & PAKTON   & \textbf{0.7032} & \textbf{0.6789} & 0.7782 & 0.2469 & 0.6828 \\

    \midrule
    \multirow{5}{1cm}{Mixtral 8x7B} & ZS & 0.5423
       & 0.5475 & 0.6445 & 0.4103 & 0.4770 \\
                              & FS & 0.6002 & 0.5804 & 0.6836 & 0.1931 & 0.5642 \\
                              & FS+Spans & 0.6150 & 0.6017 & 0.6901 & 0.1951 & 0.6060 \\
                              & PAKTON   & \textbf{0.7423} & \textbf{0.7429} & 0.7864 & 0.6655 & 0.7187 \\

    \midrule
    \multirow{5}{1cm}{Qwen 2.5 72B} & ZS &
      0.7728 & 0.7699 & 0.8248 & 0.5776 & 0.7579 \\
                                    & FS & 0.7351 & 0.7241 & 0.8094 & 0.4920 & 0.6892 \\
                                    & FS+Spans & 0.7484 & 0.7432 & 0.8196 & 0.4378 & 0.7357 \\
                                    & PAKTON   & \textbf{0.8192} & \textbf{0.8188} & 0.8353 & 0.7737 & 0.8132 \\
    \midrule
    \multirow{5}{1cm}{Gemma 3 27B} & ZS &
      0.7886 & 0.7860 & 0.8316 & 0.6348 & 0.7739 \\
                                   & FS & 0.7191 & 0.7049 & 0.7815 & 0.4608 & 0.6891 \\
                                   & FS+Spans & 0.7720 & 0.7639 & 0.8287 & 0.4728 & 0.7662 \\
                                   & PAKTON   & \textbf{\underline{0.8287}} & \textbf{\underline{0.8283}} & 0.8487 & 0.7546 & 0.8255 \\
    \midrule
    \multirow{5}{1cm}{Llama 3.3 70B} & ZS &
      0.6767 & 0.6716 & 0.7366 & 0.5378 & 0.6346 \\
                                     & FS & 0.6657 & 0.6565 & 0.7326 & 0.4431 & 0.6268 \\
                                     & FS+Spans & 0.6915 & 0.6879 & 0.7382 & 0.4244 & 0.6982 \\
                                     & PAKTON   & \textbf{0.8217} & \textbf{0.8207} & 0.8422 & 0.7488 & 0.8165 \\
    \midrule
    
    \multirow{5}{1cm}{Llama 3.1 70B} & ZS &
      0.5811 & 0.5577 & 0.5216 & 0.3152 & 0.6555 \\
                                     & FS & 0.5729 & 0.5506 & 0.5421 & 0.2381 & 0.6358 \\
                                     & FS+Spans & 0.5538 & 0.5180 & 0.4471 & 0.3014 & 0.6468 \\
                                     & PAKTON   & \textbf{0.7916} & \textbf{0.7903} & 0.8097 & 0.6846 & 0.7960 \\
    \midrule
    \multirow{5}{1cm}{Claude 3.5} & ZS &
      0.7916 & 0.7977 & 0.8757 & 0.5722 & 0.7691 \\
                                  & FS & 0.7778 & 0.7816 & 0.8588 & 0.5702 & 0.7505 \\
                                  & FS+Spans & 0.7999 & 0.8034 & 0.8678 & 0.6046 & 0.7826 \\
                                  & PAKTON   & \textbf{0.7990} & \textbf{0.8000} & 0.8157 & 0.7046 & 0.8072 \\
    \midrule
    \multirow{5}{1cm}{Claude 3.7} & ZS &
      0.7704 & 0.7781 & 0.8633 & 0.5602 & 0.7398 \\
                                  & FS & 0.7590 & 0.7602 & 0.8463 & 0.5607 & 0.7165 \\
                                  & FS+Spans & 0.7724 & 0.7766 & 0.8538 & 0.5805 & 0.7417 \\
                                  & PAKTON   & \textbf{0.8247} & \textbf{0.8254} & 0.8386 & 0.7495 & 0.8304 \\
    \midrule
    \multirow{5}{1cm}{Deep\-seek V3} & ZS &
      0.7886 & 0.7875 & 0.8487 & 0.6117 & 0.7648 \\
                                     & FS & 0.7681 & 0.7607 & 0.8346 & 0.6104 & 0.7182 \\
                                     & FS+Spans & 0.7743 & 0.7714 & 0.8377 & 0.5812 & 0.7465 \\
                                     & PAKTON   & \textbf{0.8192} & \textbf{0.8200} & 0.8315 & 0.7615 & 0.8224 \\
    \midrule
    \multirow{5}{1cm}{GPT‑4o} & ZS &
      0.6121 & 0.6366 & 0.7490 & 0.4162 & 0.5698 \\
                              & FS & 0.6640 & 0.6789 & 0.7372 & 0.4734 & 0.6666 \\
                              & FS+Spans & 0.6482 & 0.6574 & 0.6664 & 0.4636 & 0.6950 \\
                              & PAKTON   & \textbf{0.7966} & \textbf{0.7972} & 0.7964 & 0.7592 & 0.8068 \\
    \bottomrule
  \end{tabular}
  \vskip -0.02in
  \caption{Comparison of PAKTON versus other methods across models on the ContractNLI test set. The best accuracy/F1[W] per prompting method are shown in \textbf{bold}, and the best overall results are \underline{underlined}.}
  \label{tab:pakton_contractNLI_results_extended}
  \vskip -0.07in
\end{table}

% The proposed framework was evaluated using both quantitative and qualitative methods. For the quantitative evaluation, its accuracy was assessed on a well-defined and challenging legal reasoning task, where PAKTON demonstrated superior performance compared to all other methods. Additionally, to isolate and measure the contributions of individual components, we evaluated the Archivist-Researcher modules—responsible for retrieval-augmented generation (RAG)—across four benchmark datasets, achieving state-of-the-art results. Finally, given that PAKTON is a general-purpose system designed to produce outputs that are interpretable and actionable by human users, the quality of its generated responses was assessed against nine criteria. This evaluation included both a structured human assessment conducted by a panel of five expert attorneys and a Supreme Court Justice, and an automated assessment using the G-EVAL framework \cite{liu-etal-2023-g}. In both settings, PAKTON consistently outperformed GPT-4o, surpassing it in eight out of nine evaluation criteria.

Due to its model-independent design, PAKTON operates by accommodating different LLMs as agent instantiations. Specifically, our experimentation comprises Mistral \cite{mistral}, Qwen \cite{quen}, Gemma \cite{gemma}, Llama \cite{llama3}, Claude \cite{Claude3S}, Deepseek \cite{deepseek} and GPT-4 \cite{Achiam2023GPT4TR} model families.

% PAKTON is evaluated through both quantitative and qualitative methods. In the quantitative analysis, PAKTON is tested on contract analysis tasks,  outperforming all baselines in terms of accuracy. To isolate component-level contributions, since typical ablation analysis in agent-level is not applicable due to their interleaved nature, we conduct targeted evaluations to assess per-agent impact. Regarding
% the \textit{Archivist}-\textit{Researcher} modules, which are responsible for RAG, their evaluation is based on four benchmark datasets, achieving state-of-the-art results. For the \textit{Interrogator}, which is responsible for generating the answer to the user, a qualitative evaluation is conducted, with outputs assessed against nine criteria aligned with PAKTON’s goal of producing interpretable and actionable responses. This includes a structured human evaluation by five expert attorneys and a Supreme Court Justice, alongside an automated assessment using G-EVAL~\cite{liu-etal-2023-g}. In both evaluations, PAKTON consistently outperforms GPT-4o, exceeding it in eight of the nine criteria.

PAKTON is evaluated using both quantitative and qualitative methods. Quantitatively, it is tested on contract analysis tasks, consistently outperforming all baselines in accuracy. Traditional ablation studies at the agent level are not feasible, as each agent plays an essential and interdependent role. Instead, we conduct targeted evaluations to isolate contributions. The retrieval capabilities of the \textit{Archivist}–\textit{Researcher} pair are measured across four benchmark datasets, where PAKTON achieves state-of-the-art results. The overall generation performance of the end-to-end system, including the \textit{Interrogator}, is assessed quantitatively on a contract analysis task and qualitatively. The qualitative evaluation uses nine criteria aligned with PAKTON’s goal of producing interpretable, actionable responses and includes a structured human study with five expert attorneys and a Supreme Court Justice, as well as automated evaluation with G-EVAL~\cite{liu-etal-2023-g}. In both analyses, PAKTON consistently surpasses GPT-4o, outperforming it in eight of nine criteria.

\vskip -0.03in
\subsection{Quantitative Results}
\subsubsection{Performance on a classification dataset}

\paragraph{Setup.} For the quantitative evaluation, PAKTON is evaluated on the ContractNLI dataset \cite{koreeda-manning-2021-contractnli-dataset}. In this dataset, a premise denotes an entire contract document, with each premise paired with a corresponding hypothesis. The classification task involves determining whether the hypothesis is entailed by, contradicted by, or not addressed in (neutral with respect to) the associated contract. Given the substantial length of the contracts, the dataset also provides annotated spans that indicate the specific portions of text necessary for making the classification decision.

We evaluate PAKTON on  ContractNLI by comparing its performance against several baselines, including models specifically pretrained on legal corpora (Saul \cite{Colombo2024SaulLM54BS}) and LLMs employing different prompting techniques (Table \ref{tab:pakton_contractNLI_results_extended}). The evaluation metrics presented include overall accuracy and the weighted F1-score (F1[W]), alongside the individual F1-scores for the entailment E, contradiction C, and neutral N classes.
The prompting approaches examined include zero-shot (ZS), few-shot (FS)—where entire contractual documents are used as exemplars—and an alternative few-shot setting (FS-Spans), in which only the relevant spans influencing classification are provided. More details are presented in Appendix ~\ref{app:contractNLIexperiments}. 

\paragraph{Results} from Table \ref{tab:pakton_contractNLI_results_extended} indicate a clear superiority of PAKTON across all evaluated methods, notably outperforming even domain-specific models. Firstly, PAKTON consistently outperforms the Saul baselines regardless the backbone embedded LLM, with marginal benefits emerging even from small models, such as Mistral 7B, even though Saul is explicitly fine-tuned on legal data.
%Comparing the results of Mistral and Mixtral with Saul, we observe that PAKTON yields better performance even compared to fine-tuning on related legal corpora. 
%Notably, Mistral 7B with PAKTON surpasses Saul 54B—a model derived from the larger Mixtral 54B and fine-tuned on legal data. 
Furthermore, PAKTON is highly competitive against the reported prompting baselines; particularly, even the small
Mistral 7B with PAKTON outperforms standalone GPT-4o FS prompting, justifying experimentally the rationale behind the selected agentic structure and the agents' curated functionalities.

Furthermore, an interesting observation emerges from comparing the performance across different core LLMs utilizing our framework. The variation in performance among models integrated with PAKTON is minimal, highlighting the robustness of the proposed framework. For example, the F1[W] for Llama 3.1 70B score with PAKTON is 79.03\%, while for Gemma 3 27B is 82.83\%, yielding a modest difference of only 3.8 percentage points. In contrast, the performance gap between these models under the ZS prompting scenario is significantly larger (22.83\%), clearly demonstrating that Gemma 3 27B substantially outperforms Llama 3.1 70B in the baseline ZS setting. This considerable reduction in performance disparity indicates that PAKTON's architecture effectively mitigates the inherent variability among underlying LLMs, ensuring consistently high performance regardless of the core model employed.

\begin{figure}[t!]
    \centering
    \includegraphics[width=0.42\textwidth]{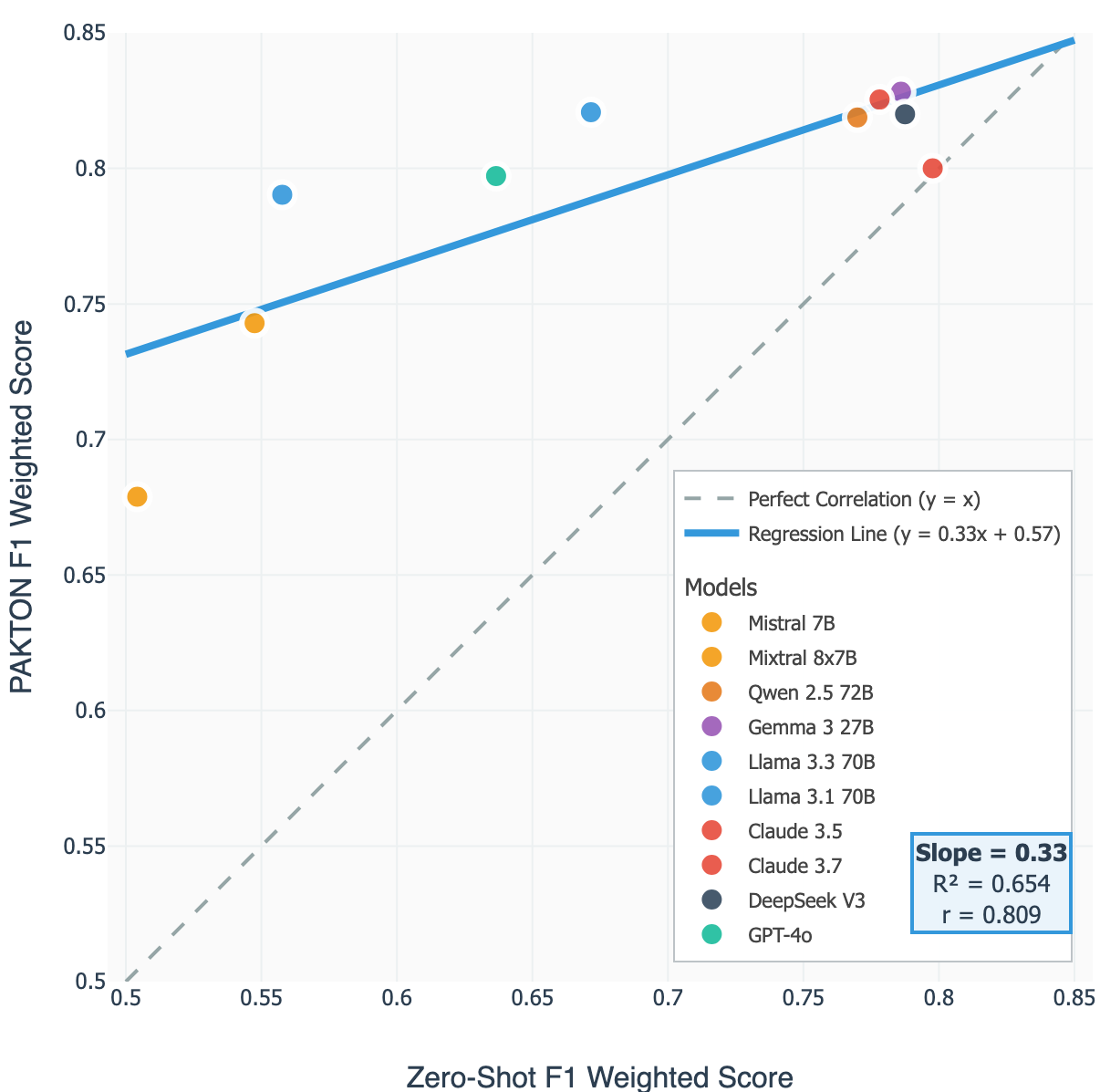}
    \caption{PAKTON F1[W] plotted against ZS scores, with the dashed line denoting perfect correlation and the solid line the regression fit.
}
    \label{fig:zs_pakton_orthogonal_regression}
\end{figure}

To further investigate PAKTON's robustness, we conduct a targeted statistical analysis focusing on variability and dependency. First, we compute the coefficient of variation (CV) across all models' PAKTON--based F1[W] scores, obtaining a low CV of $5.7$ (for comparison, ZS scores typically exhibit CVs $>16\%$). This indicates minimal relative variability—reduced to less than one third under PAKTON compared to ZS—and thus enhanced robustness, yielding consistently stable results regardless of the underlying model’s performance. Second, we perform an one--way ANOVA by splitting models into two groups according to the median of their ZS scores. The test reveals no statistically significant difference in PAKTON performance between the high-- and low--ZS groups ($F-statistic = 4.19$, $p=0.075$). 
% In the ANOVA notation, the first degree of freedom ($df=1$) captures variance due to the grouping factor (high vs.~low ZS), while the second ($df=8$) quantifies residual variance within each group; the non--significant $p$--value confirms that initial ZS ability does not materially influence \textsc{pakton} results.
Lastly, linear regression analysis assesses the direct relationship between baseline ZS performance and PAKTON, yielding a shallow slope of $0.33$, demonstrating that every unit of baseline gain translates into only about $1/3^{rd}$ under PAKTON, thereby compressing absolute performance gaps. Figure~\ref{fig:zs_pakton_orthogonal_regression} plots ZS F1[W] scores (x-axis) versus PAKTON F1[W] scores (y-axis), with the dashed line denoting perfect correlation ($y=x$). The figure shows that, regardless of their initial baseline performance, models converge to a similar outcome under PAKTON, indicating that any underlying LLM can be employed while achieving consistently stable results.

% This coefficient indicates only a minor dependency, further supporting the conclusion that the effectiveness of our framework is largely independent of the initial strength of the underlying models.

Collectively, these findings demonstrate that PAKTON substantially reduces performance disparities among diverse LLMs, by redistributing the reliance on model quality across other subcomponents of the system. Such robustness is particularly advantageous in the legal domain, where relying on open--source models alleviates the privacy risks associated with sending sensitive contractual or legal information to proprietary systems.

\subsubsection{Performance of RAG}
% \paragraph{Setup.} Beyond ContractNLI, we employ the LegalBench-RAG benchmark~\cite{pipitone2024legalbenchragbenchmarkretrievalaugmentedgeneration} to evaluate the retrieval pipeline—namely, the Archivist and Researcher modules—in isolation from answer generation. This evaluation is crucial, as PAKTON’s effectiveness depends on the quality of retrieved context; without it, the model’s performance would significantly degrade. LegalBench-RAG spans four contract-related domains—NDAs, M\&A agreements, commercial contracts, and privacy policies from consumer-facing platforms—enabling a comprehensive analysis of retrieval accuracy across diverse legal texts.
\paragraph{Setup.} In addition to ContractNLI, we leverage the LegalBench-RAG benchmark~\cite{pipitone2024legalbenchragbenchmarkretrievalaugmentedgeneration} to assess the retrieval pipeline (specifically, the \textit{Archivist} and \textit{Researcher} modules) independently of answer generation. This isolation is essential, as PAKTON’s overall performance hinges on the relevance of retrieved context. LegalBench-RAG covers four contract-related domains (NDAs, M\&A agreements, commercial contracts, and consumer-facing privacy policies), enabling a robust evaluation of retrieval accuracy across heterogeneous legal corpora.
We benchmark PAKTON’s RAG module against four baselines used in \citet{pipitone2024legalbenchragbenchmarkretrievalaugmentedgeneration}, namely: Naive (fixed 500-character chunks with OpenAI embeddings \cite{openai2024embedding}), RCTS (structure-aware splitting \cite{langchain2024rts}), Naive+Cohere (chunking with Cohere reranking\footnote{rerank-english-v3.0}), and RCTS+Cohere (structural chunking with reranking). Full details are provided in App. ~\ref{app:legalbenchragexperiments}.

\begin{table*}[h!]
\centering
\small
\resizebox{0.93\textwidth}{!}{%
\begin{tabular}{llrrrrrrr rrrrrrr}
\toprule
\multicolumn{2}{c}{} & \multicolumn{7}{c}{\textbf{Precision @ $k$}} &
\multicolumn{7}{c}{\textbf{Recall @ $k$}}\\
\cmidrule(lr){3-9}\cmidrule(lr){10-16}
\textbf{Dataset} & \textbf{Method} & 1 & 2 & 4 & 8 & 16 & 32 & 64 &
1 & 2 & 4 & 8 & 16 & 32 & 64\\
\midrule
\multirow{4}{*}{PrivacyQA}
 & Naive              & 7.86 & 7.31 & 6.41 & 5.06 & 3.58 & 2.41 & 1.54 & 7.45 & 12.53 & 20.88 & 32.38 & 42.45 & 54.27 & 66.07\\
 & RCTS               & 14.38 & 13.55 & 12.34 & 9.03 & 6.06 & 4.17 & 2.81 & 8.85 & 15.21 & 27.92 & 42.37 & 55.12 & 71.19 & 84.19\\
 & Naive + Cohere     & 14.38 & 13.55 & 12.34 & 9.02 & 6.06 & 4.17 & 2.81 & 8.85 & 15.21 & 27.92 & 42.37 & 55.12 & 71.19 & 84.19\\
 & RCTS + Cohere      & 13.94 & 15.91 & 13.32 & 9.57 & 6.88 & 4.68 & 3.28 &  7.32 & 16.12 & 25.65 & 35.60 & 51.87 & 64.98 & 79.61\\ \cmidrule{2-16}
& PAKTON & \textbf{19.94} & \textbf{16.84} & \textbf{11.44} & \textbf{8.62} & \textbf{7.38} & \textbf{6.42} & \textbf{6.08} & \textbf{13.34} & \textbf{22.43} & \textbf{32.67} & \textbf{43.39} & \textbf{61.65} & \textbf{82.30} & \textbf{89.42} \\

\midrule
\multirow{4}{*}{ContractNLI}
 & Naive              & 16.45 & 14.80 & 12.53 & 9.73 & 6.70 & 4.65 & 3.04 & 11.32 &  19.10 & 29.79 & 45.59 & 56.75 & 69.88 & 86.57\\
 & RCTS               &  6.63 &  5.29 &  3.89 & 2.81 & 1.98 & 1.29 & 0.90 & 7.63 & 11.33 & 17.34 & 24.99 & 35.80 & 46.57 & 61.72\\
 & Naive + Cohere     &  6.63 &  5.28 &  3.89 & 2.81 & 1.98 & 1.29 & 0.90 & 7.63 & 11.34 & 17.34 & 24.99 & 35.80 & 46.57 & 61.72\\
 & RCTS + Cohere      &  5.08 &  5.59 &  5.04 & 3.67 & 2.52 & 1.75 & 1.17 &  4.91 &  9.33 & 16.09 & 25.83 & 35.04 & 46.90 & 62.97\\ \cmidrule{2-16}
 & PAKTON & \textbf{33.02} & \textbf{30.34} & \textbf{17.33} & \textbf{9.98} & \textbf{5.87} & \textbf{4.68} & \textbf{4.52} & \textbf{53.14} & \textbf{67.47} & \textbf{80.06} & \textbf{89.71} & \textbf{95.50} & \textbf{99.56} & \textbf{99.82} \\

\midrule
\multirow{4}{*}{MAUD}
 & Naive              & 3.36 & 2.65 & 2.18 & 1.89 & 1.48 & 1.06 & 0.75 & 2.54 & 3.12 & 4.53 & 8.75 & 13.16 & 18.36 & 25.62\\
 & RCTS               & 2.65 & 1.77 & 1.96 & 1.40 & 1.39 & 1.15 & 0.82 &  1.65 &  2.09 & 4.59 & 6.18 & 12.93 & 21.04 & 28.28\\
 & Naive + Cohere     & 2.64 & 1.77 & 1.96 & 1.40 & 1.38 & 1.15 & 0.82 &  1.65 &  2.09 & 5.59 & 6.18 & 12.93 & 21.04 & 28.28\\
 & RCTS + Cohere      & 1.94 & 2.63 & 2.05 & 1.77 & 1.79 & 1.55 & 1.12 &  0.52 &  2.48 &  4.39 &  7.24 & 14.03 & 22.60 & 31.46\\ \cmidrule{2-16}
 & PAKTON & \textbf{25.47} & \textbf{17.45} & \textbf{10.51} & \textbf{7.24} & \textbf{5.08} & \textbf{3.18} & \textbf{1.85} & \textbf{23.99} & \textbf{30.09} & \textbf{34.49} & \textbf{46.42} & \textbf{59.74} & \textbf{74.96} & \textbf{82.80} \\

\midrule
\multirow{4}{*}{CUAD}
 & Naive              & 9.27 & 8.05 & 5.98 & 4.33 & 2.77 & 1.77 & 1.09 & 12.60 & 19.47 & 27.92 & 40.70 & 51.02 & 64.38 & 75.71\\
 & RCTS               & 1.97 & 4.03 & 4.83 & 4.20 & 2.94 & 1.99 & 1.25 & 1.62 & 8.11 & 17.72 & 31.68 & 44.38 & 60.04 & 74.70\\
 & Naive + Cohere     & 1.97 & 4.03 & 4.83 & 4.20 & 2.94 & 1.99 & 1.25 & 1.62 & 8.11 & 17.72 & 31.68 & 44.38 & 60.04 & 74.70\\
 & RCTS + Cohere      & 3.53 & 4.18 & 6.18 & 5.06 & 3.93 & 2.74 & 1.66 &  3.17 &  7.33 & 18.26 & 28.67 & 42.50 & 55.66 & 70.19\\ \cmidrule{2-16}
 & PAKTON & \textbf{11.02} & \textbf{8.83} & \textbf{6.81} & \textbf{4.72} & \textbf{2.78} & \textbf{2.07} & \textbf{1.62} & \textbf{16.52} & \textbf{24.76} & \textbf{33.34} & \textbf{46.67} & \textbf{59.53} & \textbf{77.08} & \textbf{86.23} \\

\midrule
\multirow{4}{*}{ALL}
 & Naive              & 2.40 & 3.76 & 4.97 & 4.33 & 3.39 & 2.17 & 1.29 & 3.37 & 8.44 & 21.30 & 34.51 & 48.88 & 64.47 & 76.39\\
 & RCTS               & 6.41 & 6.16 & 5.76 & 4.36 & 3.09 & 2.15 & 1.45 & 4.94 & 9.19 & 16.90 & 26.30 & 37.06 & 49.71 & 62.22\\
 & Naive + Cohere     & 6.41 & 6.16 & 5.76 & 4.36 & 3.09 & 2.15 & 1.45 & 4.94 & 9.19 & 16.90 & 26.30 & 37.05 & 49.71 & 62.22\\
 & RCTS + Cohere      & 6.13 & 7.08 & 6.65 & 5.02 & 3.78 & 2.68 & 1.81 &  3.98 &  8.82 & 16.10 & 24.34 & 35.86 & 47.54 & 61.06\\ \cmidrule{2-16}
 & PAKTON & \textbf{22.34} & \textbf{18.37} & \textbf{11.52} & \textbf{7.63} & \textbf{5.26} & \textbf{4.08} & \textbf{3.52} & \textbf{26.77} & \textbf{36.32} & \textbf{45.26} & \textbf{56.66} & \textbf{69.17} & \textbf{83.50} & \textbf{89.58} \\
\midrule
\end{tabular}}
\caption{Comparative retrieval performance on LegalBench-RAG, evaluated using Precision@k and Recall@k.}
\label{tab:legalbench_rag}
\end{table*}
\paragraph{Results.} Table~\ref{tab:legalbench_rag} presents a comparative analysis of retrieval performance across diverse legal corpora, using precision and recall at various $k$ thresholds to benchmark \textsc{PAKTON} against established baselines. Across all datasets, \textsc{PAKTON} consistently and significantly outperforms alternatives at every evaluated $k$. For instance, on ContractNLI, \textsc{PAKTON} achieves a Recall@1 of 53.14\%, nearly 5 times higher than the strongest baseline (11.32\%), a trend that holds across all $k$ values. Similar improvements are observed in other datasets: on PrivacyQA \cite{privacyqa}, MAUD \cite{maud}, and CUAD \cite{cuad}, \textsc{PAKTON} surpasses the best Recall@1 scores by margins often exceeding 20\%. Aggregate results reinforce this pattern, with \textsc{PAKTON} achieving more than five-fold increase in Recall@1 (26.77\% vs. 4.94\%). These improvements are especially critical in the legal domain, where high recall is essential. Failing to retrieve relevants spans can result in flawed reasoning or unsupported conclusions, particularly when legal documents contain conflicting clauses, exceptions, or interdependent provisions that must be interpreted in context.

% These consistently superior metrics—particularly pronounced at lower $k$—highlight PAKTON’s capacity to retrieve contextually salient information far more effectively than both naive segmentation and advanced reranking or hierarchical splitting methods, thereby significantly enhancing the quality of context available for downstream legal reasoning and retrieval-augmented generation.

\begin{figure*}[ht]
\vskip -0.09in
    \centering
    \includegraphics[width=0.9\textwidth, trim=300 990 300 300, clip]{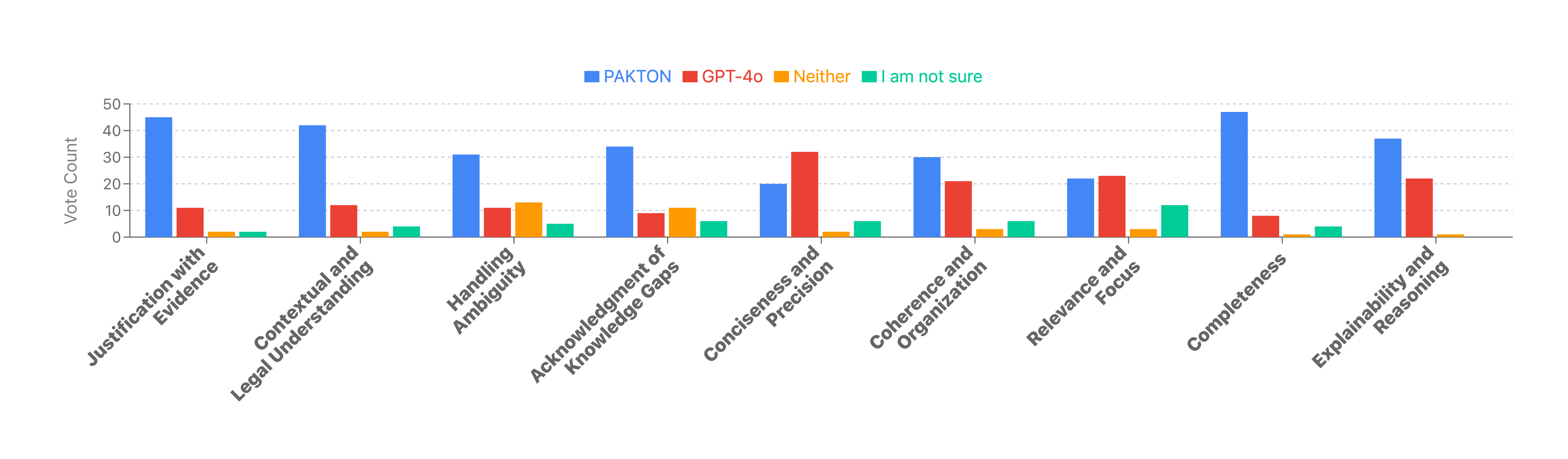}
    % \phantomcaption
    % \label{fig:geval_human_barplot}

    \vspace{0.5em}

    \includegraphics[width=0.89\textwidth, trim=250 420 300 450, clip]{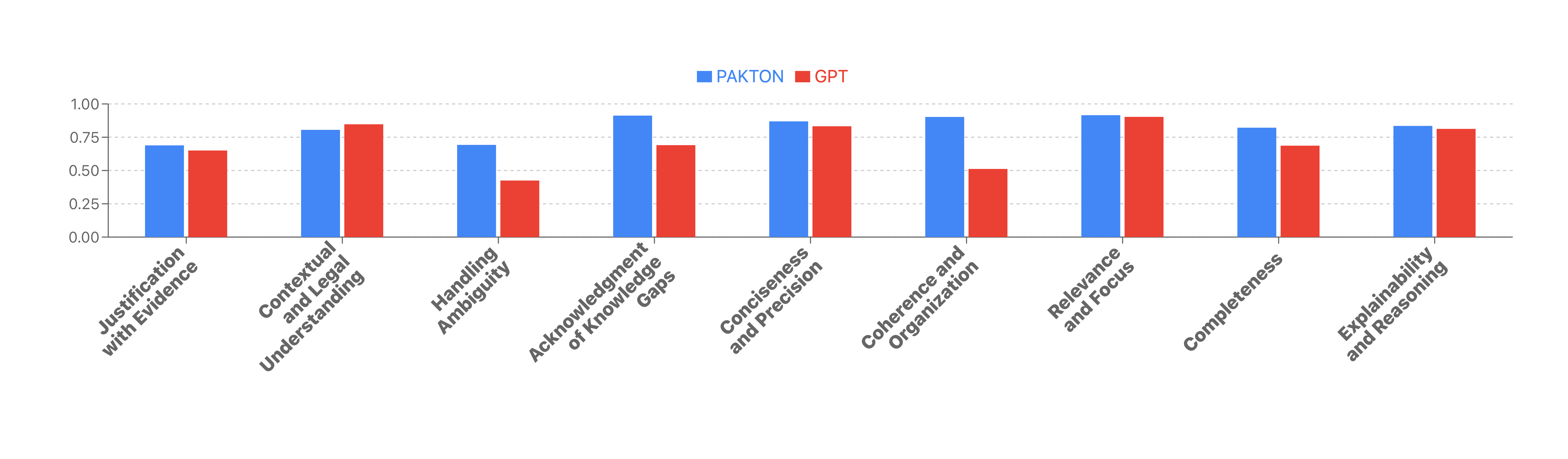}
    % \phantomcaption
    % \label{fig:geval_evaluation}
    \caption{\small Comparison analysis of PAKTON and GPT-4o. Top plot presents human preferences across nine evaluation criteria aggregated for all questions. Bottom plot shows G-EVAL scores for the same criteria, aggregated across all ContractNLI outputs.}
    \label{fig:combined_comparison}
\end{figure*}

\subsection{Qualitative Results}
\label{subsec:human_evaluation}

\paragraph{Setup.} To evaluate PAKTON’s practical utility, we conduct a human study using 15 benchmark legal questions  curated by five attorneys and a Supreme Court Justice to capture authentic legal reasoning across diverse scenarios. Each question is  chosen to test a distinct cognitive process and retrieval pattern relevant in legal analysis, ensuring breadth of evaluation.
In a \textit{Prolific}\footnote{\url{https://www.prolific.com}} survey, participants compare paired responses from PAKTON and GPT-4o\footnote{gpt-4o-2024-11-20} across 9 criteria: explainability, evidence, legal/contextual understanding, ambiguity handling, gap acknowledgment, conciseness, coherence, relevance, and completeness. 
Given PAKTON’s emphasis on explainability, evidence tracking, and accessibility, we deliberately evaluate its performance from the perspective of an average, non-expert user, which is its primary target audience.
For each criterion, they select the best answer and provided justification, with ``None'' and ``Not Sure'' options to ensure robustness. GPT-4o serves as a baseline given its adoption and, for fairness, leverage RAG when external documents are provided~\cite{openai_file_uploads}. Our human study considers a larger amount of evaluated samples over a wider range of criteria in comparison to similar endeavors \cite{jiang-etal-2024-unknown, Shao2024AssistingIW}.
Additional details are in Appendix ~\ref{app:human}.

% To assess PAKTON’s real-world applicability, we conducted a human evaluation centered on a representative legal scenario. A panel of five expert attorneys and a Supreme Court Justice developed 15 benchmark questions reflecting authentic legal reasoning challenges. Using these, we deployed a survey on \textit{Prolific}\footnote{\url{https://www.prolific.com}}, presenting participants with paired responses from PAKTON and ChatGPT\footnote{gpt-4o-2024-11-20} for each query and instructing them to evaluate outputs across nine predefined criteria (see Table~\ref{tab:evaluation_criteria}): explainability and reasoning, justification with evidence, contextual and legal understanding, handling ambiguity, acknowledgment of knowledge gaps, conciseness and precision, coherence and organization, relevance and focus, and completeness. For each criterion, participants selected the superior response and provided a brief justification; ``None'' and ``Not Sure'' options were available to ensure reliability and reduce response noise. 

% Notably, ChatGPT was chosen as a baseline due to its widespread adoption, and our comparison remains equitable, as the evaluated version of ChatGPT employs a RAG mechanism when external documents are supplied~\cite{openai_file_uploads}. Further methodological details are provided in Appendix~\ref{app:human}.

\paragraph{Results.} Sixty participants completed the evaluation,  comparing PAKTON and GPT-4o outputs across 9 criteria for 15 legal questions, yielding 540 individual judgments. As illustrated in Figure~\ref{fig:combined_comparison}, PAKTON is favored over GPT-4o on the majority of evaluation dimensions, a trend consistent across all legal scenarios assessed. 
%Aggregate results confirm PAKTON’s superior overall performance. 
The most pronounced advantage is observed in “Completeness,” where PAKTON consistently provides more comprehensive responses. PAKTON also excels in “Explainability and Reasoning,” offering clearer, stepwise rationales, an important attribute for non-specialist users. To assess the participants' discriminative capacity, we include competing criteria such as “Completeness” versus “Conciseness and Precision,” recognizing the inherent tension between exhaustive and succinct responses; as expected, PAKTON excels in completeness, while GPT-4o is preferred for conciseness. For “Relevance and Focus,” both models receive comparable scores, reflecting a trade-off between breadth and specificity. These outcomes confirm PAKTON’s core objective to generate detailed, report-like answers, with the observed completeness-precision trade-off both anticipated and justified. 

% \begin{figure*}[ht]
%     \centering
%     \includegraphics[width=0.9\textwidth]{static/HumanEvaluationBarChart.png}
%     \caption{Model preference by criterion aggregated for all questions.}
%     \label{fig:geval_human_barplot}
% \end{figure*}

% \begin{figure*}[ht]
%     \centering
%     \includegraphics[width=1\textwidth, trim=300 400 300 300, clip]{static/votes-chart-HD.png}
%     \caption{\small Human preferences across nine evaluation criteria comparing PAKTON and GPT-4o, aggregated for all questions.}
%     \label{fig:geval_human_barplot}
% \end{figure*}

% \begin{figure*}[ht]
%     \centering
%     \includegraphics[width=1\textwidth, trim=300 280 300 300, clip]{static/votes-chart-HD.jpg}
%     \caption{\small Human preferences across nine evaluation criteria comparing PAKTON and GPT-4o, aggregated for all questions.}
%     \label{fig:geval_human_barplot}
% \end{figure*}

\subsection{Evaluation using LLMs}
\label{subsec:geval_evaluation}

\paragraph{Setup.} To comprehensively assess PAKTON at scale, we employ G-EVAL~\citep{liu-etal-2023-g}, a widely used LLM-as-a-Judge evaluation framework~\citep{liusie-etal-2024-llm, wang-etal-2024-revisiting, chiang-lee-2023-large} for NLG output quality across the same criteria as the human study. We evaluate 102 randomly selected ContractNLI samples, comparing PAKTON and GPT-4o on matched inputs. G-EVAL generates criterion-specific scores, supporting fine-grained analysis of response quality, explainability, and reasoning. Further details are demonstrated in Appendix~\ref{app:geval}.

% To thougly evaluate PAKTON at scale we also used an automated framework named G-EVAL. G-EVAL ~\citep{liu-etal-2023-g} has emerged as a widely adopted ~\citep{liusie-etal-2024-llm, wang-etal-2024-revisiting} evaluation framework grounded in the LLM-as-a-Judge paradigm ~\citep{chiang-lee-2023-large}, which leverages large language models to assess the quality of natural language generation (NLG) outputs across the same dimensions used in human survey.
% The evaluation was conducted on 102 samples randomly drawn from the ContractNLI dataset ~\citep{koreeda-manning-2021-contractnli-dataset}. Once again, PAKTON and GPT-4o were evaluated on identical inputs, and the framework produced criterion-specific scores for each response, enabling a more nuanced analysis of model behavior beyond accuracy, particularly in terms of response quality, explainability, and reasoning. Additional details can be found in Appendix~\ref{app:geval}.

% As part of our evaluation design, we ensured fair model comparison settings. For ChatGPT, we implemented a (RAG) pipeline. For PAKTON, we limited tool usage to strictly in-document retrieval (disabling access to external tools like web search), and capped the number of interrogation turns at five. GPT-4o served as the underlying model for both systems to eliminate base model performance discrepancies and isolate differences due to architecture and orchestration.

\paragraph{Results} across all samples and evaluation criteria prove PAKTON's superiority over GPT-4o in eight out of nine dimensions (Figure ~\ref{fig:combined_comparison}): PAKTON scores higher in “Explainability and Reasoning”, “Justification with Evidence”, “Completeness”, and “Handling Ambiguity”. These outcomes are consistent with PAKTON’s design objectives that prioritize detailed and well-supported responses that explicitly reason through legal content. For  “Conciseness and Precision”, and “Relevance and Focus” criteria, both models demonstrate comparable performance, an  anticipated outcome, as these dimensions often conflict with “Completeness”, a domain in which PAKTON significantly surpasses GPT-4o, thereby illustrating the inherent trade-off between brevity and depth.
The sole criterion in which PAKTON underperforms relative to GPT-4o is “Contextual and Legal Understanding”; while this initially appears counterintuitive, a closer examination of the G-EVAL rationale reveals that PAKTON’s responses frequently acknowledge knowledge gaps under uncertainty. Although this is desirable from a transparency perspective, the evaluation framework interprets such acknowledgments as evidence of limited understanding, resulting in lower scores. %Further details are provided in Appendix \ref{app:geval}.
The alignment between G-EVAL and human judgments is evident in Fig. ~\ref{fig:combined_comparison}, where results for both methods appear nearly superimposed. A more rigorous statistical analysis that strengthens and intervalidates this agreement is presented in App. \ref{app:statistical_agreement}.

\section{Conclusion}

We propose \textsc{PAKTON}, a multi-agent framework for contract analysis that promotes explainability, grounded reasoning, and modular retrieval. By coordinating three specialized agents in a model-agnostic manner, and without further tuning needed, \textsc{PAKTON} generates transparent, well-justified, and query-specific legal reports. %, and is readily adaptable to diverse deployment settings.
Automated and human evaluations prove \textsc{PAKTON}'s  superiority against both prompted and fine-tuned baselines, not only in accuracy but also in critical dimensions such as explainability, completeness, and evidentiary support—core requirements for legal applications characterized by ambiguity and high-stakes interpretation.

\section*{Acknowledgments}
%This work was carried out within the framework of the Pharos AI Factory project, funded by the European High-Performance Computing Joint Undertaking (EuroHPC JU) under Grant Agreement No. 101234269 as part of the Horizon Europe and by the Greek Public Investments Program programme.

This work was supported by the FCT project ``OptiGov'', ref. 2024.07385.IACDC (DOI 10.54499/2024.07385.IACDC), funded by the PRR under the measure RE-C05-i08.m04.

\section*{Limitations}

While PAKTON demonstrates strong capabilities in contract analysis, it has several limitations that should be considered when interpreting its results and deploying it in real-world scenarios.

\textbf{Language Scope.}  
Our system has been tested only on English-language contracts. As legal language varies significantly across languages and cultures, additional adaptation and evaluation would be necessary for multilingual or cross-lingual applications.

\textbf{Contract and Jurisdiction Coverage.}  
PAKTON has been evaluated on a subset of contract types and does not currently cover the full diversity of legal documents. Similarly, the system has not been tested across different legal jurisdictions. These factors may affect the system’s generalizability and legal relevance in broader contexts.

\textbf{Latency and Cost.}  
Given our focus on the quality and depth of the generated reports, the system prioritizes multi-step reasoning over speed. As a result, response times may be longer compared to general-purpose language models, particularly due to the iterative communication between agents. This design also increases computational cost, making it less suitable for low-latency or resource-constrained environments.

\textbf{Explainability vs. Efficiency Tradeoff.}  
Our framework is explicitly designed to enhance transparency and reasoning. However, this emphasis on explainability can sometimes result in longer or less concise responses. In prioritizing clarity and justification, the system may occasionally sacrifice brevity or even slightly reduce precision, especially in cases where ambiguity is high and reasoning chains are extended.

\textbf{Structural Parsing Generalization.}  
The system’s structural parsing component is optimized for standard contract formats, which follow consistent patterns of headings, clauses, and subclauses. When documents deviate significantly from these conventions or lack a clearly defined structure, the benefits of structural parsing are reduced. In such cases, the parsing mechanism defaults to a more general-purpose chunking strategy. This fallback does not hinder the system’s functionality but may limit the advantages gained from fine-grained hierarchical representation.

\section*{Ethical Considerations}

\textbf{Legal Expertise and Overreliance.}  
Our system is developed to aid in contract analysis and increase access to legal information, but it does not serve as a substitute for qualified legal advice. There is a risk that users, particularly non-experts, may over-rely on its outputs without proper legal verification. To mitigate this, we recommend clearly communicating the system’s limitations and encouraging users to consult legal professionals when making important decisions. \textbf{PAKTON should be viewed as an assistive tool, not a definitive authority on legal interpretation}.

\textbf{Accessibility.}  
Contract analysis is often inaccessible to non-professionals due to its complexity and reliance on legal expertise (as discussed previously). PAKTON's goal is to help democratize contract understanding by providing explainable, user-friendly outputs that can assist individuals without legal backgrounds. We are also trying to offer free access to a publicly deployed version of the system, to the extent that it remains practically and financially feasible. Nevertheless, disparities in access to computational resources may limit the ability of low-resourced groups to deploy or benefit from the system effectively. Future iterations should explore lightweight deployments support to enhance accessibility across a broader range of users.

\textbf{Security concerns and misuse.}  
As with any system built on large language models, PAKTON may be susceptible to potential misuse and adversarial attacks. Malicious users might attempt to exploit the system to bypass safeguards, distort outputs, or extract sensitive information from retrieved content. While we employ various techniques to reduce some of these risks, further work is needed to strengthen the system’s robustness against such threats. We strongly recommend responsible deployment practices and ongoing monitoring when integrating our system into real-world applications.

\textbf{Human Evaluation.}  
Our human evaluation was carried out by anonymous volunteers who were fairly compensated for their time. All participants were fully informed about the purpose of the study and followed a structured annotation protocol. No personal data was collected or stored at any stage of the evaluation.

\bibliography{custom}

\begin{thebibliography}{72}
\providecommand{\natexlab}[1]{#1}

\bibitem[{Achiam et~al.(2023)Achiam, Adler, Agarwal, Ahmad, Akkaya, Aleman, Almeida, Altenschmidt, Altman, Anadkat, Avila, Babuschkin, Balaji, Balcom, Baltescu, ing Bao, Bavarian, Belgum, and et~al.}]{Achiam2023GPT4TR}
OpenAI~Josh Achiam, Steven Adler, Sandhini Agarwal, Lama Ahmad, Ilge Akkaya, Florencia~Leoni Aleman, Diogo Almeida, Janko Altenschmidt, Sam Altman, Shyamal Anadkat, Red Avila, Igor Babuschkin, Suchir Balaji, Valerie Balcom, Paul Baltescu, Haim ing Bao, Mo~Bavarian, Jeff Belgum, and Irwan~Bello et~al. 2023.
\newblock \href {https://api.semanticscholar.org/CorpusID:257532815} {Gpt-4 technical report}.

\bibitem[{Aires et~al.(2019)Aires, Granada, and Meneguzzi}]{aires2019concon}
Jo{\~a}o~Paulo Aires, Roger~Leitzke Granada, and Felipe~Rech Meneguzzi. 2019.
\newblock Concon: a contract conflict identifier.
\newblock In \emph{Proceedings of the 18th International Conference on Autonomous Agents and Multiagent Systems, 2019, Canad{\'a}.}

\bibitem[{Akarajaradwong et~al.(2025)Akarajaradwong, Pothavorn, Chaksangchaichot, Tasawong, Nopparatbundit, and Nutanong}]{akarajaradwong2025nitibench}
Pawitsapak Akarajaradwong, Pirat Pothavorn, Chompakorn Chaksangchaichot, Panuthep Tasawong, Thitiwat Nopparatbundit, and Sarana Nutanong. 2025.
\newblock Nitibench: A comprehensive study of llm framework capabilities for thai legal question answering.
\newblock \emph{arXiv preprint arXiv:2502.10868}.

\bibitem[{Anthropic(2024{\natexlab{a}})}]{anthropic2024contextual}
Anthropic. 2024{\natexlab{a}}.
\newblock Introducing contextual retrieval.
\newblock \url{https://www.anthropic.com/news/contextual-retrieval}.

\bibitem[{Anthropic(2024{\natexlab{b}})}]{anthropic2024modelcontext}
Anthropic. 2024{\natexlab{b}}.
\newblock \href {https://www.anthropic.com/news/model-context-protocol} {Introducing the model context protocol}.
\newblock Accessed: 2025-04-17.

\bibitem[{Anthropic(2025)}]{Claude3S}
Anthropic. 2025.
\newblock \href {https://api.semanticscholar.org/CorpusID:276612236} {Claude 3.7 sonnet system card}.

\bibitem[{Bianchini et~al.(2024)Bianchini, Calamo, De~Luzi, Macr{\`\i}, and Mecella}]{bianchini2024enhancing}
Filippo Bianchini, Marco Calamo, Francesca De~Luzi, Mattia Macr{\`\i}, and Massimo Mecella. 2024.
\newblock Enhancing complex linguistic tasks resolution through fine-tuning llms, rag and knowledge graphs (short paper).
\newblock In \emph{International Conference on Advanced Information Systems Engineering}, pages 147--155. Springer.

\bibitem[{Brown et~al.(2020)Brown, Mann, Ryder, Subbiah, Kaplan, Dhariwal, Neelakantan, Shyam, Sastry, Askell, Agarwal, Herbert-Voss, Krueger, Henighan, Child, Ramesh, Ziegler, Wu, Winter, Hesse, Chen, Sigler, teusz Litwin, Gray, Chess, Clark, Berner, McCandlish, Radford, Sutskever, and Amodei}]{Brown2020LanguageMA}
Tom~B. Brown, Benjamin Mann, Nick Ryder, Melanie Subbiah, Jared Kaplan, Prafulla Dhariwal, Arvind Neelakantan, Pranav Shyam, Girish Sastry, Amanda Askell, Sandhini Agarwal, Ariel Herbert-Voss, Gretchen Krueger, Tom Henighan, Rewon Child, Aditya Ramesh, Daniel~M. Ziegler, Jeff Wu, Clemens Winter, and 12 others. 2020.
\newblock \href {https://api.semanticscholar.org/CorpusID:218971783} {Language models are few-shot learners}.
\newblock \emph{ArXiv}, abs/2005.14165.

\bibitem[{Chiang and Lee(2023)}]{chiang-lee-2023-large}
Cheng-Han Chiang and Hung-yi Lee. 2023.
\newblock \href {https://doi.org/10.18653/v1/2023.acl-long.870} {Can large language models be an alternative to human evaluations?}
\newblock In \emph{Proceedings of the 61st Annual Meeting of the Association for Computational Linguistics (Volume 1: Long Papers)}, pages 15607--15631, Toronto, Canada. Association for Computational Linguistics.

\bibitem[{Colombo et~al.(2024{\natexlab{a}})Colombo, Pires, Boudiaf, Culver, Melo, Corro, Martins, Esposito, Raposo, Morgado, and Desa}]{Colombo2024SaulLM7BAP}
Pierre Colombo, Telmo~Pessoa Pires, Malik Boudiaf, Dominic Culver, Rui Melo, Caio Corro, Andr{\'e} Martins, Fabrizio Esposito, Vera~L'ucia Raposo, Sofia Morgado, and Michael Desa. 2024{\natexlab{a}}.
\newblock \href {https://api.semanticscholar.org/CorpusID:268253108} {Saullm-7b: A pioneering large language model for law}.
\newblock \emph{ArXiv}, abs/2403.03883.

\bibitem[{Colombo et~al.(2024{\natexlab{b}})Colombo, Pires, Boudiaf, Melo, Culver, Morgado, Malaboeuf, Hautreux, Charpentier, and Desa}]{Colombo2024SaulLM54BS}
Pierre Colombo, Telmo~Pessoa Pires, Malik Boudiaf, Rui Melo, Dominic Culver, Sofia Morgado, Etienne Malaboeuf, Gabriel Hautreux, Johanne Charpentier, and Michael Desa. 2024{\natexlab{b}}.
\newblock \href {https://api.semanticscholar.org/CorpusID:271533610} {Saullm-54b \& saullm-141b: Scaling up domain adaptation for the legal domain}.
\newblock \emph{ArXiv}, abs/2407.19584.

\bibitem[{Cormack et~al.(2009)Cormack, Clarke, and B{\"u}ttcher}]{Cormack2009ReciprocalRF}
Gordon~V. Cormack, Charles L.~A. Clarke, and Stefan B{\"u}ttcher. 2009.
\newblock \href {https://api.semanticscholar.org/CorpusID:12408211} {Reciprocal rank fusion outperforms condorcet and individual rank learning methods}.
\newblock \emph{Proceedings of the 32nd international ACM SIGIR conference on Research and development in information retrieval}.

\bibitem[{Cui et~al.(2023)Cui, Li, Yan, Chen, and Yuan}]{Cui2023ChatlawAM}
Jiaxi Cui, Zongjia Li, Yang Yan, Bohua Chen, and Li~Yuan. 2023.
\newblock \href {https://api.semanticscholar.org/CorpusID:259274889} {Chatlaw: A multi-agent collaborative legal assistant with knowledge graph enhanced mixture-of-experts large language model}.

\bibitem[{DeepSeek-AI et~al.(2025)DeepSeek-AI, Liu, Feng, Xue, Wang, Wu, Lu, Zhao, Deng, Zhang, Ruan, Dai, Guo, Yang, Chen, and et~al.}]{deepseek}
DeepSeek-AI, Aixin Liu, Bei Feng, Bing Xue, Bingxuan Wang, Bochao Wu, Chengda Lu, Chenggang Zhao, Chengqi Deng, Chenyu Zhang, Chong Ruan, Damai Dai, Daya Guo, Dejian Yang, Deli Chen, and Dongjie~Ji et~al. 2025.
\newblock \href {https://arxiv.org/abs/2412.19437} {Deepseek-v3 technical report}.
\newblock \emph{Preprint}, arXiv:2412.19437.

\bibitem[{Duarte et~al.(2024)Duarte, Marques, Gra{\c{c}}a, Freire, Li, and Oliveira}]{duarte-etal-2024-lumberchunker}
Andr{\'e}~V. Duarte, Jo{\~a}o~DS Marques, Miguel Gra{\c{c}}a, Miguel Freire, Lei Li, and Arlindo~L. Oliveira. 2024.
\newblock \href {https://doi.org/10.18653/v1/2024.findings-emnlp.377} {{L}umber{C}hunker: Long-form narrative document segmentation}.
\newblock In \emph{Findings of the Association for Computational Linguistics: EMNLP 2024}, pages 6473--6486, Miami, Florida, USA. Association for Computational Linguistics.

\bibitem[{Edge et~al.(2024)Edge, Trinh, Cheng, Bradley, Chao, Mody, Truitt, and Larson}]{Edge2024FromLT}
Darren Edge, Ha~Trinh, Newman Cheng, Joshua Bradley, Alex Chao, Apurva Mody, Steven Truitt, and Jonathan Larson. 2024.
\newblock \href {https://api.semanticscholar.org/CorpusID:269363075} {From local to global: A graph rag approach to query-focused summarization}.
\newblock \emph{ArXiv}, abs/2404.16130.

\bibitem[{Eisenberg(2022)}]{Eisenberg2022LegalR}
Melvin~A. Eisenberg. 2022.
\newblock \href {https://api.semanticscholar.org/CorpusID:262079114} {Legal reasoning}.

\bibitem[{Fan et~al.(2024)Fan, Ding, bo~Ning, Wang, Li, Yin, Chua, and Li}]{Fan2024ASO}
Wenqi Fan, Yujuan Ding, Liang bo~Ning, Shijie Wang, Hengyun Li, Dawei Yin, Tat-Seng Chua, and Qing Li. 2024.
\newblock \href {https://api.semanticscholar.org/CorpusID:269740933} {A survey on rag meeting llms: Towards retrieval-augmented large language models}.
\newblock In \emph{Knowledge Discovery and Data Mining}.

\bibitem[{Frei(2016)}]{Frei2016AnAI}
Thomas Frei. 2016.
\newblock \href {https://api.semanticscholar.org/CorpusID:8873886} {An artificial intelligence approach to legal reasoning}.

\bibitem[{Gao et~al.(2023)Gao, Xiong, Gao, Jia, Pan, Bi, Dai, Sun, Guo, Wang, and Wang}]{Gao2023RetrievalAugmentedGF}
Yunfan Gao, Yun Xiong, Xinyu Gao, Kangxiang Jia, Jinliu Pan, Yuxi Bi, Yi~Dai, Jiawei Sun, Qianyu Guo, Meng Wang, and Haofen Wang. 2023.
\newblock \href {https://api.semanticscholar.org/CorpusID:266359151} {Retrieval-augmented generation for large language models: A survey}.
\newblock \emph{ArXiv}, abs/2312.10997.

\bibitem[{Gao et~al.(2024)Gao, Xiong, Gao, Jia, Pan, Bi, Dai, Sun, Wang, and Wang}]{gao2024retrievalaugmentedgenerationlargelanguage}
Yunfan Gao, Yun Xiong, Xinyu Gao, Kangxiang Jia, Jinliu Pan, Yuxi Bi, Yi~Dai, Jiawei Sun, Meng Wang, and Haofen Wang. 2024.
\newblock \href {https://arxiv.org/abs/2312.10997} {Retrieval-augmented generation for large language models: A survey}.
\newblock \emph{Preprint}, arXiv:2312.10997.

\bibitem[{Grattafiori et~al.(2024)Grattafiori, Dubey, Jauhri, Pandey, Kadian, Al-Dahle, Letman, Mathur, Schelten, Vaughan, Yang, Fan, Goyal, Hartshorn, Yang, Mitra, Sravankumar, Korenev, Hinsvark, Rao, Zhang, Rodriguez, Gregerson, and et~al.}]{llama3}
Aaron Grattafiori, Abhimanyu Dubey, Abhinav Jauhri, Abhinav Pandey, Abhishek Kadian, Ahmad Al-Dahle, Aiesha Letman, Akhil Mathur, Alan Schelten, Alex Vaughan, Amy Yang, Angela Fan, Anirudh Goyal, Anthony Hartshorn, Aobo Yang, Archi Mitra, Archie Sravankumar, Artem Korenev, Arthur Hinsvark, and 5 others. 2024.
\newblock \href {https://arxiv.org/abs/2407.21783} {The llama 3 herd of models}.
\newblock \emph{Preprint}, arXiv:2407.21783.

\bibitem[{Guo et~al.(2024)Guo, Xia, Yu, Ao, and Huang}]{Guo2024LightRAGSA}
Zirui Guo, Lianghao Xia, Yanhua Yu, Tu~Ao, and Chao Huang. 2024.
\newblock \href {https://api.semanticscholar.org/CorpusID:273227829} {Lightrag: Simple and fast retrieval-augmented generation}.
\newblock \emph{ArXiv}, abs/2410.05779.

\bibitem[{Hendrycks et~al.(2021)Hendrycks, Burns, Chen, and Ball}]{cuad}
Dan Hendrycks, Collin Burns, Anya Chen, and Spencer Ball. 2021.
\newblock Cuad: An expert-annotated nlp dataset for legal contract review.
\newblock \emph{NeurIPS}.

\bibitem[{Huang et~al.(2023)Huang, Tao, Zhang, An, Jiang, Chen, Wu, and Feng}]{huang2023lawyer}
Quzhe Huang, Mingxu Tao, Chen Zhang, Zhenwei An, Cong Jiang, Zhibin Chen, Zirui Wu, and Yansong Feng. 2023.
\newblock Lawyer llama technical report.
\newblock \emph{arXiv preprint arXiv:2305.15062}.

\bibitem[{Ichida and Meneguzzi(2021)}]{ichida2021detecting}
Alexandre~Yukio Ichida and Felipe Meneguzzi. 2021.
\newblock Detecting logical relation in contract clauses.
\newblock \emph{arXiv preprint arXiv:2111.01856}.

\bibitem[{{Institute for Supply Management}()}]{ism2024homepage}
{Institute for Supply Management}.
\newblock Institute for supply management (ism).
\newblock \url{https://www.ismworld.org/}.
\newblock Accessed: 2025-04-01.

\bibitem[{Jiang et~al.(2023)Jiang, Sablayrolles, Mensch, Bamford, Chaplot, de~las Casas, Bressand, Lengyel, Lample, Saulnier, Lavaud, Lachaux, Stock, Scao, Lavril, Wang, Lacroix, and Sayed}]{mistral}
Albert~Q. Jiang, Alexandre Sablayrolles, Arthur Mensch, Chris Bamford, Devendra~Singh Chaplot, Diego de~las Casas, Florian Bressand, Gianna Lengyel, Guillaume Lample, Lucile Saulnier, Lélio~Renard Lavaud, Marie-Anne Lachaux, Pierre Stock, Teven~Le Scao, Thibaut Lavril, Thomas Wang, Timothée Lacroix, and William~El Sayed. 2023.
\newblock \href {https://arxiv.org/abs/2310.06825} {Mistral 7b}.
\newblock \emph{Preprint}, arXiv:2310.06825.

\bibitem[{Jiang et~al.(2024)Jiang, Shao, Ma, Semnani, and Lam}]{jiang-etal-2024-unknown}
Yucheng Jiang, Yijia Shao, Dekun Ma, Sina Semnani, and Monica Lam. 2024.
\newblock \href {https://doi.org/10.18653/v1/2024.emnlp-main.554} {Into the unknown unknowns: Engaged human learning through participation in language model agent conversations}.
\newblock In \emph{Proceedings of the 2024 Conference on Empirical Methods in Natural Language Processing}, pages 9917--9955, Miami, Florida, USA. Association for Computational Linguistics.

\bibitem[{Kant et~al.(2025)Kant, Nabi, Kant, Scharrer, Ma, and Nabi}]{kant2025towards}
Manuj Kant, Sareh Nabi, Manav Kant, Roland Scharrer, Megan Ma, and Marzieh Nabi. 2025.
\newblock Towards robust legal reasoning: Harnessing logical llms in law.
\newblock \emph{arXiv preprint arXiv:2502.17638}.

\bibitem[{Karpukhin et~al.(2020)Karpukhin, Oğuz, Min, Lewis, Wu, Edunov, Chen, and tau Yih}]{Karpukhin2020DensePR}
Vladimir Karpukhin, Barlas Oğuz, Sewon Min, Patrick Lewis, Ledell~Yu Wu, Sergey Edunov, Danqi Chen, and Wen tau Yih. 2020.
\newblock \href {https://api.semanticscholar.org/CorpusID:215737187} {Dense passage retrieval for open-domain question answering}.
\newblock \emph{ArXiv}, abs/2004.04906.

\bibitem[{Koreeda and Manning(2021)}]{koreeda-manning-2021-contractnli-dataset}
Yuta Koreeda and Christopher Manning. 2021.
\newblock \href {https://doi.org/10.18653/v1/2021.findings-emnlp.164} {{C}ontract{NLI}: A dataset for document-level natural language inference for contracts}.
\newblock In \emph{Findings of the Association for Computational Linguistics: EMNLP 2021}, pages 1907--1919, Punta Cana, Dominican Republic. Association for Computational Linguistics.

\bibitem[{Lai et~al.(2024)Lai, Gan, Wu, Qi, and Yu}]{lai2024large}
Jinqi Lai, Wensheng Gan, Jiayang Wu, Zhenlian Qi, and Philip~S Yu. 2024.
\newblock Large language models in law: A survey.
\newblock \emph{AI Open}.

\bibitem[{{LangChain}(2024)}]{langchain2024rts}
{LangChain}. 2024.
\newblock Recursive text splitter.
\newblock \url{https://python.langchain.com/v0.1/docs/modules/data_connection/document_transformers/recursive_text_splitter/}.
\newblock Accessed: 2024-08-15.

\bibitem[{Lewis et~al.(2020)Lewis, Perez, Piktus, Petroni, Karpukhin, Goyal, Kuttler, Lewis, tau Yih, Rockt{\"a}schel, Riedel, and Kiela}]{Lewis2020RetrievalAugmentedGF}
Patrick Lewis, Ethan Perez, Aleksandara Piktus, Fabio Petroni, Vladimir Karpukhin, Naman Goyal, Heinrich Kuttler, Mike Lewis, Wen tau Yih, Tim Rockt{\"a}schel, Sebastian Riedel, and Douwe Kiela. 2020.
\newblock \href {https://api.semanticscholar.org/CorpusID:218869575} {Retrieval-augmented generation for knowledge-intensive nlp tasks}.
\newblock \emph{ArXiv}, abs/2005.11401.

\bibitem[{Liu et~al.(2025)Liu, Liu, Yao, Liu, Meng, Wang, and Ma}]{liu2025hm}
Pei Liu, Xin Liu, Ruoyu Yao, Junming Liu, Siyuan Meng, Ding Wang, and Jun Ma. 2025.
\newblock Hm-rag: Hierarchical multi-agent multimodal retrieval augmented generation.
\newblock \emph{arXiv preprint arXiv:2504.12330}.

\bibitem[{Liu et~al.(2023)Liu, Iter, Xu, Wang, Xu, and Zhu}]{liu-etal-2023-g}
Yang Liu, Dan Iter, Yichong Xu, Shuohang Wang, Ruochen Xu, and Chenguang Zhu. 2023.
\newblock \href {https://doi.org/10.18653/v1/2023.emnlp-main.153} {{G}-eval: {NLG} evaluation using gpt-4 with better human alignment}.
\newblock In \emph{Proceedings of the 2023 Conference on Empirical Methods in Natural Language Processing}, pages 2511--2522, Singapore. Association for Computational Linguistics.

\bibitem[{Liusie et~al.(2024)Liusie, Manakul, and Gales}]{liusie-etal-2024-llm}
Adian Liusie, Potsawee Manakul, and Mark Gales. 2024.
\newblock \href {https://aclanthology.org/2024.eacl-long.8/} {{LLM} comparative assessment: Zero-shot {NLG} evaluation through pairwise comparisons using large language models}.
\newblock In \emph{Proceedings of the 18th Conference of the European Chapter of the Association for Computational Linguistics (Volume 1: Long Papers)}, pages 139--151, St. Julian{'}s, Malta. Association for Computational Linguistics.

\bibitem[{Marques et~al.(2024)Marques, de~Lamare Bastian-Pinto, and Brand{\~a}o}]{marques2024contract}
Naielly~Lopes Marques, Carlos de~Lamare Bastian-Pinto, and Luiz Eduardo~Teixeira Brand{\~a}o. 2024.
\newblock When contract clauses conflict: The salvador light rail vehicle concession.
\newblock \emph{Case Studies on Transport Policy}, 16:101168.

\bibitem[{Mekala et~al.(2024)Mekala, Razeghi, and Singh}]{mekala-etal-2024-echoprompt}
Raja Sekhar~Reddy Mekala, Yasaman Razeghi, and Sameer Singh. 2024.
\newblock \href {https://doi.org/10.18653/v1/2024.naacl-short.35} {{E}cho{P}rompt: Instructing the model to rephrase queries for improved in-context learning}.
\newblock In \emph{Proceedings of the 2024 Conference of the North American Chapter of the Association for Computational Linguistics: Human Language Technologies (Volume 2: Short Papers)}, pages 399--432, Mexico City, Mexico. Association for Computational Linguistics.

\bibitem[{{OpenAI}(2024{\natexlab{a}})}]{openai2024embedding}
{OpenAI}. 2024{\natexlab{a}}.
\newblock Embedding models.
\newblock \url{https://platform.openai.com/docs/guides/embeddings/embedding-models}.
\newblock Accessed: 2024-08-15.

\bibitem[{{OpenAI}(2024{\natexlab{b}})}]{openai_file_uploads}
{OpenAI}. 2024{\natexlab{b}}.
\newblock \href {https://help.openai.com/en/articles/10029836-optimizing-file-uploads-in-chatgpt-enterprise} {Optimizing file uploads in chatgpt enterprise}.
\newblock Accessed: 2025-04-12.

\bibitem[{Pipitone and Alami(2024)}]{pipitone2024legalbenchragbenchmarkretrievalaugmentedgeneration}
Nicholas Pipitone and Ghita~Houir Alami. 2024.
\newblock \href {https://arxiv.org/abs/2408.10343} {Legalbench-rag: A benchmark for retrieval-augmented generation in the legal domain}.
\newblock \emph{Preprint}, arXiv:2408.10343.

\bibitem[{Qwen et~al.(2025)Qwen, :, Yang, Yang, Zhang, Hui, Zheng, Yu, Li, Liu, Huang, Wei, Lin, Yang, and et~al.}]{quen}
Qwen, :, An~Yang, Baosong Yang, Beichen Zhang, Binyuan Hui, Bo~Zheng, Bowen Yu, Chengyuan Li, Dayiheng Liu, Fei Huang, Haoran Wei, Huan Lin, Jian Yang, and Jianhong~Tu et~al. 2025.
\newblock \href {https://arxiv.org/abs/2412.15115} {Qwen2.5 technical report}.
\newblock \emph{Preprint}, arXiv:2412.15115.

\bibitem[{Ravichander et~al.(2019)Ravichander, Black, Wilson, Norton, and Sadeh}]{privacyqa}
Abhilasha Ravichander, Alan~W Black, Shomir Wilson, Thomas Norton, and Norman Sadeh. 2019.
\newblock \href {https://arxiv.org/abs/1911.00841} {Question answering for privacy policies: Combining computational and legal perspectives}.
\newblock \emph{Preprint}, arXiv:1911.00841.

\bibitem[{Robertson and Zaragoza(2009)}]{Robertson2009ThePR}
Stephen~E. Robertson and Hugo Zaragoza. 2009.
\newblock \href {https://api.semanticscholar.org/CorpusID:207178704} {The probabilistic relevance framework: Bm25 and beyond}.
\newblock \emph{Found. Trends Inf. Retr.}, 3:333--389.

\bibitem[{Ryu et~al.(2023)Ryu, Lee, Pang, Choi, Choi, Min, and Sohn}]{ryu2023retrieval}
Cheol Ryu, Seolhwa Lee, Subeen Pang, Chanyeol Choi, Hojun Choi, Myeonggee Min, and Jy-Yong Sohn. 2023.
\newblock Retrieval-based evaluation for llms: a case study in korean legal qa.
\newblock In \emph{Proceedings of the Natural Legal Language Processing Workshop 2023}, pages 132--137.

\bibitem[{Saha et~al.(2024)Saha, Saha, and Malik}]{saha2024advancing}
Binita Saha, Utsha Saha, and Muhammad~Zubair Malik. 2024.
\newblock Advancing retrieval-augmented generation with inverted question matching for enhanced qa performance.
\newblock \emph{IEEE Access}.

\bibitem[{Shao et~al.(2024)Shao, Jiang, Kanell, Xu, Khattab, and Lam}]{Shao2024AssistingIW}
Yijia Shao, Yucheng Jiang, Theodore~A. Kanell, Peter Xu, Omar Khattab, and Monica~S. Lam. 2024.
\newblock \href {https://api.semanticscholar.org/CorpusID:267782917} {Assisting in writing wikipedia-like articles from scratch with large language models}.
\newblock In \emph{North American Chapter of the Association for Computational Linguistics}.

\bibitem[{ShengbinYue et~al.(2025)ShengbinYue, Huang, Jia, Wang, Liu, Song, Huang, and Wei}]{shengbinyue2025multi}
ShengbinYue ShengbinYue, Ting Huang, Zheng Jia, Siyuan Wang, Shujun Liu, Yun Song, Xuan-Jing Huang, and Zhongyu Wei. 2025.
\newblock Multi-agent simulator drives language models for legal intensive interaction.
\newblock In \emph{Findings of the Association for Computational Linguistics: NAACL 2025}, pages 6537--6570.

\bibitem[{Shu et~al.(2024)Shu, Zhao, Liu, Demeter, Du, and Zhang}]{shu2024lawllm}
Dong Shu, Haoran Zhao, Xukun Liu, David Demeter, Mengnan Du, and Yongfeng Zhang. 2024.
\newblock Lawllm: Law large language model for the us legal system.
\newblock In \emph{Proceedings of the 33rd ACM International Conference on Information and Knowledge Management}, pages 4882--4889.

\bibitem[{Siino et~al.(2025)Siino, Falco, Croce, and Rosso}]{siino2025exploring}
Marco Siino, Mariana Falco, Daniele Croce, and Paolo Rosso. 2025.
\newblock Exploring llms applications in law: A literature review on current legal nlp approaches.
\newblock \emph{IEEE Access}.

\bibitem[{Smith(2007)}]{Smith2007AnOO}
Raymond~W. Smith. 2007.
\newblock \href {https://api.semanticscholar.org/CorpusID:7038773} {An overview of the tesseract ocr engine}.
\newblock \emph{Ninth International Conference on Document Analysis and Recognition (ICDAR 2007)}, 2:629--633.

\bibitem[{Smith(2013)}]{Smith2013HistoryOT}
Raymond~W. Smith. 2013.
\newblock \href {https://api.semanticscholar.org/CorpusID:35474080} {History of the tesseract ocr engine: what worked and what didn't}.
\newblock In \emph{Electronic imaging}.

\bibitem[{Sun et~al.(2024)Sun, Dai, Luo, Chang, and Li}]{sun2024lawluo}
Jingyun Sun, Chengxiao Dai, Zhongze Luo, Yangbo Chang, and Yang Li. 2024.
\newblock Lawluo: A chinese law firm co-run by llm agents.
\newblock \emph{arXiv preprint arXiv:2407.16252}.

\bibitem[{Team et~al.(2025)Team, Kamath, Ferret, Pathak, Vieillard, Merhej, Perrin, Matejovicova, Ramé, Rivière, Rouillard, Mesnard, Cideron, bastien Grill, Ramos, Yvinec, Casbon, Pot, and et~al.}]{gemma}
Gemma Team, Aishwarya Kamath, Johan Ferret, Shreya Pathak, Nino Vieillard, Ramona Merhej, Sarah Perrin, Tatiana Matejovicova, Alexandre Ramé, Morgane Rivière, Louis Rouillard, Thomas Mesnard, Geoffrey Cideron, Jean bastien Grill, Sabela Ramos, Edouard Yvinec, Michelle Casbon, Etienne Pot, and Ivo~Penchev et~al. 2025.
\newblock \href {https://arxiv.org/abs/2503.19786} {Gemma 3 technical report}.
\newblock \emph{Preprint}, arXiv:2503.19786.

\bibitem[{Veturi et~al.(2024)Veturi, Vaichal, Jagadheesh, Tripto, and Yan}]{veturi2024rag}
Sriram Veturi, Saurabh Vaichal, Reshma~Lal Jagadheesh, Nafis~Irtiza Tripto, and Nian Yan. 2024.
\newblock Rag based question-answering for contextual response prediction system.
\newblock \emph{arXiv preprint arXiv:2409.03708}.

\bibitem[{Wahidur et~al.(2025)Wahidur, Kim, Choi, Bhatti, and Lee}]{wahidur2025legal}
Rahman~SM Wahidur, Sumin Kim, Haeung Choi, David~S Bhatti, and Heung-No Lee. 2025.
\newblock Legal query rag.
\newblock \emph{IEEE Access}.

\bibitem[{Wang et~al.(2025)Wang, Chen, Yang, Huang, Dou, and Wei}]{Wang2025ChainofRetrievalAG}
Liang Wang, Haonan Chen, Nan Yang, Xiaolong Huang, Zhicheng Dou, and Furu Wei. 2025.
\newblock \href {https://api.semanticscholar.org/CorpusID:275906944} {Chain-of-retrieval augmented generation}.
\newblock \emph{ArXiv}, abs/2501.14342.

\bibitem[{Wang et~al.(2023)Wang, Scardigli, Tang, Chen, Levkin, Chen, Ball, Woodside, Zhang, and Hendrycks}]{maud}
Steven~H. Wang, Antoine Scardigli, Leonard Tang, Wei Chen, Dimitry Levkin, Anya Chen, Spencer Ball, Thomas Woodside, Oliver Zhang, and Dan Hendrycks. 2023.
\newblock \href {https://arxiv.org/abs/2301.00876} {Maud: An expert-annotated legal nlp dataset for merger agreement understanding}.
\newblock \emph{Preprint}, arXiv:2301.00876.

\bibitem[{Wang et~al.(2020)Wang, Wei, Dong, Bao, Yang, and Zhou}]{wang2020minilm}
Wenhui Wang, Furu Wei, Li~Dong, Hangbo Bao, Nan Yang, and Ming Zhou. 2020.
\newblock Minilm: Deep self-attention distillation for task-agnostic compression of pre-trained transformers.
\newblock \emph{Advances in neural information processing systems}, 33:5776--5788.

\bibitem[{Wang et~al.(2024{\natexlab{a}})Wang, Wang, Gao, Zhang, Wu, Xu, Shi, Wang, Li, Qian, Yin, Lv, Zheng, and Huang}]{Wang2024SearchingFB}
Xiaohua Wang, Zhenghua Wang, Xuan Gao, Feiran Zhang, Yixin Wu, Zhibo Xu, Tianyuan Shi, Zhengyuan Wang, Shizheng Li, Qi~Qian, Ruicheng Yin, Changze Lv, Xiaoqing Zheng, and Xuanjing Huang. 2024{\natexlab{a}}.
\newblock \href {https://api.semanticscholar.org/CorpusID:270870251} {Searching for best practices in retrieval-augmented generation}.
\newblock In \emph{Conference on Empirical Methods in Natural Language Processing}.

\bibitem[{Wang et~al.(2024{\natexlab{b}})Wang, Chen, Cai, Xu, and Zhao}]{wang-etal-2024-revisiting}
Yuqi Wang, Lyuhao Chen, Songcheng Cai, Zhijian Xu, and Yilun Zhao. 2024{\natexlab{b}}.
\newblock \href {https://doi.org/10.18653/v1/2024.emnlp-main.815} {Revisiting automated evaluation for long-form table question answering}.
\newblock In \emph{Proceedings of the 2024 Conference on Empirical Methods in Natural Language Processing}, pages 14696--14706, Miami, Florida, USA. Association for Computational Linguistics.

\bibitem[{Wei et~al.(2022)Wei, Wang, Schuurmans, Bosma, Chi, Xia, Le, and Zhou}]{Wei2022ChainOT}
Jason Wei, Xuezhi Wang, Dale Schuurmans, Maarten Bosma, Ed~H. Chi, F.~Xia, Quoc Le, and Denny Zhou. 2022.
\newblock \href {https://api.semanticscholar.org/CorpusID:246411621} {Chain of thought prompting elicits reasoning in large language models}.
\newblock \emph{ArXiv}, abs/2201.11903.

\bibitem[{Wiratunga et~al.()Wiratunga, Abeyratne, Lasal~Jayawardena, Massie, Nkisi-Orji, Weerasinghe, Liret, and Fleisch}]{wiratungacbr}
Nirmalie Wiratunga, Ramitha Abeyratne, Kyle~Martin Lasal~Jayawardena, Stewart Massie, Ikechukwu Nkisi-Orji, Ruvan Weerasinghe, Anne Liret, and Bruno Fleisch.
\newblock Cbr-rag: Case-based reasoning for retrieval augmented generation in llms for legal question answering.” arxiv, april 4, 2024.

\bibitem[{{World Commerce \& Contracting}(2020)}]{worldcc2020value}
{World Commerce \& Contracting}. 2020.
\newblock The real value of contracting.
\newblock \url{https://www.worldcc.com/Resources/Content-Search/Details/articleid/885}.
\newblock Accessed: 2025-04-01.

\bibitem[{Xu et~al.(2024)Xu, Tao, Shen, Xu, Xu, Long, Lou, and Ma}]{xu-etal-2024-reading}
Xiaohan Xu, Chongyang Tao, Tao Shen, Can Xu, Hongbo Xu, Guodong Long, Jian-Guang Lou, and Shuai Ma. 2024.
\newblock \href {https://doi.org/10.18653/v1/2024.emnlp-main.871} {Re-reading improves reasoning in large language models}.
\newblock In \emph{Proceedings of the 2024 Conference on Empirical Methods in Natural Language Processing}, pages 15549--15575, Miami, Florida, USA. Association for Computational Linguistics.

\bibitem[{Yao et~al.(2022)Yao, Zhao, Yu, Du, Shafran, Narasimhan, and Cao}]{Yao2022ReActSR}
Shunyu Yao, Jeffrey Zhao, Dian Yu, Nan Du, Izhak Shafran, Karthik Narasimhan, and Yuan Cao. 2022.
\newblock \href {https://api.semanticscholar.org/CorpusID:252762395} {React: Synergizing reasoning and acting in language models}.
\newblock \emph{ArXiv}, abs/2210.03629.

\bibitem[{Zhang et~al.(2025)Zhang, Yu, Sun, and Xu}]{zhang2025syler}
Kepu Zhang, Weijie Yu, Zhongxiang Sun, and Jun Xu. 2025.
\newblock Syler: A framework for explicit syllogistic legal reasoning in large language models.
\newblock \emph{arXiv preprint arXiv:2504.04042}.

\bibitem[{Zhao et~al.(2024{\natexlab{a}})Zhao, Zu, Hao, Lu, He, Ding, Gui, Zhang, and Huang}]{zhao2024longagent}
Jun Zhao, Can Zu, Xu~Hao, Yi~Lu, Wei He, Yiwen Ding, Tao Gui, Qi~Zhang, and Xuan-Jing Huang. 2024{\natexlab{a}}.
\newblock Longagent: Achieving question answering for 128k-token-long documents through multi-agent collaboration.
\newblock In \emph{Proceedings of the 2024 Conference on Empirical Methods in Natural Language Processing}, pages 16310--16324.

\bibitem[{Zhao et~al.(2024{\natexlab{b}})Zhao, Zhou, and Li}]{zhao2024chat2data}
Xinyang Zhao, Xuanhe Zhou, and Guoliang Li. 2024{\natexlab{b}}.
\newblock Chat2data: An interactive data analysis system with rag, vector databases and llms.
\newblock \emph{Proceedings of the VLDB Endowment}, 17(12):4481--4484.

\bibitem[{Zhou et~al.(2024)Zhou, Zhou, Gao, Huang, Huang, Zuo, and Zhao}]{zhou2024intelligent}
Hong Zhou, Li~Zhou, Binwei Gao, Wen Huang, Wenlu Huang, Jian Zuo, and Xianbo Zhao. 2024.
\newblock Intelligent identification of risks in construction contract clauses based on semantic reasoning.
\newblock \emph{Engineering, Construction and Architectural Management}.

\end{thebibliography}
% \clearpage
\appendix

\section{Framework's Implementation Details}
\label{app:framework_implementation_details}
\begin{figure*}[h!]
    \centering

    % First image
    \includegraphics[width=0.6\textwidth]{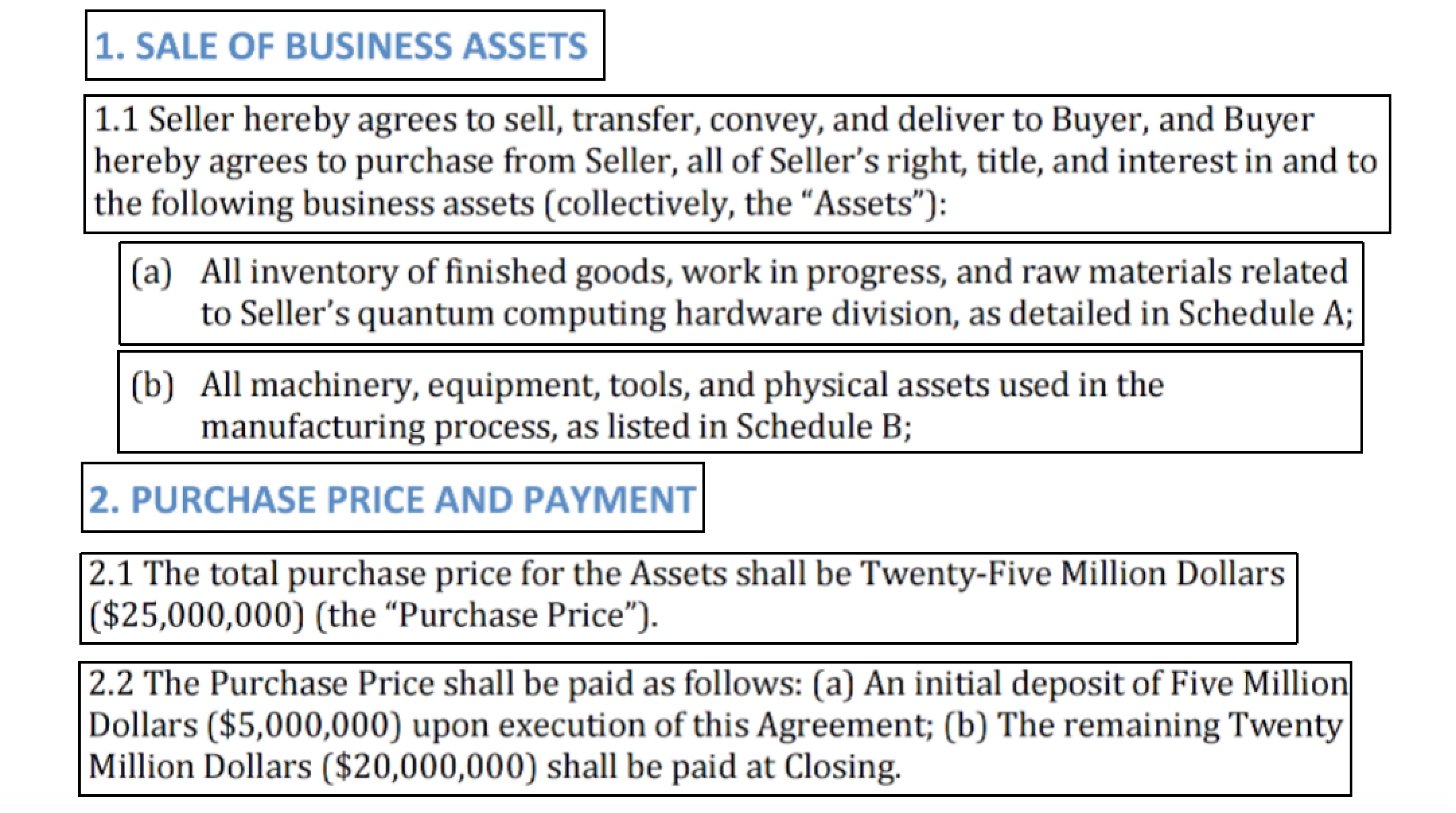}
    
    \vspace{0.5ex}
    (a) Section Detection 

    \vspace{2ex}

    % Second image
    \includegraphics[width=0.95\textwidth]{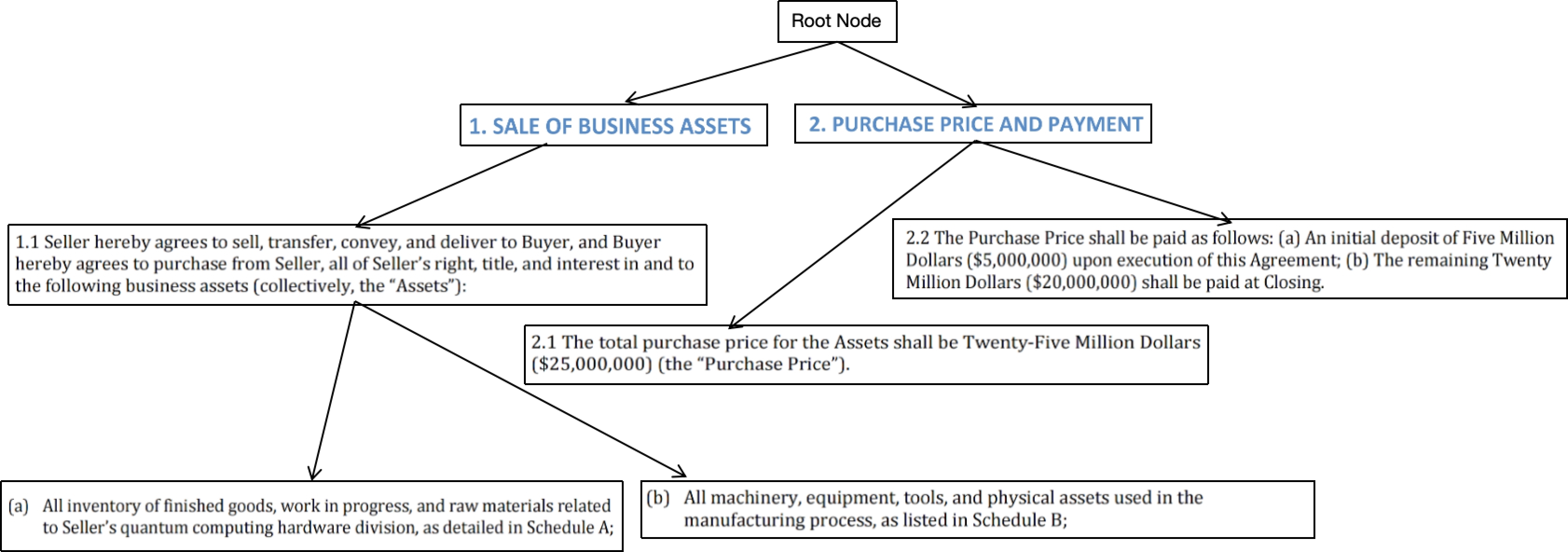}

    \vspace{0.5ex}
    (b) Hierarchical Organization of sections.

    \vspace{2ex}

    % Third image
    \includegraphics[width=1\textwidth]{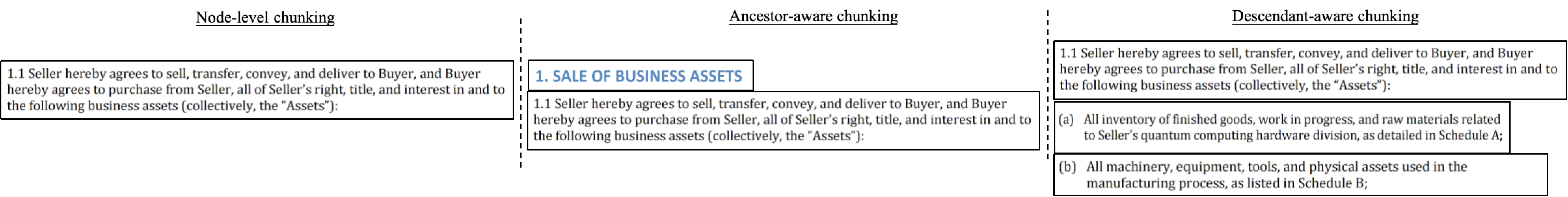}
    
    \vspace{0.5ex}
    (c) Contextual Embeddings for node "1.1 Seller ..."

    \vspace{2ex}
        
    \caption{Tree Representation and Chunking Workflow for Contract Documents}
    \label{fig:graph_representation}
\end{figure*}

Figure~\ref{fig:graph_representation} depicts the Tree Representation Process of the \textit{Archivist} as described in ~\ref{subsection:archivist}.

\newpage \clearpage
\subsection{Prompts}

Throughout the framework, various prompts are employed across multiple stages to guide the system's behavior. %Due to space constraints, it is not feasible to include all of them in this appendix. 
We provide the prompts used in two of the core components of the framework: the generation of interrogation questions and the refinement of the report.
The complete set of prompts is available in our GitHub repository.

\begin{tcolorbox}[colback=gray!5!white, colframe=black!75!black, title=INTERROGATION SYSTEM PROMPT, fonttitle=\bfseries, sharp corners=south]
\scriptsize
You are a skilled legal interrogator conducting an in-depth interview with a legal researcher.  
Your objective is to extract \textbf{comprehensive, well-supported legal information} by formulating precise, strategic questions.  

The goal is \textbf{not simply to obtain answers}, but to gather authoritative legal evidence, reasoning, and precedents to thoroughly address the following legal question:

\textless question\textgreater  
\{userQuery\}  
\textless/question\textgreater

\textbf{Additional Context:}  
The following background information relevant to the question is provided:

\textless context\textgreater  
\{userContext\}  
\textless/context\textgreater

\textbf{Additional Instructions:}  
You must take into account the following instructions:

\textless intructions\textgreater  
\{userInstructions\}  
\textless/intructions\textgreater

---

\textbf{Critically Consider the Existing Report Before Asking New Questions:}  \\
You have been provided with a \textbf{report summarizing the interrogation so far}. This report serves as a \textbf{synthesis of key legal arguments, findings, acknowledged knowledge gaps, and preliminary reasoning} extracted from the conversation.  

Before forming your next question, \textbf{carefully analyze this report}, which includes:  \\
- The \textbf{preliminary reasoning and draft interpretation}---a tentative legal direction that has emerged, but is still subject to revision.  \\
- \textbf{Explicitly acknowledged knowledge gaps}---areas where the legal researcher did not provide sufficient clarity, evidence, or citations.  \\
- \textbf{Remaining uncertainties and conflicting viewpoints}, including legal areas where additional investigation is required.  \\
- \textbf{Follow-up questions that have already been identified} to refine the legal analysis further.  \\

You must use this information \textbf{strategically} to craft your next question.  

---

\textbf{Your Role:}  \\
- You have \textbf{\{remaining\_questions\} questions remaining}, so each question must be maximally informative.  \\
- Your goal is to \textbf{clarify uncertainties, challenge assumptions, and press for concrete legal sources} to fill the knowledge gaps. \\  
- Your questions should probe deeper into weak or vague responses, pressing for \textbf{specific legal precedents, case law, statutory references, and counterarguments}.  \\
- Avoid redundancy---do not ask questions that have already been answered in the report. Instead, \textbf{build upon previous insights} and push the conversation forward.  \\

---

\textbf{...}\\
---

\textbf{Completion:}  
Once you are fully satisfied that you have gathered all necessary legal insights, you may conclude the interrogation by stating:  
\textit{"Thank you, I am now in a position to answer the question with confidence."}  

You will be given:  \\
- The report summarizing the previous exchange with the legal researcher.  \\
- The list of previous questions asked so far.  \\

Use this information to ensure your next question is targeted, strategic, and maximally informative.
\end{tcolorbox}

\begin{tcolorbox}[colback=gray!5!white, colframe=black!75!black, title=INTERROGATION USER PROMPT, fonttitle=\bfseries, sharp corners=south]
\scriptsize
The following report summarizes the previous exchange between you and the legal researcher.  

\textless report\textgreater  
\{report\}  
\textless/report\textgreater  

This report contains: \\
- \textbf{A preliminary interpretation or draft answer}, which is subject to revision.  \\
- \textbf{Explicitly acknowledged gaps in legal reasoning}---areas that require further clarification.  \\
- \textbf{Conflicting viewpoints or legal uncertainties} that need to be resolved.  \\
- \textbf{Follow-up questions that have been identified} to improve the legal analysis.  \\

The following questions have been asked so far: \\
\textless questions\textgreater  
\{questions\}  
\textless/questions\textgreater  \\

You must carefully analyze the above report before crafting your next question.  

Your next question should:  \\
- \textbf{Push the conversation forward}---do not repeat questions that have already been asked.  \\
- \textbf{Target unresolved knowledge gaps} and press for \textbf{specific legal references}.  \\
- \textbf{Challenge weak or unsupported reasoning}---seek case law, statutes, or counterarguments.  \\
- \textbf{Refine or reassess the preliminary interpretation}, if needed.  \\
- \textbf{Help move toward a stronger, well-supported legal answer}. \\ 

Now, continue your interrogation.
\end{tcolorbox}

\begin{tcolorbox}[colback=gray!5!white, colframe=black!75!black, title=REPORT REFINEMENT SYSTEM PROMPT, fonttitle=\bfseries\small, sharp corners=south]
\scriptsize
You are a legal technical writer tasked with \textbf{refining} a structured, professional legal report based on new information from an interrogation-style conversation between a legal interrogator and a legal researcher.

---

\textbf{Your Objective:} You will be given a legal question and an \textbf{existing draft report}. Your goal is to \textbf{analyze the updated conversation} and integrate the new insights, arguments, and legal interpretations into the existing report—always ensuring that the refinements directly contribute to answering the legal question—while maintaining a \textbf{structured, authoritative, and professional} legal analysis. DO NOT just append the new information at the end. Rewrite the report so it reads as one clear, complete, and updated version.

The final/refined report must be written as if it is the only version that exists. DO NOT acknowledge the existence of the previous report and any conversation.

Your role is \textbf{not to provide a final answer or definitive conclusion}, but to further develop the \textbf{key insights, arguments, and reasoning gaps} necessary to reach a legally sound conclusion. The refined report may challenge or revise the preliminary direction taken earlier.

---

\textbf{Guidelines for Writing the Report:}

\textbf{1. Analyze the Updated Conversation:} \\
- Carefully review the \textbf{existing legal report} and the \textbf{new conversation transcript}. \\
- Identify \textbf{new legal arguments, precedents, counterarguments, or reasoning} that emerge and \textbf{critically evaluate} whether they change or reinforce the preliminary findings. \\
- Challenge any previous interpretations if needed—\textbf{do not assume the original direction is correct}. \\
- Identify \textbf{knowledge gaps or missing legal evidence} that still prevent a definitive answer.

\textbf{2. Refine the Legal Report (Markdown Formatting):} \\
- \textbf{Preserve the original report structure but enhance it where needed:}
  
  \textbf{\#\# Title:} Keep or modify the title if the updated information suggests a more precise framing.
  
  \textbf{\#\#\# Summary:} \\
  Keep or modify the summary if the updated information suggests a more precise introduction to the topic.
  
  \textbf{\#\#\# Legal Reasoning \& Analysis:} \\
  - Expand the reasoning section with \textbf{new legal arguments or counterarguments} introduced. \\
  - Clearly \textbf{indicate changes or clarifications} while ensuring logical consistency. \\
  - Ensure that all conclusions remain legally sound and properly substantiated.
  
  \textbf{\#\#\# Preliminary Answer \& Direction for Further Research:} \\
  - Instead of refining toward a definitive answer, provide an \textbf{updated draft interpretation} or \textbf{alternative possible directions} based on new findings. \\
  - If previous reasoning is now in doubt, \textbf{state why and explore alternative legal views}. \\
  - Clarify \textbf{what would be required} to reach a more confident answer.
  
  \textbf{\#\#\# Gaps \& Next Questions:} \\
  - Explicitly state what \textbf{additional legal information, precedents, or sources} are needed to refine the analysis. \\
  - List \textbf{follow-up questions} that could help clarify uncertainties.
  
  \textbf{\#\#\# Sources:} \\
  - List all cited legal sources using numbered references \textbf{[1], [2]}, etc. \\
  - If URLs or case references exist, include them in this section. \\
  - Incorporate new references, direct quotes, and citations from the conversation where relevant. \\
  - Ensure each reference includes metadata to help locate the original text (e.g., clause number, page number, section name, etc.).

\textbf{3. Writing Style \& Formatting:} \\
- Use \textbf{formal legal writing}—precise, objective, and authoritative. \\
- Be \textbf{concise yet comprehensive} (approximately \textbf{500 words max}). \\
- Ensure \textbf{clarity and logical flow}—no redundant or unclear statements. \\
- \textbf{Do not reference the interrogator or researcher}—present findings as a \textbf{standalone report}. \\
- \textbf{Whenever possible, include direct quotes from the original context} in your references to justify your claim. Enclose these quotes in quotation marks ("") to highlight the exact supporting spans. \\
- \textbf{For each reference, specify how to locate the relevant information} in the original text (like clause number, page number, section name, etc.) \\
\textbf{4. Handling Insufficient Data:} \\
- If the conversation still lacks sufficient legal clarity or citations, \textbf{explicitly acknowledge these gaps}. \\
% \textbf{Final Review Checklist:} \\
% - The report follows the outlined structure. \\
% - Legal reasoning is \textbf{coherent, logical, and well-supported}. \\
% - All sources are correctly cited and listed. \\
% - There are \textbf{no references to the interrogator/researcher's role}. \\
% - The final version presents a \textbf{coherent and fully refined legal report} that addresses the original question. \\
% - Prioritize the most important and relevant information from both the existing report and the new conversation—keeping only the content that meaningfully impacts the answer to the legal question. \\
% - Any \textbf{new legal insights, contradictions, or alternative directions} are integrated. DO NOT just append the new information at the end. Rewrite the report so it reads as one clear, complete, and updated version. \\
% - \textbf{The preliminary answer is revisited and questioned if necessary}—it does not assume the original interpretation is correct. \\
% - Information gaps and next steps are clearly stated. \\
% - Do not mention what you changed, don't mention 'old' or 'new' information, just present the final refined report. The final/refined report must be written as if it is the only version that exists. DO NOT acknowledge the existence of the previous report and any conversation.
| ... | \\
Now, analyze the new conversation and \textbf{refine the existing legal report} accordingly.
\end{tcolorbox}

\begin{tcolorbox}[colback=gray!5!white, colframe=black!75!black, title=REPORT REFINEMENT USER PROMPT, fonttitle=\bfseries\small, sharp corners=south]
\scriptsize
Refine the following \textbf{legal report} based on the newly provided conversation between a \textbf{legal interrogator} and a \textbf{legal researcher}. \\
Prioritize the most important and relevant information from both the existing report and the new conversation—keeping only the content that meaningfully impacts the answer to the legal question.

---

\textbf{Legal Question:} \\
\textless question\textgreater \\
\{userQuery\} \\
\textless/question\textgreater

\textbf{Additional Context:} \\
The following background information relevant to the question is provided: \\
\textless context\textgreater \\
\{userContext\} \\
\textless/context\textgreater

\textbf{Updated Legal Conversation Transcript:} \\
\textless conversation\textgreater \\
\{conversation\} \\
\textless/conversation\textgreater

\textbf{Existing Legal Report:} \\
\textless legal\_report\textgreater \\
\{existingReport\} \\
\textless/legal\_report\textgreater

---

\textbf{Refinement Guidelines:} \\
- Carefully \textbf{incorporate relevant new legal arguments, precedents, and reasoning} from the conversation. DO NOT just append the new information at the end. Rewrite the report so it reads as one clear, complete, and updated version. \\
- \textbf{Critically evaluate} the existing legal report against the new conversation transcript. \\
- Do \textbf{not assume the existing direction is correct}—if the new insights challenge prior reasoning, revise accordingly. \\
- Identify \textbf{knowledge gaps and missing evidence} that prevent a definitive answer. \\
- \textbf{Explicitly highlight any contradictions or multiple possible legal interpretations}. \\
- List \textbf{follow-up questions} that need to be answered to reach a more well-founded conclusion. \\
- \textbf{Cite new references} where applicable and preserve the report's structured format. \\
- \textbf{ALWAYS ensure that every refinement you make directly enhances the accuracy and clarity of the answer to the legal question.} \\
- The final/refined report must be written as if it is the only version that exists. DO NOT acknowledge the existence of the previous report and any conversation.

Now, refine the legal report based on the new information.
\end{tcolorbox}

\newpage
\section{Human Evaluation}
\label{app:human}

We deployed a survey on \textit{Prolific}, an online research platform widely used in academic studies. Prolific ensures participant anonymity and does not allow researchers to interfere with or influence responses. Participants were asked to compare answers from PAKTON and ChatGPT for the same legal question, based on specific evaluation criteria.

We intentionally did not restrict participation to legal professionals. Our goal was to understand how well PAKTON’s responses are received by the general public, especially in comparison to a widely used baseline like ChatGPT. The only eligibility requirements were that participants must be fluent in English and have completed at least compulsory education (e.g., high school level), to ensure basic reading comprehension and critical reasoning skills.

Moreover, we opted for maintaining labeled samples regarding which one stems from PAKTON and which from ChatGPT; this way, human evaluators were fully aware of the source of each response, aiming to simulate a real-world usage scenario: in practice, users do know whether they are interacting with ChatGPT or PAKTON. This transparency was crucial to ensure that any preference expressed for PAKTON over ChatGPT reflected a conscious, informed choice, especially given the fact that ChatGPT is the most widely used publicly available LLM interface, particularly in the context of contract review. In doing so, we can be confident that any favoring of PAKTON was not due to anonymity, but instead due to users deliberately valuing PAKTON’s output over that of an industry-standard tool and signaling that they would be willing to use PAKTON over ChatGPT for contract analysis if it were publicly available.

Our human evaluation was carried out by anonymous volunteers who were fairly compensated for their time. All participants were fully informed about the purpose of the study and followed a structured annotation protocol. No personal data was collected or stored at any stage of the evaluation. The instructions given to human annotators are presented in Table \ref{tab:evaluation_criteria}. \begin{table*}[h!]
\centering
\small
\begin{tabular}{>{\raggedright\arraybackslash}p{4cm}|>{\raggedright\arraybackslash}p{11cm}}
\toprule
\textbf{Criterion} & \textbf{Instructions} \\
\midrule

\textbf{Explainability and Reasoning} & Evaluate whether the report clearly and transparently explains not only the final conclusion, but also the reasoning process and supporting evidence in a step-by-step, understandable manner. The explanation should guide the reader through the logic in a way that supports comprehension, avoiding unexplained jumps in logic. \\

\midrule

\textbf{Justification with Evidence} & Determine whether the statements and claims are explicitly justified with relevant, specific, and clearly cited evidence (e.g., direct quotations, clause references). The justification should be traceable, allowing the reader to locate the original source material. \\

\midrule

\textbf{Contextual and Legal Understanding} & Assess whether the report demonstrates a deep and accurate understanding of the document, its legal terminology, and the broader context. Consider whether it correctly interprets clauses and captures implied assumptions or legal concerns behind the question. \\

\midrule

\textbf{Handling Ambiguity} & Determine whether the report identifies and handles ambiguities in the source material appropriately, such as by presenting multiple interpretations or justifying a chosen one clearly. \\

\midrule

\textbf{Acknowledgment of Knowledge Gaps} & Evaluate whether the report explicitly acknowledges when available information is insufficient to support a conclusion, avoiding speculation or overconfidence. \\

\midrule

\textbf{Conciseness and Precision} & Assess whether the report communicates clearly and efficiently, avoiding unnecessary repetition or verbosity, while still covering all key points. \\

\midrule

\textbf{Coherence and Organization} & Check whether the report is logically structured, flows smoothly, and maintains clarity across sections. Transitions between ideas should be natural and helpful. \\

\midrule

\textbf{Relevance and Focus} & Evaluate whether the report stays on topic and maintains focus on answering the core question, avoiding tangents or irrelevant content. \\

\midrule

\textbf{Completeness} & Assess whether the report addresses all important aspects of the question and offers a contextually broad and holistic answer. It should not omit any major points or perspectives. \\

\hline
\end{tabular}
\caption{Instructions given to human annotators for each evaluation criterion used in the PAKTON vs. ChatGPT comparison. Similar instructions were given to the G-EVAL framework.}
\label{tab:evaluation_criteria}
\end{table*}

\begin{figure*}[h!]
    \centering

    \includegraphics[width=0.6\textwidth]{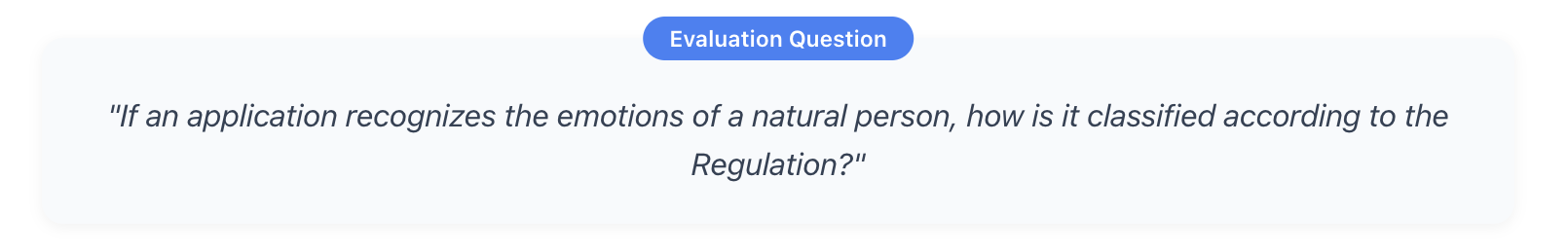}

    \includegraphics[width=1\textwidth]{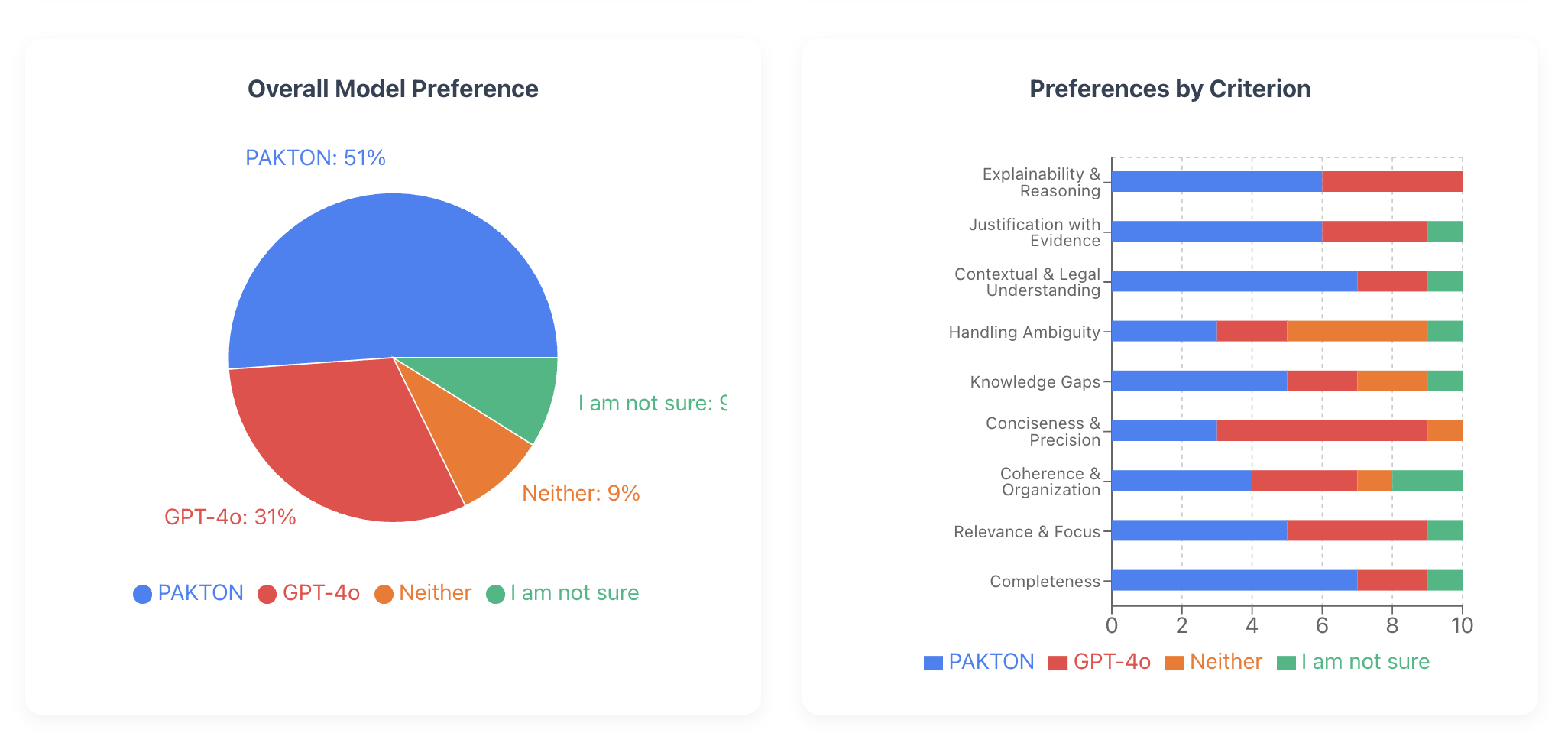}

    (a) Preference based on responses for a single question

    \vspace{1ex}

    \includegraphics[width=0.8\textwidth]{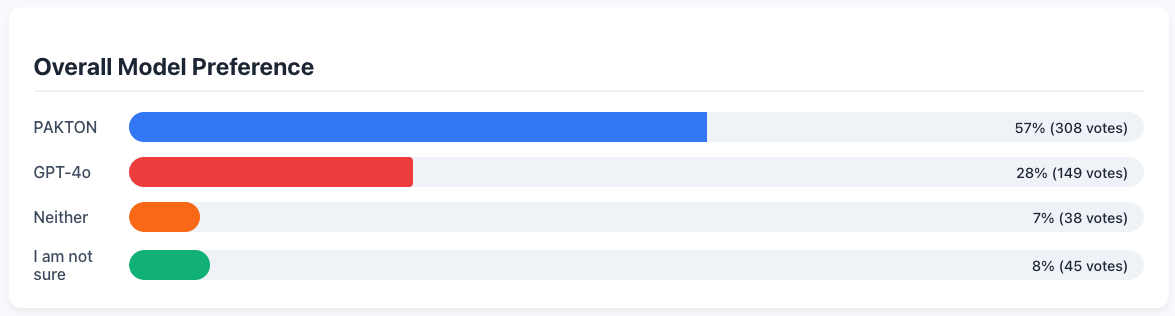}

    (b) Overall Model Preference aggregated across all criteria and all questions

    \vspace{1ex}
    
    \caption{Comparative analysis of PAKTON vs. GPT-4o based on human evaluator judgments across different criteria}
    \label{fig:human_evaluation_emotions}
\end{figure*}

Figure~\ref{fig:human_evaluation_emotions} presents a comparative assessment between PAKTON and GPT-4o, based on human judgments across multiple criteria. PAKTON consistently outperformed GPT-4o, both at the individual response level and in aggregate preferences.
% Further details about the experiments can be found here: \url{https://pakton.site/evaluation/geval}

% Further details about the survey can be found here: \url{https://pakton.site/evaluation/human-evaluation}

% \begin{figure*}[ht]
%     \centering
%     \includegraphics[width=1\textwidth]{static/PAKTON_UI_1.png}\\[1ex]
%     \includegraphics[width=1\textwidth]{static/PAKTON_UI_2.png}\\[1ex]
%     \includegraphics[width=1\textwidth]{static/PAKTON_UI_3.png}
%     \caption{The User Interface (UI) adapted for the EU AI Act question-answering task. It showcases the user interaction flow and the structure of the generated responses.}
%     % This example is accessible at \url{https://pakton.site/examples/?num=9}. You can view other examples by changing the 'num' value in the URL (1–12).}
%     \label{fig:pakton_examples}
% \end{figure*}

Participants were paid £9/hour, exceeding Prolific’s minimum compensation rate of £6/hour. While the expected duration of the survey was 5 minutes, the median completion time was closer to 15 minutes. To fairly account for the additional effort, participants who spent more time were rewarded with bonuses, calculated based on Prolific’s minimum rate.

% The outputs generated by PAKTON were presented to participants through the user interface illustrated in Figure~\ref{fig:pakton_examples}. In addition to reviewing predefined examples, participants were also given the opportunity to interact with the system in real time and evaluate its responses to their own questions. The system was met with enthusiastic and favorable feedback from participants, particularly with regard to its practical usefulness.
\begin{figure*}[h!]
    \centering
    \includegraphics[width=1\textwidth]{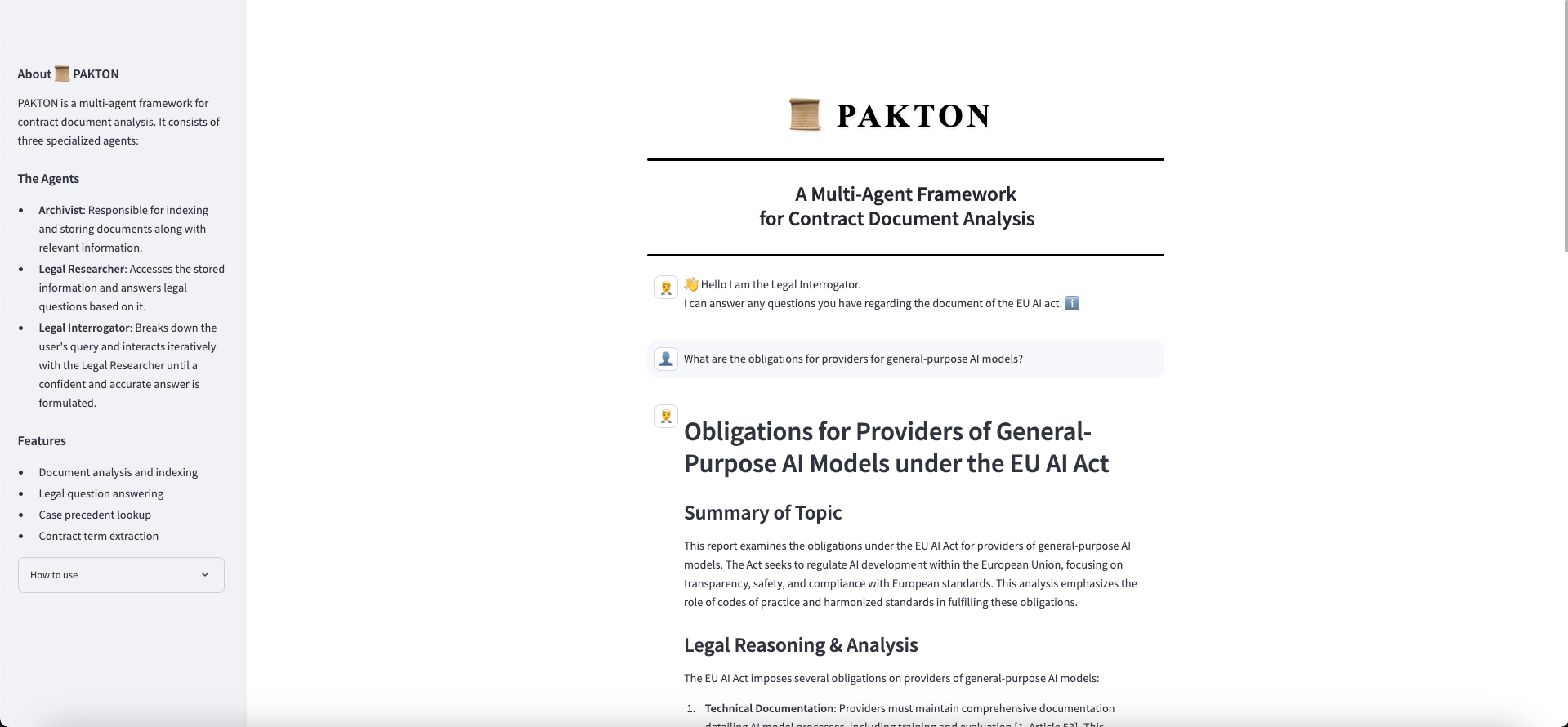}\\[1ex]
    \includegraphics[width=1\textwidth]{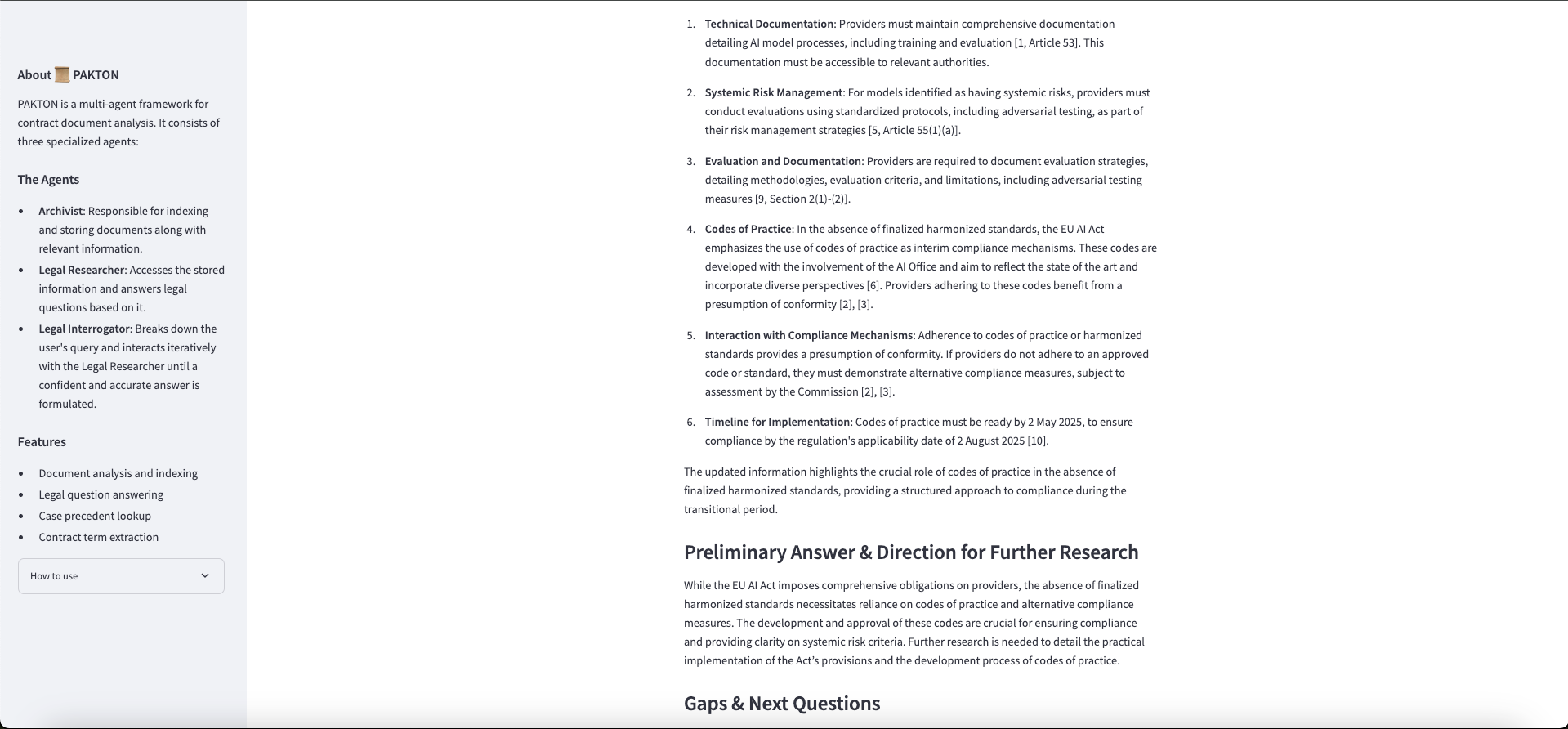}\\[1ex]
    \includegraphics[width=1\textwidth]{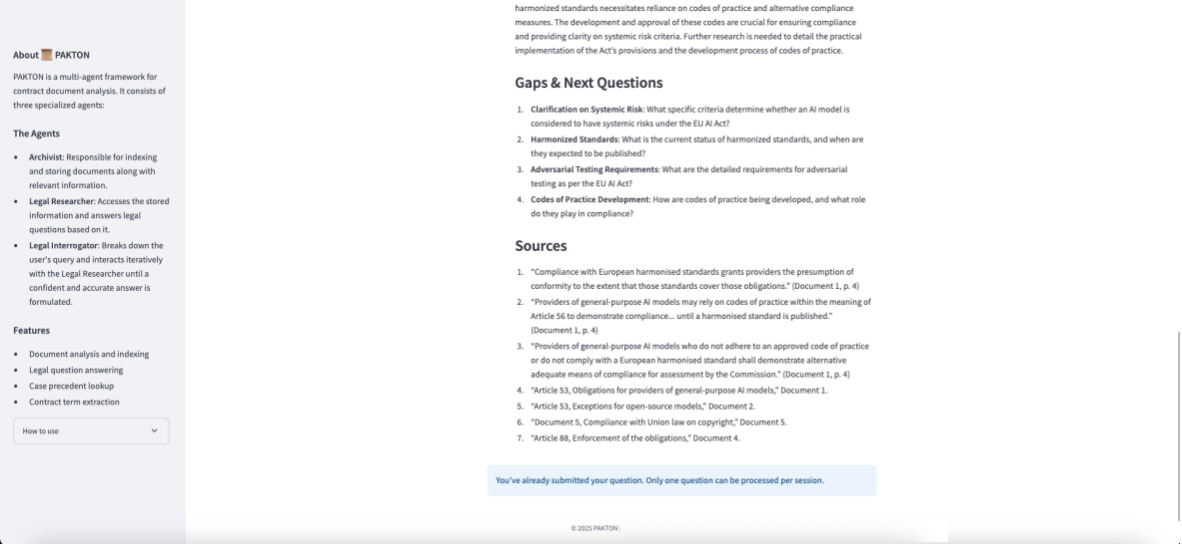}
    \caption{The user interface (UI) of PAKTON employed during the human evaluation with study participants.}
    % This example is accessible at \url{https://pakton.site/examples/?num=9}. You can view other examples by changing the 'num' value in the URL (1–12).}
    \label{fig:pakton_examples}
\end{figure*}
Evaluators assessed PAKTON’s outputs via the interface shown in Figure~\ref{fig:pakton_examples}, reviewing both predefined examples and interacting with the system in real time using their own queries. Feedback was overwhelmingly positive, highlighting the tool’s practical utility.

% A comparative analysis of PAKTON and GPT-4o, based on human evaluator judgments across multiple criteria, is presented in Figure~\ref{fig:human_evaluation_emotions}. The figure displays both the response-level preferences for individual questions and the overall aggregated preferences across all criteria. PAKTON is consistently preferred over GPT-4o in both the detailed and aggregated evaluations.

\newpage
\section{G-EVAL Experiments}
\label{app:geval}
We set the temperature to 0 to ensure deterministic responses, facilitating reproducibility.

As part of our evaluation design, we ensured fair model comparison settings. For ``GPT'', we implemented a (RAG) pipeline. For PAKTON, we limited tool usage to strictly in-document retrieval (disabling access to external tools like web search), and capped the number of interrogation turns at five. GPT-4o served as the underlying model for both systems to eliminate base model performance discrepancies and isolate differences due to architecture and orchestration.
\begin{figure}[t!]
    \centering
    \includegraphics[width=\columnwidth, trim=20 0 0 0, clip]{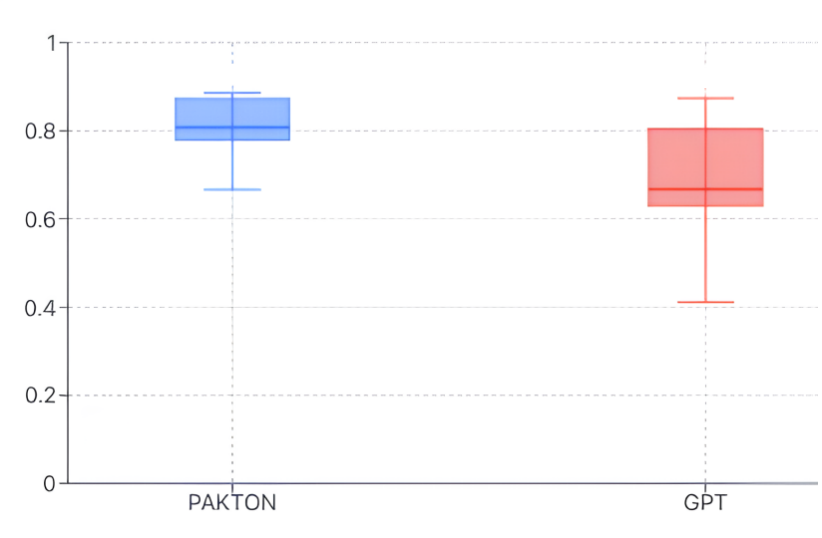}
    \caption{Comparative analysis of PAKTON vs.\ GPT-4o based on G-EVAL across different criteria. Distribution of the aggregated scores across all criteria.}
    \label{fig:geval_evaluation_box_plot}
\end{figure}

The aggregated scores for all evaluated outputs of PAKTON and GPT-4o on the ContractNLI dataset are presented in Table~\ref{tab:pakton_gpt_comparison}.
The distribution of the aggregated scores across all evaluation criteria, as computed by the G-EVAL framework, is illustrated in Figure~\ref{fig:geval_evaluation_box_plot}.
Table~\ref{tab:understanding_penalty} provides a concrete example demonstrating how PAKTON's transparency in acknowledging knowledge gaps leads to penalized scores in the "Contextual and Legal Understanding" criterion.

\subsection{Statistical Agreement between LLM and Human Judgments}
\label{app:statistical_agreement}
The alignment between the LLM-based evaluation and human judgments is visually evident in Figure ~\ref{fig:combined_comparison}, where the results of both methods appear nearly superimposed. While such visual inspection already suggests consistency between the two evaluation approaches, a more rigorous statistical analysis was performed to strengthen and intervalidate this agreement.

\paragraph{Transformation process.}
To enable a direct comparison with human evaluation, the absolute G-EVAL scores were transformed into categorical votes. Specifically, for each data sample and evaluation criterion, if the difference between PAKTON and GPT-4o was less than $1\%$, the outcome was considered a tie and placed in the \textit{Neither} category (mirroring the human ``Neither'' option). Otherwise, the vote was assigned to the system with the higher score. This procedure yielded vote distributions across the three categories (\textit{PAKTON}, \textit{GPT-4o}, \textit{Neither}), which were normalized to percentages. Human evaluation data were processed analogously, excluding ``I am not sure'' responses.

\paragraph{Distributional similarity and variance consistency.}
Agreement was assessed on these percentage distributions. Across criteria, the average cosine similarity between LLM- and human-derived distributions was $0.88$, with an average MAE of $12.6\%$. Crucially, sensitivity analysis excluding the single outlier criterion (\textit{Contextual and Legal Understanding})---which, as noted in Section~4.3, was misinterpreted by the G-EVAL framework---further strengthens the agreement: average cosine similarity rises to $0.9164$, while average MAE falls to $10.88\%$ (RMSE $14.06\%$). Variance-comparison tests further indicate no significant differences in dispersion between methods (all F-tests $p \geq 0.05$ across categories), and two-sample distribution tests (Kolmogorov--Smirnov and Mann--Whitney U) similarly do not detect significant differences between the LLM and human distributions (all $p \geq 0.05$), including in the outlier-excluded analysis (e.g., K--S $p = 0.28$--$0.98$ across categories).
\begin{table}[t!]
\footnotesize
\centering
\begin{tabular}{p{3.7cm}|cc}
\toprule
\textbf{Evaluation Criterion} & \textbf{PAKTON} & \textbf{GPT} \\
\midrule
Justification with Evidence        & 0.6887 & 0.6503 \\
Contextual and Legal Understanding & 0.8054 & 0.8472 \\
Handling Ambiguity                & 0.6925 & 0.4248 \\
Acknowledgment of Knowledge Gaps  & 0.9124 & 0.6905 \\
Conciseness and Precision          & 0.8691 & 0.8323 \\
Coherence and Organization         & 0.9024 & 0.5118 \\
Relevance and Focus                & 0.9155 & 0.9030 \\
Completeness                      & 0.8216 & 0.6866 \\
Explainability and Reasoning       & 0.8350 & 0.8127 \\
\bottomrule
\end{tabular}
\caption{Comparison of PAKTON and GPT scores acquired across different evaluation criteria aggregated for all examined outputs.}
\label{tab:pakton_gpt_comparison}
\end{table}

\begin{table}[h!]
\centering \small
\begin{tabular}{|p{1.5cm}|p{5cm}|}
\hline
\textbf{Criterion} & Contextual and Legal Understanding \\ \hline
\textbf{Score} & 0.87549 \\ \hline
\textbf{Reason} & The report accurately interprets legal terminology and context, correctly identifies that employees are not third parties, and aligns with the contract's confidentiality obligations. It fully understands the user's query, addressing legal issues and assumptions, \textbf{but suggests further research for comprehensive understanding.} \\ \hline
\end{tabular}
\caption{Example illustrating score penalization of "Contextual and Legal Understanding" due to further research section of the report.}
% You can see the outputs of the G-EVAL here: \url{https://pakton.site/evaluation/geval}}
\label{tab:understanding_penalty}
\end{table}

\paragraph{Where agreement is strongest.}
With the outlier removed, several criteria exhibit near-identity between methods:
\begin{itemize}
    \item \textbf{Completeness:} cosine $0.9992$, MAE $1.77\%$
    \item \textbf{Explainability and Reasoning:} cosine $0.9915$, MAE $4.07\%$
    \item \textbf{Relevance and Focus:} cosine $0.9901$, MAE $3.86\%$
    \item \textbf{Justification with Evidence:} cosine $0.9751$, MAE $6.69\%$
    \item \textbf{Acknowledgment of Knowledge Gaps:} cosine $0.9769$, MAE $7.24\%$
\end{itemize}

Even for more demanding dimensions such as \textit{Conciseness and Precision} and \textit{Coherence and Organization}, the methods preserve broadly similar distributional shapes, despite larger absolute gaps (e.g., MAE up to $\sim 22$--$23\%$). Category-level calibration offsets are small to moderate (e.g., mean differences of $+13.7\%$ for ``PAKTON,'' $-11.6\%$ for ``GPT,'' and $-2.1\%$ for ``Neither'' in the outlier-excluded analysis) and do not overturn these distributional agreements.

\paragraph{Takeaway.}
Beyond the visual overlap in Figure \ref{fig:combined_comparison}, the quantitative evidence, particularly after excluding a single outlier criterion, shows that the LLM-based judging and human evaluation yield highly similar vote distributions across \textit{PAKTON}, \textit{GPT-4o}, and \textit{Neither}. The combination of high cosine similarity, reduced MAE/RMSE, matched variances, and non-significant distribution tests substantiates the claim that the two methods are in strong statistical agreement, thereby mutually reinforcing the validity of our evaluation results.

% The code and experimental outputs related to the G-EVAL evaluations will be made publicly available repository upon publication.

\clearpage
% \appendix
\section{Representative Output from PAKTON and Comparative Analysis with GPT (RAG)}

The following examples illustrate the outputs generated by PAKTON and GPT (RAG) in response to the same ContractNLI input, consisting of a contract-hypothesis pair. The input is as follows:
\begin{tcolorbox}[colback=gray!5!white, colframe=black!75!black, title=CONTRACT NLI INPUT, fonttitle=\bfseries, sharp corners=south]
\small
Is the following hypothesis ENTAILMENT, CONTRADICTION, or NEUTRAL according to the content of the contract: \textless hypothesis\textgreater Receiving Party shall destroy or return some Confidential Information upon the termination of Agreement.\textless/hypothesis\textgreater? \\

The problem is categorized as a Natural Language Inference (NLI) task for contract analysis. The answer can be one of the following: \\

\textbf{ENTAILMENT:} \\
- The hypothesis is logically true based on the content of the contract. \\
- It is explicitly stated or can be directly inferred. \\

\textbf{CONTRADICTION:} \\
- The hypothesis directly or logically conflicts with the contract's content. \\

\textbf{NEUTRAL:} \\
- There is insufficient information to conclude if it is CONTRADICTION or ENTAILMENT. \\
- If the hypothesis isn't mentioned in the contract or is unrelated.
\end{tcolorbox}

% \vspace{1em}
Based on this input, GPT produced the following output:

\begin{tcolorbox}[colback=gray!5!white, colframe=black!75!black, title=GPT (RAG) OUTPUT, fonttitle=\bfseries, sharp corners=south]
\small
The contract does indeed entail the hypothesis that the Receiving Party shall destroy or return some Confidential Information upon the termination of the Agreement. \\

As outlined in section 9 of the contract, labeled \textquotedblleft Return of JEA Confidential Information\textquotedblright, it specifies that \\
\textquotedblleft the Recipient shall immediately return and redeliver to the other all tangible material embodying the JEA Confidential Information provided hereunder\ldots and all other documents or materials ('Notes')\ldots upon the earlier of: \\
I. the completion or termination of the dealings between the parties contemplated hereunder; or \\
II. the termination of this Agreement; or \\
III. at such time as JEA may so request.\textquotedblright \\

It also provides an alternative where, with the written consent of JEA, the Recipient may destroy such information, and it provides details on how this destruction should be carried out and verified. \\

These obligations regarding handling Confidential Information are specific to the event of the termination of the Agreement, directly supporting the hypothesis.
\end{tcolorbox}

While the output of PAKTON for the same input is presented as follows:

% \vspace{1em}

\begin{tcolorbox}[
  colback=gray!5!white,
  colframe=black!75!black,
  title=PAKTON OUTPUT,
  fonttitle=\bfseries,
  sharp corners=south,
  enhanced,
  breakable
]\small
\textbf{Title:} Analysis of Contractual Obligations Regarding Confidential Information Post-Termination \\

\textbf{Summary of Topic:} \\
The legal question under consideration is whether the hypothesis that "Receiving Party shall destroy or return some Confidential Information upon the termination of Agreement" is classified as ENTAILMENT, CONTRADICTION, or NEUTRAL according to the contract's content. This involves determining whether the contract explicitly or implicitly supports, contradicts, or is neutral regarding this hypothesis. \\

\textbf{Legal Reasoning \& Analysis:} \\
The contract under review contains specific clauses that delineate the obligations of the Receiving Party regarding the handling of confidential information upon the termination of the agreement. The updated analysis provides a more nuanced understanding of these obligations and introduces potential exceptions and modifications: \\
\textbf{1. Return of Confidential Information:} \\
- The contract specifies that upon request by the disclosing party, the Receiving Party is obligated to return all confidential information. \\
- Exceptions exist for analyses prepared by or for the Receiving Party, and they may retain one copy for compliance with applicable laws or record retention policies [1], [2], [3]. \\

\textbf{2. Destruction of Confidential Information:} \\
- The contract provides an option for the Receiving Party to destroy the confidential information instead of returning it, provided there is written consent from the disclosing party. \\
- The destruction must be verifiable, and computerized data must be non-recoverably erased [2] (Clause 9). \\

\textbf{3. Exceptions to Obligations:} \\
- Certain types of confidential information are exempt from the return or destruction obligations. \\
- This includes information received from a third party without breach, information known prior to disclosure, publicly known information, independently developed information, and information ordered for public release by a government agency [1], [3], [4], [5]. \\

\textbf{4. Retention for Legal Compliance:} \\
- The receiving party may retain one copy of the confidential information as needed to comply with applicable law and/or the party's record retention policies. \\
- This retained information must be kept subject to the terms of the agreement or destroyed [1], [2], [3]. \\
- The retention is allowed to enable compliance with statutory document retention policies and must be necessary for legal compliance, suggesting that any retention beyond this scope would not be permissible [4]. \\

\textbf{5. Impact of Legal Compliance and Record Retention Policies:} \\
- The contract explicitly allows retention for compliance with legal and record retention requirements, thereby modifying the general obligation to return or destroy the information upon termination. \\
- This impacts the classification of the hypothesis as ENTAILMENT, as the contract articulates exceptions to the return or destruction requirement for compliance purposes [1], [2], [3]. \\

\textbf{Preliminary Answer \& Direction for Further Research:} \\
The hypothesis that the Receiving Party shall destroy or return some Confidential Information upon termination appears to be \textbf{ENTAILMENT} with specific conditions and exceptions. \\
The contract outlines obligations for returning or destroying confidential information, with allowances for retention due to legal compliance and record retention policies. \\
The determination of ENTAILMENT assumes that the specific conditions outlined in the contract are met, such as the disclosing party's request or consent and the applicability of exceptions for certain types of information. \\
Further research should focus on verifying whether these conditions apply universally to all confidential information or are subject to specific circumstances or exceptions not covered in the provided excerpts. \\

\textbf{Gaps \& Next Questions:} \\
- Clarification on Universal Application: Is the obligation to return or destroy confidential information upon termination applicable to all types of confidential information, or are there exceptions not covered in the provided clauses? \\
\textbf{...} \\
% - Verification of Consent Procedures: What are the specific procedures and requirements for obtaining written consent for destruction, and do these affect the classification of the hypothesis? \\
% - Examination of Additional Clauses: Are there other clauses in the contract that might modify or provide exceptions to the obligations outlined in the discussed clauses? \\
- Impact of Legal Compliance: How does the retention of information for legal compliance or record retention policies affect the overall obligations regarding confidential information? \\

\textbf{Sources:} \\
1. Clause 5 - Limits on Confidential Information \\
2. Clause 4.e - Obligations of Receiving Party \\
3. Clause 4.e - Obligations of Receiving Party \\
4. Clause 5 - Limits on Confidential Information \\
5. Clause 5.e - Limits on Confidential Information \\

% Further analysis of the complete contract and additional legal texts would be necessary to address these gaps and reach a definitive legal conclusion.
\end{tcolorbox}

A comparison of the outputs generated by GPT and PAKTON in response to the same ContractNLI input reveals substantial differences in terms of depth, structure, and legal interpretability. The output from GPT is concise and directly addresses the hypothesis by selecting the label \textit{ENTAILMENT}. It supports this classification by referencing specific contractual clauses that align with the hypothesis and briefly justifies the decision. While this approach is effective for providing a quick assessment, the explanation remains relatively shallow. It does not explore potential exceptions, conditional obligations, or broader legal implications that may influence the interpretation.

On the other hand, the output from \textbf{PAKTON} exhibits a much more detailed and structured analysis. It decomposes the legal obligations regarding confidential information into separate categories: return, destruction, exceptions, and legal compliance. Each element is examined thoroughly, with references to specific clauses and relevant conditions. PAKTON also integrates legal reasoning, raises critical questions for further investigation, and acknowledges limitations or ambiguities in the contract's language. Additionally, PAKTON considers practical legal factors such as retention for statutory compliance and record-keeping policies—factors which are absent from GPT’s output. These aspects contribute to a more nuanced and legally grounded evaluation of the hypothesis.

Overall, PAKTON’s response reflects a deeper engagement with the legal content, demonstrating an interpretive capacity closer to that of a legal expert. GPT, while effective in recognizing textual entailment, remains more limited in its reasoning and scope of analysis.

\section{Computational Costs}
\label{appendix:computational_costs}

The runtime of \textsc{PAKTON} primarily reflects the computational characteristics of the LLM used within the system, rather than the framework itself. This is because, first, the invocation of the LLM constitutes the most compute-intensive component and represents the dominant runtime bottleneck, and second, \textsc{PAKTON} is explicitly designed to be model-agnostic. Consequently, we analyze computational cost primarily in terms of LLM calls, while accounting for auxiliary operations through a unified time-based cost model.

\subsection{Cost Metrics}

We employ two complementary cost metrics:

\begin{enumerate}
    \item \textbf{LLM Call Count} ($N^{\textit{LLM}}$): The number of distinct LLM inference calls, which directly determines the dominant computational cost.
    
    \item \textbf{Total Runtime} ($T_{\textit{total}}$): The end-to-end execution time, which includes LLM calls and auxiliary operations:
    $$T_{\textit{total}} = \sum_{i=1}^{N^{\textit{LLM}}} t_{\textit{LLM}}^i + T_{\textit{aux}}$$
    where $t_{\textit{LLM}}^i$ is the latency of the $i$-th LLM call and $T_{\textit{aux}}$ represents the cumulative time for auxiliary operations (embedding, reranking, OCR, etc.).
\end{enumerate}

\subsection{LLM Call Analysis}

Let us denote:
\begin{itemize}
    \item $N_{\textit{turns}}$: number of conversation turns in the user interaction
    \item $D_{\textit{int}}$: interrogation depth (number of iterative reasoning rounds)
    \item $\mathbb{I}[\cdot]$: indicator function (returns 1 if condition is true, 0 otherwise)
\end{itemize}

\paragraph{Archivist.}
The Archivist's LLM calls comprise conversation management and optional document parsing: \\
\begin{align}
N^{\textit{LLM}}_{\textit{Archivist}} &= \underbrace{(N_{\textit{turns}} + 1)}_{\textit{Conversation}} + \underbrace{\mathbb{I}[\textit{LLM\_parsing}]}_{\textit{Document\_parsing}}
\end{align}

The conversation component includes $N_{\textit{turns}}$ calls for the interactive dialogue with the user plus one call for final summarization and extraction of the query, context, and instructions. The document parsing term takes the value 1 when zero-shot LLM-based parsing is used rather than BERT-based parsing.

\paragraph{Interrogator.}
The Interrogator executes $D_{\textit{int}}$ reasoning rounds, each involving question generation, research, and report refinement:
\begin{multline}
N^{\textit{LLM}}_{\textit{Interrogator}} = D_{\textit{int}} \times \underbrace{(1}_{\textit{Question}} + \underbrace{N^{\textit{LLM}}_{\textit{Researcher}}}_{\textit{Research}} + \underbrace{1)}_{\textit{Report}}
\end{multline}

\paragraph{Researcher.}
Each Researcher invocation requires query extraction, with optional natural language response:
\begin{align}
N^{\textit{LLM}}_{\textit{Researcher}} = \underbrace{1}_{\textit{Query extraction}} + \underbrace{\mathbb{I}[\textit{NL\_response}]}_{\textit{Response generation}}
\end{align}

When natural language response is disabled, the Researcher returns raw retrieved spans without additional LLM processing.

\paragraph{Total LLM Calls.}
Combining all components:
\begin{multline}
N^{\textit{LLM}}_{\textit{Total}} = N^{\textit{LLM}}_{\textit{Archivist}} + N^{\textit{LLM}}_{\textit{Interrogator}}\\
= (N_{\textit{turns}} + 1) + \mathbb{I}[\textit{LLM\_parsing}]\\
\quad + D_{\textit{int}} \times \left(2 + 1 + \mathbb{I}[\textit{NL\_response}]\right)
\end{multline}

\subsection{Auxiliary Operations}

While LLM calls dominate the computational cost, auxiliary operations contribute to the total runtime:

\paragraph{Document Processing.}
\begin{itemize}
    \item $T_{\textit{OCR}}$: Cost of Optical Character Recognition. \\ This step relies on lightweight neural networks and the runtime is negligible compared to LLM inference.
    \item $T_{\textit{parse}}$: Cost of Hierarchical BERT parsing
    \item $T_{\textit{encode}}$: Cost of Document Encoding \\ Each resulting document chunk is transformed into an embedding vector using an embedding model. This cost is strongly influenced by the size of the embedding model and it is proportional to the number of chunks produced.
\end{itemize}

\paragraph{Retrieval Operations (per interrogation round).}
\begin{itemize}
    \item $T_{\textit{query}}$: Cost of Query Embedding
    \item $T_{\textit{search}}$: Cost of Vector Search \\ Nearest-neighbor lookup in the vector store, Complexity: $O(\log n \times d)$ where $n$ is the number of stored vectors and $d$ is the vector dimensionality
    \item $T_{\textit{rerank}}$: Cost of Reranking
\end{itemize}

\subsection{Unified Cost Model}

To compare total computational costs, we express all operations in terms of equivalent runtime:

\begin{multline}
T_{\textit{total}} = \underbrace{\sum_{i=1}^{N^{\textit{LLM}}} t_{\textit{LLM}}^i}_{\textit{LLM inference}} + \underbrace{T_{\textit{OCR}} + T_{\textit{parse}} + T_{\textit{encode}}}_{\textit{Document processing}}\\
\quad + \underbrace{D_{\textit{int}} \times (T_{\textit{query}} + T_{\textit{search}} + T_{\textit{rerank}})}_{\textit{Retrieval operations}}
\end{multline}

\subsection{Model-Specific Considerations}
\textsc{PAKTON} supports heterogeneous model configurations, enabling different LLMs to be assigned to different stages of the pipeline. For example, a larger and more capable model may be used for the final report generation, while a much smaller, cheaper model may suffice for query extraction. This flexibility allows for an effective trade-off between performance and computational cost.

\subsection{Comparison with Baseline Models}
When comparing computational costs against baseline methods (e.g., on the \textsc{ContractNLI} benchmark), we observe a key distinction:
\begin{itemize}
    \item \textbf{Baseline models:} Require a \emph{single} LLM call, but the input is much larger, as it must include the entire contract.
    \item \textbf{\textsc{PAKTON}:} Issues multiple LLM calls and uses larger system prompts, but only processes relevant spans of the contract, reducing the number of input tokens per call.
\end{itemize}
This design leads to a more efficient use of computational resources, particularly for long contracts, while also enabling deeper reasoning through iterative interrogation.

\section{Experiments on LegalBenchRAG}
\label{app:legalbenchragexperiments}

\subsection{Baselines}

In the original LegalBenchRAG paper, the authors evaluated several retrieval configurations for in-document retrieval using the LegalBench benchmark. The configurations and their corresponding experimental setups are summarized below:

\begin{itemize}
    \item \textbf{Method 1}: Naive fixed-size chunking with a window of 500 characters and no overlap; no reranker; embeddings generated using \texttt{text-embedding-3-large}. Results reported in Table~\ref{tab:legalbench_rag} as "Naive".
    
    \item \textbf{Method 2}: Recursive Character Text Splitter (RCTS) with no overlap; no reranker; embeddings generated using \texttt{text-embedding-3-large}. Results reported in Table~\ref{tab:legalbench_rag} as "RCTS".
    
    \item \textbf{Method 3}: Naive fixed-size chunking (500 characters, no overlap) with reranking using the Cohere reranker \texttt{rerank-english-v3.0}; embeddings from \texttt{text-embedding-3-large}. Results reported in Table~\ref{tab:legalbench_rag} as "Naive + Cohere".
    
    \item \textbf{Method 4}: Recursive Character Text Splitter with no overlap and reranking using \texttt{rerank-english-v3.0}; embeddings from \texttt{text-embedding-3-large}. Results reported in Table~\ref{tab:legalbench_rag} as "RCTS + Cohere".
\end{itemize}

Among these, Method 2 was selected by the authors as the preferred configuration. However, in this work we compare PAKTON against all four baseline configurations to demonstrate consistent performance improvements.

\subsection{PAKTON Evaluation Strategy}

To assess the retrieval capabilities of the \textbf{PAKTON} system on LegalBenchRAG, we evaluated the indexing and in-document retrieval functionality, which constitute the core components of the pipeline. In this setup:

\begin{itemize}
    \item The \textbf{Archivist} component is responsible for document indexing.
    \item The \textbf{Researcher} performs in-document retrieval.
    \item Interactions through the \textbf{Interrogator} were bypassed to focus exclusively on retrieval performance.
\end{itemize}

Each document from the dataset was indexed by the \textit{Archivist}, and the corresponding queries were directly submitted to the \textit{Researcher}. The retrieved spans were evaluated using the LegalBenchRAG scoring methodology.

\subsubsection*{Configuration 1}
\label{subsubsection:configuration_1}
\textbf{Archivist}:
\begin{itemize}
    \item Primary strategy: structural parsing.
    \item Fallback: Recursive Character Text Splitter with 1000-character chunks and no overlap if structural parsing failed.
    \item Embedding model:\\ \texttt{text-embedding-3-large}.
\end{itemize}

\textbf{Researcher}:
\begin{itemize}
    \item Query optimization using gpt-4o and only in-document search as tool enabled.
    \item BM25 retrieves top-100 chunks with a similarity threshold of 0.6.
    \item Dense embedding retriever returns top-100 chunks with no similarity filtering using embeddings model \texttt{text-embedding-3-large}
    \item Reciprocal Rank Fusion with equal weights for both retrievers to rerank chunks, pick top-64 chunks.
    \item Reranker: \texttt{BAAI/bge-reranker-v2-m3}, producing a top-64 reranked final list of chunks.
    \item Strip structural information of the chunk and keep only the original span.
\end{itemize}

\textbf{LLM Filtering}:
\begin{itemize}
    \item An additional post-reranking filtering stage is applied using \texttt{command-R} (Cohere), an open-source 35B parameter model specifically fine-tuned for Retrieval-Augmented Generation (RAG) applications.
    \item From the top-10 reranked chunks, the model identifies and extracts the most relevant sub-span(s) within each chunk, aiming to isolate highly precise evidence.
    \item This step is designed to enhance overall precision by focusing retrieval results on the most contextually pertinent portions of the content.
\end{itemize}

Results for this configuration are provided in Table~\ref{tab:pakton_legal_bench_rag_results}. Two variants are compared: one with and one without the LLM filtering step.
The results reported and compared in Table~\ref{tab:legalbench_rag} correspond to Configuration 1, specifically the variant without LLM filtering.

\subsubsection*{Configuration 2}

Same as Configuration 1 with the only change being the use of an alternative LLM-based reranker:
\begin{itemize}
    \item Reranker:\\ \texttt{AAI/bge-reranker-v2-minicpm-layerwise} (2.72B parameters) with a 28-layer cutoff.
\end{itemize}

Results for this configuration are provided in Table~\ref{tab:pakton_legal_bench_rag_results_conf_2}.

\subsection{Discussion}
\begin{table*}[h!]
\centering
\resizebox{\textwidth}{!}{%
\begin{tabular}{l|l|cccccccccccccc}
\toprule
& \multicolumn{1}{c|}{} & \multicolumn{7}{c}{Precision @ $k$} & \multicolumn{7}{c}{Recall @ $k$} \\
\cmidrule(lr){3-9} \cmidrule(lr){10-16}
\textbf{Dataset} & \textbf{LLM filtering} & 1 & 2 & 4 & 8 & 16 & 32 & 64 & 1 & 2 & 4 & 8 & 16 & 32 & 64 \\
\midrule
\multirow{2}{*}{\textbf{PrivacyQA}}    
  & without & 19.94 & 16.84 & 11.44 & 8.62 & 7.38 & 6.42 & 6.08 & 13.34 & 22.43 & 32.67 & 43.39 & 61.65 & 82.30 & 89.42 \\
  & with & 29.11 & 30.33 & 25.99 & 23.35  & 22.64 & 22.33 & 22.33 & 10.51 & 18.13 & 23.33 & 26.35 & 27.65 & 28.19 & 28.19 \\
\midrule
\multirow{2}{*}{\textbf{ContractNLI}} 
  & without & 33.02 & 30.34 & 17.33 & 9.98  & 5.87  & 4.68 & 4.52  & 53.14 & 67.47 & 80.06 & 89.71 & 95.50 & 99.56 & 99.82 \\
  & with & 59.59 & 51.36 & 46.32 & 45.00 & 45.00 & 44.87 & 44.87 & 38.53 & 45.25 & 51.95 & 54.94 & 58.00 & 58.69 & 58.69 \\
\midrule
\multirow{2}{*}{\textbf{MAUD}}        
  & without & 25.47 & 17.45 & 10.51 & 7.24 & 5.08 & 3.18 & 1.85 & 23.99 & 30.09 & 34.49  & 46.42 & 59.74 & 74.96 & 82.80 \\
  & with & 38.87 & 36.99 & 33.54 & 33.12 & 32.77 & 32.33 & 32.29 & 19.06 & 22.60 & 24.06 & 26.52 & 27.51 & 27.64 & 27.64 \\
\midrule
\multirow{2}{*}{\textbf{CUAD}}        
  & without & 11.02 & 8.83 & 6.81 & 4.72 & 2.78 & 2.07 & 1.62 & 16.52 & 24.76 & 33.34 & 46.67 & 59.53 & 77.08 & 86.23 \\
  & with & 29.14 & 29.53 & 29.31 & 28.86 & 28.89 & 28.79 & 28.77 & 25.31 & 30.30 & 34.33 & 37.68 & 38.24 & 38.68 & 38.68 \\
\midrule
\multirow{2}{*}{\textbf{ALL}}         
  & without & \textbf{22.34} & \textbf{18.37} & \textbf{11.52} & \textbf{7.63} & \textbf{5.26} & \textbf{4.08} & \textbf{3.52} & \textbf{26.77} & \textbf{36.32} & \textbf{45.26} & \textbf{56.66} & \textbf{69.17} & \textbf{83.50} & \textbf{89.58} \\
  & with & \textbf{39.17} & \textbf{37.03} & \textbf{33.78} & \textbf{32.58} & \textbf{32.26} & \textbf{32.08} & \textbf{32.05} & \textbf{23.37} & \textbf{29.07} & \textbf{33.42} & \textbf{36.36} & \textbf{37.84} & \textbf{38.29} & \textbf{38.29} \\
\bottomrule
\end{tabular}
}
\caption{Performance comparison on different datasets for Precision and Recall at various k values for PAKTON's \textit{Researcher} and \textit{Archivist} under Configuration 1.}
\label{tab:pakton_legal_bench_rag_results}
\end{table*}

\begin{table*}[h!]
\centering
\resizebox{\textwidth}{!}{%
\begin{tabular}{l|l|cccccccccccccc}
\toprule
& \multicolumn{1}{c|}{} & \multicolumn{7}{c}{Precision @ $k$} & \multicolumn{7}{c}{Recall @ $k$} \\
\cmidrule(lr){3-9} \cmidrule(lr){10-16}
\textbf{Dataset} & \textbf{LLM filtering} & 1 & 2 & 4 & 8 & 16 & 32 & 64 & 1 & 2 & 4 & 8 & 16 & 32 & 64 \\
\midrule
\multirow{2}{*}{\textbf{PrivacyQA}}    
  & without & 35.08 & 30.37 & 23.69 & 17.86 & 14.79 & 11.73 & 10.88 & 19.65 & 32.23 & 43.52 & 58.46 & 75.93 & 89.34 & 94.18 \\
  & with    & 33.51 & 32.98 & 28.01 & 26.16 & 25.61 & 25.43 & 25.39 & 20.36 & 31.65 & 37.62 & 42.49 & 43.94 & 46.89 & 46.89 \\
\midrule
\multirow{2}{*}{\textbf{ContractNLI}} 
  & without & 58.76 & 39.69 & 24.10 & 13.63 & 7.86 & 5.99 & 5.87 & 53.74 & 68.56 & 80.50 & 89.86 & 95.45 & 99.48 & 99.74 \\
  & with    & 63.73 & 54.12 & 49.40 & 48.01 & 47.90 & 47.81 & 47.81 & 54.10 & 61.30 & 68.17 & 71.60 & 72.90 & 73.41 & 73.41 \\
\midrule
\multirow{2}{*}{\textbf{MAUD}}        
  & without & 35.05 & 23.97 & 15.34 & 11.08 & 8.02 & 5.27 & 3.08 & 32.32 & 38.24 & 43.56 & 54.98 & 67.17 & 81.12 & 85.45 \\
  & with    & 41.58 & 36.01 & 32.22 & 31.34 & 30.72 & 30.33 & 30.29 & 36.90 & 41.18 & 44.18 & 47.77 & 48.55 & 48.81 & 48.81 \\
\midrule
\multirow{2}{*}{\textbf{CUAD}}        
  & without & 20.10 & 15.46 & 10.82 & 8.13 & 5.30 & 3.91 & 3.08 & 16.86 & 24.98 & 33.95 & 47.84 & 60.85 & 77.66 & 84.38 \\
  & with    & 35.75 & 30.93 & 29.38 & 29.11 & 28.83 & 28.82 & 28.81 & 30.99 & 37.81 & 43.14 & 47.04 & 48.85 & 49.28 & 49.28 \\
\midrule
\multirow{2}{*}{\textbf{ALL}}         
  & without & 37.24 & 27.37 & 18.49 & 12.68 & 8.99 & 6.73 & 5.73 & 30.64 & 41.00 & 50.38 & 62.79 & 74.85 & 86.90 & 90.94 \\
  & with    & 43.64 & 38.51 & 34.75 & 33.66 & 33.27 & 33.10 & 33.08 & 35.59 & 42.99 & 48.28 & 52.23 & 53.56 & 54.60 & 54.60 \\
\bottomrule
\end{tabular}
}
\caption{Performance comparison across different datasets in terms of Precision and Recall at various k values, using PAKTON’s \textit{Researcher} and \textit{Archivist} components under Configuration 1, based on span-based calculation of Precision and Recall.}
\label{tab:pakton_legal_bench_rag_results_span_based}
\end{table*}
\begin{table*}[h!]
\centering
\resizebox{\textwidth}{!}{%
\begin{tabular}{l|l|ccccccc|ccccccc}
\toprule
& \multicolumn{1}{c|}{} & \multicolumn{7}{c}{Precision @ $k$} & \multicolumn{7}{c}{Recall @ $k$} \\
\cmidrule(lr){3-9} \cmidrule(lr){10-16}
\textbf{Dataset} & \textbf{LLM filtering} & 1 & 2 & 4 & 8 & 16 & 32 & 64 & 1 & 2 & 4 & 8 & 16 & 32 & 64 \\
\midrule
\multirow{2}{*}{\textbf{PrivacyQA}}    
  & without & 18.64 & 17.66 & 14.13 & 10.39 & 7.51 & 6.34 & 6.01 & 17.00 & 26.39 & 39.99 & 62.73 & 79.24 & 94.90 & 97.35 \\
  & with    & 33.02 & 31.85 & 29.33 & 26.66 & 26.11 & 25.70 & 25.68 & 10.35 & 18.20 & 23.12 & 26.95 & 28.20 & 28.58 & 28.76 \\
\midrule
\multirow{2}{*}{\textbf{ContractNLI}} 
  & without & \textbf{42.21} & \textbf{36.86} & \textbf{19.81} & \textbf{10.58} & \textbf{6.13} & \textbf{4.63} & \textbf{4.53} & \textbf{67.67} & \textbf{83.35} & \textbf{90.87} & \textbf{95.16} & \textbf{98.19} & \textbf{99.81} & \textbf{100.00} \\
  & with    & 77.51 & 67.15 & 64.02 & 63.02 & 63.22 & 63.23 & 63.23 & 51.78 & 58.93 & 63.10 & 64.52 & 65.07 & 65.24 & 65.24 \\
\midrule
\multirow{2}{*}{\textbf{MAUD}}        
  & without & 17.77 & 11.35 & 9.41 & 8.32 & 5.84 & 3.50 & 1.92 & 17.06 & 19.94 & 30.57 & 49.94 & 68.82 & 80.10 & 86.51 \\
  & with    & 37.71 & 33.60 & 31.01 & 32.12 & 32.11 & 31.96 & 31.92 & 17.92 & 21.62 & 24.69 & 28.02 & 29.90 & 30.48 & 30.49 \\
\midrule
\multirow{2}{*}{\textbf{CUAD}}        
  & without & 2.61 & 5.10 & 5.72 & 4.18 & 2.84 & 1.84 & 1.29 & 4.38 & 16.11 & 34.09 & 53.00 & 69.13 & 80.65 & 86.20 \\
  & with    & 25.51 & 28.62 & 30.28 & 31.13 & 30.27 & 30.21 & 30.22 & 15.54 & 23.74 & 30.06 & 33.11 & 33.60 & 33.60 & 33.65 \\
\midrule
\multirow{2}{*}{\textbf{ALL}}         
  & without & 20.31 & 17.74 & 12.27 & 8.37 & 5.58 & 4.08 & 3.44 & 26.53 & 36.45 & 48.88 & 65.21 & 78.845 & 88.87 & 92.52 \\
  & with    & 43.44 & 40.30 & 38.66 & 38.23 & 37.93 & 37.78 & 37.76 & 23.90 & 30.62 & 35.24 & 38.15 & 39.19 & 39.48 & 39.54 \\
\bottomrule
\end{tabular}
}
\caption{Performance comparison across different datasets in terms of Precision and Recall at various $k$ values, using PAKTON’s \textit{Researcher} and \textit{Archivist} under Configuration 2.}
\label{tab:pakton_legal_bench_rag_results_conf_2}
\end{table*}

\paragraph{Configuration 1 vs Configuration 2:}
While Configuration~1 serves as the primary setup for PAKTON, our evaluation indicates that Configuration~2 yields superior performance in terms of retrieval accuracy. However, this improvement comes at a cost: the second reranker in Configuration~2 is a large language model with 2.72 billion parameters, which introduces a significant latency overhead compared to the more lightweight setup in Configuration~1.

Considering the trade-off between reranking accuracy and computational efficiency, Configuration~1 represents the most suitable choice for practical deployment scenarios where speed is a critical factor. Nevertheless, for applications where performance is prioritized over inference time, Configuration~2 may be preferred to achieve more competitive results.

\paragraph{No LLM Filtering vs.\ LLM Filtering:}  
In the legal domain, \textbf{high recall} is often of paramount importance, as omitting relevant spans can lead to incomplete or flawed legal reasoning. The \textit{No LLM Filtering} configuration aligns better with this paradigm minimizing the risk of excluding potentially critical information. For this reason, it is selected as the primary setup in our evaluation.

Nonetheless, there are scenarios where \textbf{precision} is more desirable—particularly when mitigating hallucination risks or when users require concise, targeted evidence rather than exhaustive retrieval. To accommodate such use cases, we evaluate an additional post-reranking stage employing the \textit{LLM Filtering} variant, which identifies and extracts the most relevant sub-spans within each of the top-10 reranked chunks, thereby filtering the retrieval output to focus on the most contextually pertinent segments.

As shown in Tables~\ref{tab:pakton_legal_bench_rag_results} and~\ref{tab:pakton_legal_bench_rag_results_conf_2}, this LLM Filtering variant significantly improves \textbf{precision}. However, this gain comes at the cost of \textbf{recall}, especially as the value of \texttt{top\_k} increases, due to the stricter content selection. Based on these findings, we conclude that LLM Filtering is particularly advantageous for low \texttt{top\_k} settings, where focused and precise evidence is preferred. In contrast, for higher \texttt{top\_k} values, the unfiltered setup is more appropriate to maintain broader recall.

\paragraph{Character-Based vs.\ Span-Based Calculation of Precision and Recall:}  
The \textsc{LegalBench-RAG} paper adopts a character-based approach for computing precision and recall, and we follow the same protocol for our primary evaluation. However, we observe that this method may penalize retrieval strategies—particularly those targeting precision like the \textit{LLM Filtering} variant—due to the fine-grained nature of the retrieved spans. Specifically, in many cases the retrieved content consists of subspans (often smaller than a sentence) that lie within the annotated answer span. Under the character-based metric, such partial matches are treated as incomplete, thereby reducing recall—even when the retrieved content is semantically relevant and informative.

To further investigate this effect, we introduce a complementary \textit{span-based} evaluation. In this setting, a retrieved span is considered a \textit{hit} if it overlaps with any ground truth span, and a \textit{miss} otherwise. This binary overlap-based metric provides clearer insight into how often irrelevant spans are retrieved or relevant spans are entirely missed. The corresponding results are reported in Table~\ref{tab:pakton_legal_bench_rag_results_span_based} and should be interpreted in comparison with the character-based results in Table~\ref{tab:pakton_legal_bench_rag_results}.

Overall, the span-based evaluation yields consistently higher values, with a particularly notable improvement in recall for the \textit{LLM Filtering} variant. This suggests that character-level metrics may disproportionately penalize methods optimized for precision, potentially underestimating their effectiveness. It is also important to consider that different datasets contain varying numbers of ground truth spans per example, which can limit achievable recall at low \texttt{top\_k} values. For reference, Table~\ref{tab:perfect_recall_only} presents the upper bounds on recall achievable under 100\% precision.
\begin{table}[h!]
\centering
\resizebox{0.98\columnwidth}{!}{%
\begin{tabular}{l|ccccccc}
\toprule
& \multicolumn{7}{c}{Recall @ $k$} \\
\cmidrule(lr){2-8}
\textbf{Dataset} & 1 & 2 & 4 & 8 & 16 & 32 & 64 \\
\midrule
PrivacyQA     & 61.91 & 86.97 & 96.42 & 99.12 & 100.00 & 100.00 & 100.00 \\
ContractNLI   & 82.16 & 98.69 & 100.00 & 100.00 & 100.00 & 100.00 & 100.00 \\
MAUD          & 68.19 & 92.35 & 98.21 & 99.86 & 100.00 & 100.00 & 100.00 \\
CUAD          & 82.14 & 94.30 & 98.44 & 100.00 & 100.00 & 100.00 & 100.00 \\
\midrule
\textbf{ALL} & \textbf{73.63} & \textbf{93.09} & \textbf{98.27} & \textbf{99.75} & \textbf{100.00} & \textbf{100.00} & \textbf{100.00} \\
\bottomrule
\end{tabular}%
}
\caption{Recall @ $k$ for perfect retrieval using ground-truth snippets. Note: Precision is 100\% for all cases.}
\label{tab:perfect_recall_only}
\end{table}

\paragraph{Variation on Character Volume per chunk for different retrieval methods:}  
Our analysis of precision and recall is conducted across varying \texttt{top\_k} values. However, it is important to note that different retrieval methods return varying volumes of text, even when the same number of chunks is retrieved. For instance, one method may retrieve significantly fewer characters on average per chunk compared to another, despite retrieving the same number of chunks. To better understand this disparity, Table~\ref{tab:retrieved_characters_with_lengths} reports the total number of characters retrieved at various \texttt{k} values, the average answer length, and the average document length for each dataset.
\begin{table*}[h!]
\small
\centering
\begin{tabular}{l|l|ccccccc|cc}
\toprule
& & \multicolumn{7}{c|}{Number of Characters @ $k$} & \multicolumn{2}{c}{Avg. Lengths} \\
\cmidrule(lr){3-9} \cmidrule(lr){10-11}
\textbf{Dataset} & \textbf{LLM Filtering} & 1 & 2 & 4 & 8 & 16 & 32 & 64 & \textbf{Answer} & \textbf{Document} \\
\midrule
\multirow{2}{*}{PrivacyQA}     
  & without & 682 & 1415 & 2985 & 5706 & 10341 & 17961 & 21080 &  &  \\
  & with    & 207 & 403 & 711 & 1088 & 1362 & 1476 & 1523 & 1200 & 25266 \\
\midrule
\multirow{2}{*}{ContractNLI}   
  & without & 789 & 1371 & 2738 & 4869 & 8126 & 10774 & 11112 &  &  \\
  & with    & 250 & 412 & 651 & 885 & 1036 & 1081 & 1081 & 438 & 10782 \\
\midrule
\multirow{2}{*}{MAUD}          
  & without & 908 & 1765 & 3542 & 7069 & 14248 & 28699 & 56523 &  &  \\
  & with    & 301 & 477 & 678 & 906 & 1027 & 1059 & 1071 & 1259 & 353718 \\
\midrule
\multirow{2}{*}{CUAD}          
  & without & 771 & 1643 & 3496 & 7002 & 13694 & 24057 & 35750 &  &  \\
  & with    & 250 & 383 & 527 & 749 & 844 & 896 & 971 & 600 & 69303 \\
\bottomrule
\end{tabular}
\caption{Average number of characters retrieved @ $k$ for each dataset for PAKTON configuration 1. Comparison with average length of ground truth (answer) and document lengths.}
\label{tab:retrieved_characters_with_lengths}
\end{table*}

This analysis allows us to assess the actual amount of information passed to the LLM across methods. Notably, the \textit{LLM Filtering} variant consistently retrieves fewer characters than its unfiltered counterpart for the same number of chunks.

Moreover, this analysis provides insight into the degree of document compression achieved during retrieval. For example, in the MAUD dataset, our \textit{span-based} Recall@64 reaches 85.45\% while retrieving, on average, 56{,}523 characters—compared to an average document length of 353{,}718 characters—indicating an approximate 84\% compression of the original document content. Similarly, in the ContractNLI dataset, the \textit{LLM Filtering} variant achieves a Recall@32 of 74.41\% using only 1{,}081 characters, which corresponds to nearly 90\% information compression.

\begin{figure*}[h!]
  \centering
  \hspace*{-0.12\textwidth} % shift first image left
  \begin{subfigure}[b]{0.68\textwidth}
    \centering
    \includegraphics[width=\textwidth]{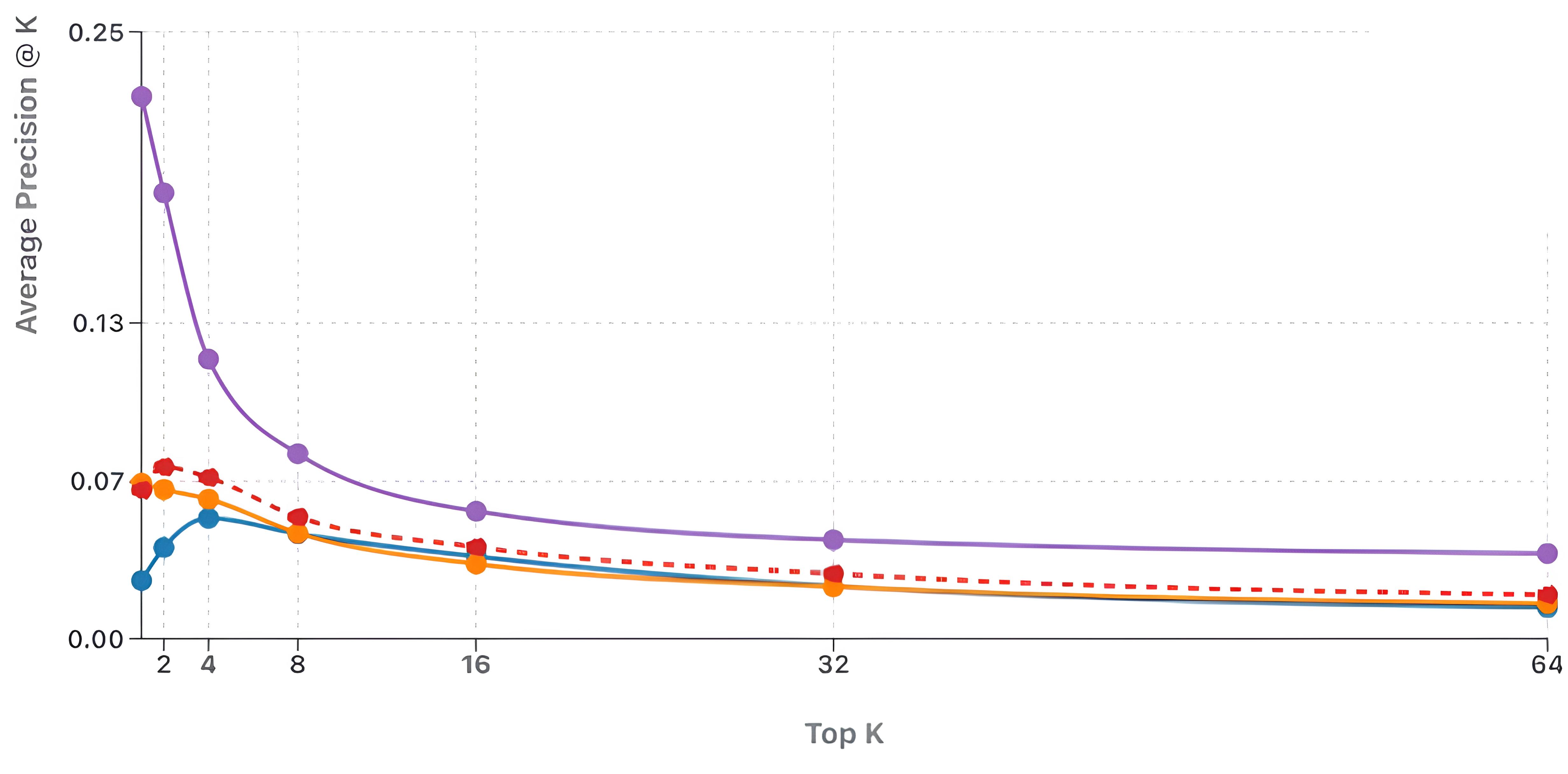}
    \caption{Precision @ k across all datasets}
    \label{fig:plot1}
  \end{subfigure}
  \hspace{0.05\textwidth}
  \begin{subfigure}[b]{0.8\textwidth}
    \centering
    \includegraphics[width=\textwidth]{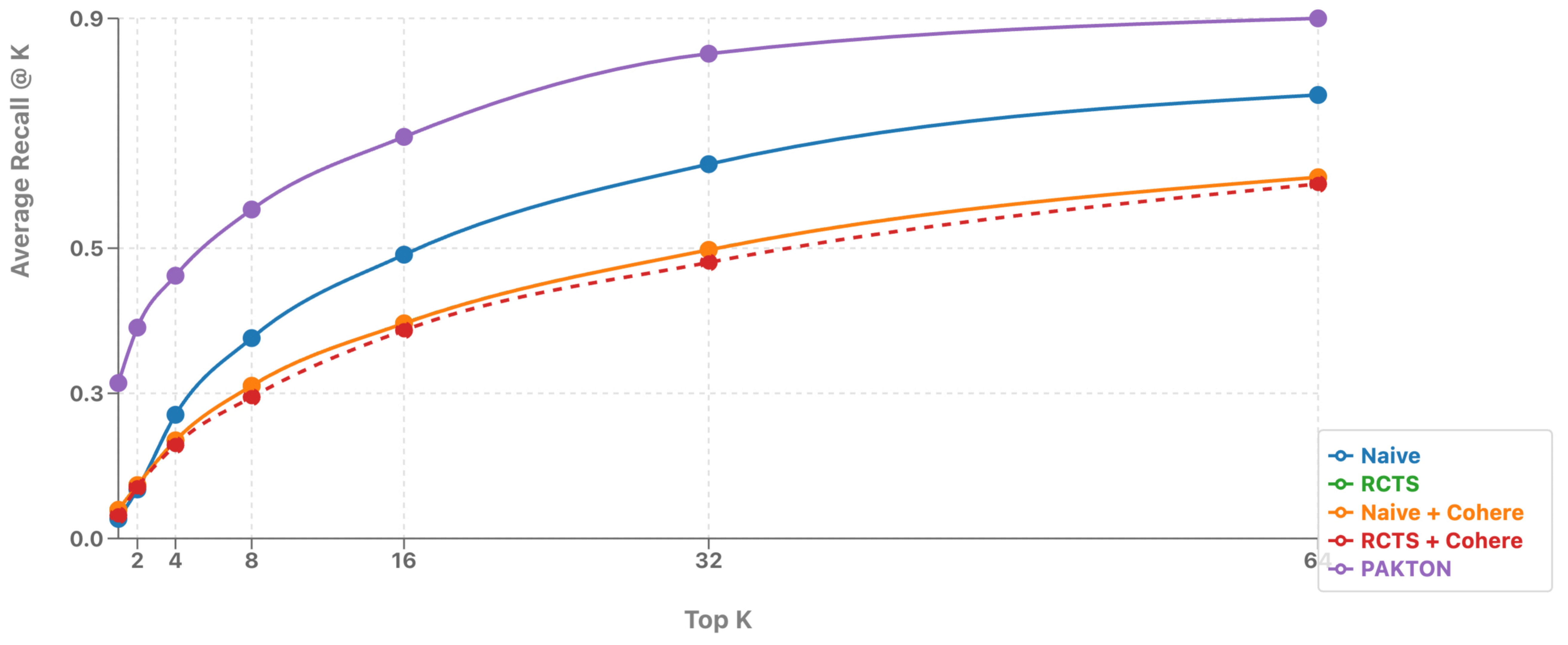}
    \caption{Recall @ k across all datasets}
    \label{fig:plot2}
  \end{subfigure}
  \hspace*{-0.12\textwidth} % shift first image left
  \begin{subfigure}[b]{0.68\textwidth}
    \centering
    \includegraphics[width=\textwidth]{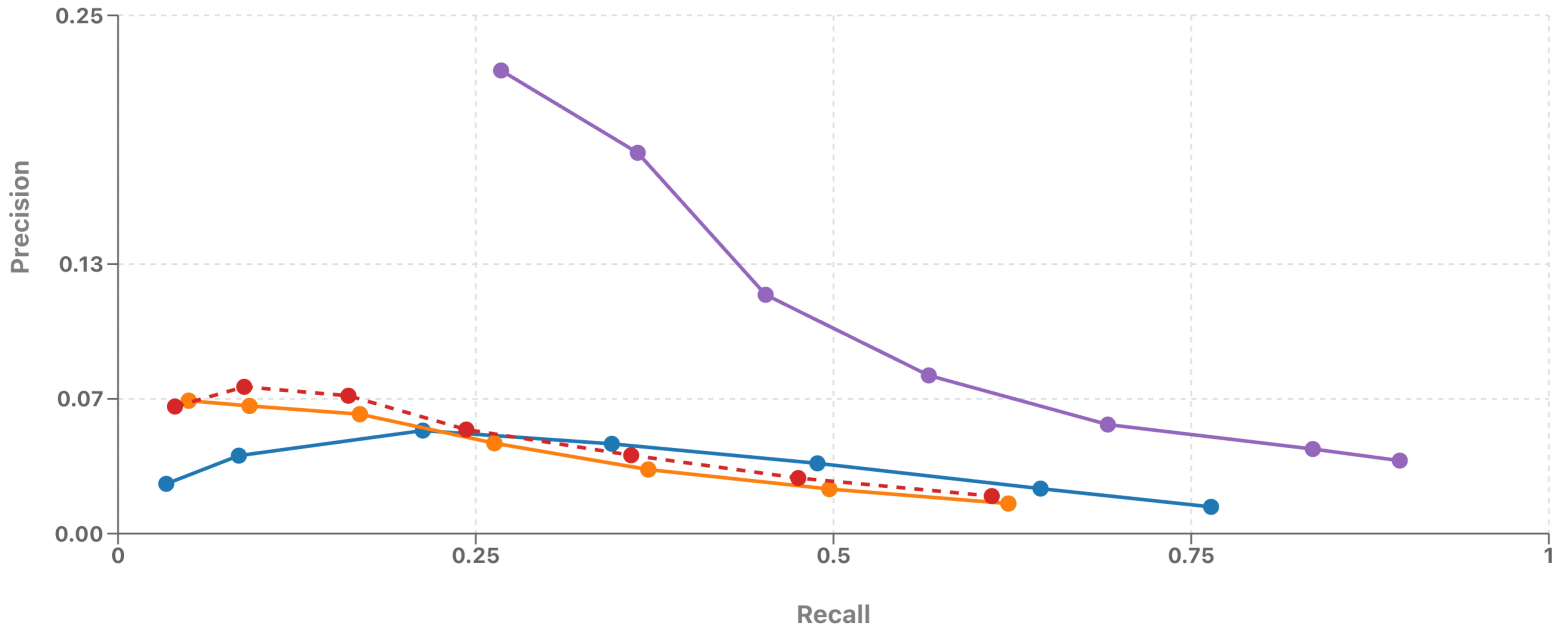}
    \caption{Precision - Recall across all datasets}
    \label{fig:plot3}
  \end{subfigure}
  \caption{Precision and Recall values for different k across all datasets for all methods for LegalBenchRAG.}
  \label{fig:retrieval_plots}
\end{figure*}
\paragraph{Plots.} Figure~\ref{fig:retrieval_plots} presents the evaluation plots for all four methods of LegalBenchRAG and PAKTON, including Precision@k, Recall@k, and Precision–Recall curves across all datasets. The results indicate that PAKTON consistently outperforms the other methods across all metrics.

\paragraph{Conclusion:} Taking into account the span-based evaluation metrics—which provide a more representative measure of retrieval quality—and the observed degree of information compression, the \textit{Researcher} module demonstrates strong performance in the task of long-document retrieval within a practical deployment context.

\section{Experiments on ContractNLI}
\label{app:contractNLIexperiments}

\subsection{Dataset Overview}

We conduct all experiments on the test split of the \texttt{ContractNLI} dataset~\citep{koreeda-manning-2021-contractnli-dataset}, which contains 2{,}091 samples. Each sample consists of a full non-disclosure agreement (NDA) as the \textit{premise}, a legal statement as the \textit{hypothesis}, and an \textit{inference label} indicating whether the hypothesis is entailed, contradicted, or neutral with respect to the contract.

\paragraph{Data Access.} We use the Hugging Face implementation of the dataset:
\url{https://huggingface.co/datasets/kiddothe2b/contract-nli}

\paragraph{Predictions.} For each experimental configuration, predictions and their associated data points are stored in structured \texttt{.json} files, which will be made publicly available upon publication.

\subsection{Dataset Subsets}

\begin{itemize}
    \item \textbf{\texttt{contractnli\_b}:} The full version of the dataset. It includes 7.19K samples in the training split, 1.04K in the validation split, and 2.09K in the test split. The full contract is used as the premise for each example.
    
    \item \textbf{\texttt{contractnli\_a}:} A filtered version of \texttt{contractnli\_b} in which only the minimal spans necessary to determine the correct label are retained as the premise, significantly reducing the input length. Experiments using this subset simulate ideal retrieval conditions and serve as an upper bound for the potential performance of a perfect RAG system.
\end{itemize}

\subsection{Prompting Strategies}

We experiment with the following prompting techniques:

\begin{itemize}
    \item \textbf{\texttt{naive zero-shot (ZS):}} No examples are given. A basic description of the label classes is provided, and the full contract is used as the premise. This serves as a solid baseline for performance.
    
    \item \textbf{\texttt{optimized zero-shot (opt. ZS):}} Uses hardcoded explanations of the classes and improved prompt structure. Still uses the full contract as the premise. Explores the effect of manual prompt engineering.
    
    \item \textbf{\texttt{naive few-shot (FS):}} Based on ~\citep{Brown2020LanguageMA}, builds on the optimized zero-shot format, but includes three random training examples (contract, hypothesis, and label) in the prompt. The full contract is used as the premise in both the examples and the current input. Demonstrates the effect of using cross-document retrieval.
    
    \item \textbf{\texttt{naive few-shot isolated spans (FS+Spans):}} Similar to naive few-shot, but in the training examples, only the relevant spans (rather than the entire contract) are provided as the premise.
    
    \item \textbf{\texttt{naive few-shot isolated spans (same hypothesis) (FS+Spans+Hyp):}} A refinement of the above method, in which few-shot examples are dynamically selected to match the hypothesis of the current sample, increasing semantic alignment.
    Shows the effect of using effective cross-document retrieval and chunking setting the upper boundary for the ideal cross-document RAG system.
    
    \item \textbf{\texttt{Chain of Thought (CoT):}} Based on ~\citep{Wei2022ChainOT}, extends the previous method by also including reasoning steps (i.e., rationales or justifications) in the answers of the few-shot examples, encouraging more explicit reasoning in the final output. Examines the impact of explicitly articulating the reasoning behind decisions.
\end{itemize}

\subsection{Experimental Setup}

\textbf{Hardware:} All local experiments were conducted on a server equipped with 4x NVIDIA A6000 GPUs, each with 48GB VRAM, using the AI daptiv infrastructure.

\textbf{Quantization:} We employed multiple quantization levels to evaluate performance under varying resource constraints. The specific techniques and configurations used will be made available in our GitHub repository upon publication.

\textbf{Scale:} In total, we ran 210 distinct experiments (each one for the whole test set of ContractNLI) across model, prompting, and input configurations.

\begin{table*}[ht]
\centering \small
% \resizebox{\textwidth}{!}{%
\begin{tabular}{c|c|c|c|c|c|c|c|c|c}
\toprule
\textbf{No} & \textbf{Model ID} & \textbf{Acc.} & \textbf{F1[W]} & \textbf{F1[E]} & \textbf{F1[C]} & \textbf{F1[N]} & \textbf{Quant.} & \textbf{Try} & \textbf{Prompting} \\
\midrule
0 & Llama-3-8B-Instr  & 0.4868 & 0.4612 & 0.5677 & 0.0327 & 0.4514 & no & 1 & ZS \\
1 & Llama-3-8B-Instr  & 0.4883 & 0.4613 & 0.5639 & 0.0252 & 0.4576 & no & 2 & ZS \\
2 & Llama-3-8B-Instr  & 0.4969  & 0.4691 & 0.5758 & 0.0166  & 0.4651 & no & 3 & ZS \\
3 & Mistral-7B-Instr   & 0.5301 & 0.4804 & 0.4551 & 0.0261 & 0.6178 & no & 1 & ZS \\
4 & Mistral-7B-Instr   & 0.5340 & 0.4843 & 0.4627 & 0.0175  & 0.6206 & no & 2 & ZS \\
5 & Mistral-7B-Instr   & 0.5328 & 0.4839 & 0.4590 & 0.0342 & 0.6201 & no & 3 & ZS \\
6 & Saul-7B-Instr          & 0.4223  & 0.2933  & 0.0570  & 0.0910 & 0.5960 & no & 1 & ZS \\
7 & Saul-7B-Instr          & 0.4218  & 0.2960 & 0.0760 & 0.0542 & 0.5907 & no & 2 & ZS \\
8 & Saul-7B-Instr          & 0.4146  & 0.2809 & 0.0437 & 0.0588 & 0.5893 & no & 3 & ZS \\
9 & Mixtral-8x7B-Instr  & 0.5608  & 0.5659 & 0.6548 & 0.4087 & 0.5088 & normal & 1 & ZS \\
10 & Mixtral-8x7B-Instr  & 0.5605  & 0.5652 & 0.6551 & 0.4099 & 0.5065 & normal & 2 & ZS \\
11 & Mixtral-8x7B-Instr  & 0.5612  & 0.5661 & 0.6545 & 0.4105 & 0.5092 & normal & 3 & ZS \\
12 & Llama-3-8B-Instr  & 0.5151  & 0.4670 & 0.6331 & 0.0000 & 0.4026 & no & 1 & opt. ZS \\
13 & Llama-3-8B-Instr  & 0.5227  & 0.4738 & 0.6372 & 0.0000 & 0.4140 & no & 2 & opt. ZS \\
14 & Llama-3-8B-Instr  & 0.5261  & 0.4785 & 0.6384 & 0.0000 & 0.4237 & no & 3 & opt. ZS \\
15 & Llama-3-8B-Instr  & 0.5332  & 0.5013 & 0.6144 & 0.0090 & 0.5000 & no & 1 & FS+Spans \\
16 & Llama-3-8B-Instr  & 0.5232  & 0.4938 & 0.5977 & 0.0177 & 0.4986 & no & 2 & FS+Spans \\
17 & Llama-3-8B-Instr  & 0.5333  & 0.5024 & 0.6089 & 0.0089 & 0.5086 & no & 3 & FS+Spans \\
18 & SaulLM-54B-Instr  & 0.7021  & 0.6806 & 0.7726 & 0.1832 & 0.7032 & lowest & 1 & ZS \\
19 & SaulLM-54B-Instr  & 0.7001  & 0.6764 & 0.7724 & 0.1594 & 0.6995 & lowest & 2 & ZS \\
20 & SaulLM-54B-Instr  & 0.7040  & 0.6807 & 0.7731 & 0.1760 & 0.7046 & lowest & 3 & ZS \\
21 & Llama-3-70B-Instr  & 0.6241 & 0.5990 & 0.7313 & 0.4972 & 0.4819 & normal & 1 & ZS \\
22 & Llama-3-70B-Instr  & 0.6236  & 0.5987 & 0.7320 & 0.4943 & 0.4813  & normal & 2 & ZS \\
23 & Llama-3-70B-Instr  & 0.6208  & 0.5953 & 0.7297 & 0.4943 & 0.4759  & normal & 3 & ZS \\
24 & DS-R1-Dist.-Llama-70B  & 0.6054  & 0.6078 & 0.6552 & 0.4837 & 0.5874 & normal & 1 & ZS \\
25 & gpt-4o  & 0.6112  & 0.6367 & 0.7497 & 0.4119 & 0.5704  & - & 1 & ZS \\
26 & gpt-4o  & 0.6112  & 0.6356 & 0.75 & 0.4151 & 0.5667  & - & 2 & ZS \\
27 & gpt-4o  & 0.6141  & 0.6376 & 0.7474 & 0.4216 & 0.5725  & - & 3 & ZS \\
28 & gpt-4o  & 0.6275  & 0.6265 & 0.5957 & 0.4416 & 0.7046  & - & 1 & opt. ZS \\
29 & gpt-4o  & 0.6165  & 0.6169 & 0.5916 & 0.4455 & 0.6856 & - & 2 & opt. ZS \\
30 & gpt-4o  & 0.6227  & 0.6240 & 0.5973 & 0.4364 & 0.6983  & - & 3 & opt. ZS \\
31 & gpt-4o  & 0.6523  & 0.6693 & 0.7315 & 0.4577 & 0.6542  & - & 1 & FS \\
32 & gpt-4o  & 0.6590  & 0.6733 & 0.7297 & 0.4624 & 0.6642  & - & 2 & FS \\
33 & gpt-4o  & 0.6805  & 0.6941 & 0.7503 & 0.5 & 0.6813 & - & 3 & FS \\
34 & gpt-4o  & 0.6413  & 0.6507 & 0.6556 & 0.4533 & 0.6936  & - & 1 & FS+Spans \\
35 & gpt-4o  & 0.6538  & 0.6627 & 0.6724 & 0.4719 & 0.6989  & - & 2 & FS+Spans \\
36 & gpt-4o  & 0.6495  & 0.6587 & 0.6712 & 0.4656 & 0.6924  & - & 3 & FS+Spans \\
37 & gpt-4o  & 0.6906  & 0.7108 & 0.7919 & 0.4866 & 0.6784  & - & 1 & ZS RAG \\
38 & Llama3-8b-Instr  & 0.4776  & 0.4278 & 0.6029 & 0.0536 & 0.3314  & - & 1 & ZS \\
39 & Mistral-7b-Instr  & 0.527  & 0.4876 & 0.4863 & 0.0248 & 0.6018  & - & 1 & ZS \\
40 & Mistral-8x7b-Instr  & 0.5872  & 0.5948 & 0.656 & 0.4427 & 0.5664  & - & 1 & ZS \\
\bottomrule
\end{tabular}
% }
\caption{Baseline performance of models across multiple evaluation runs on the ContractNLI test set. (Part 1 of 3)}
\label{tab:baseline_results_1}
\end{table*}

\begin{table*}[ht]
\centering \small
% \resizebox{\textwidth}{!}{%
\begin{tabular}{c|c|c|c|c|c|c|c|c|c}
\toprule
\textbf{No} & \textbf{Model ID} & \textbf{Acc.} & \textbf{F1[W]} & \textbf{F1[E]} & \textbf{F1[C]} & \textbf{F1[N]} & \textbf{Quant.} & \textbf{Try} & \textbf{Prompting} \\
\midrule
41 & Llama3-70b-Instr  & 0.571  & 0.536 & 0.4493 & 0.3969 & 0.6628  & - & 1 & ZS \\
42 & Claude-3-Opus  & 0.7547 & 0.7676 & 0.8339 & 0.5083 & 0.7596 & - & 1 & ZS \\
43 & Claude-3-Opus  & 0.7461 & 0.7592 & 0.8279 & 0.4992 & 0.7489 & - & 2 & ZS \\
44 & Claude-3-Opus  & 0.7475 & 0.7608 & 0.8319 & 0.4970 & 0.7489 & - & 3 & ZS \\
45 & Claude-3.5-Sonnet  & 0.7944 & 0.8002 & 0.8757 & 0.5789 & 0.7732 & - & 1 & ZS \\
46 & Claude-3.5-Sonnet  & 0.7881 & 0.7949 & 0.8749 & 0.5609 & 0.7661 & - & 2 & ZS \\
47 & Claude-3.5-Sonnet  & 0.7924 & 0.7981 & 0.8764 & 0.5770 & 0.7679 & - & 3 & ZS \\
48 & Claude-3-Opus  & 0.7819 & 0.7891 & 0.8432 & 0.5622 & 0.7863 & - & 1 & opt. ZS \\
49 & Claude-3-Opus  & 0.7857 & 0.7935 & 0.8502 & 0.5559 & 0.7906 & - & 2 & opt. ZS \\
50 & Claude-3-Opus  & 0.7819 & 0.7886 & 0.8441 & 0.5606 & 0.7847 & - & 3 & opt. ZS \\
51 & Claude-3.5-Sonnet  & 0.7901 & 0.7960 & 0.8676 & 0.5710 & 0.7740 & - & 1 & opt. ZS \\
52 & Claude-3.5-Sonnet  & 0.7905 & 0.7967 & 0.8643 & 0.5728 & 0.7787 & - & 2 & opt. ZS \\
53 & Claude-3.5-Sonnet  & 0.7915 & 0.7975 & 0.8694 & 0.5681 & 0.7762 & - & 3 & opt. ZS \\
54 & Claude-3-Opus  & 0.7580 & 0.7593 & 0.8382 & 0.5628 & 0.7226 & - & 1 & FS \\
55 & Claude-3-Opus  & 0.7676 & 0.7680 & 0.8410 & 0.5945 & 0.732 & - & 2 & FS \\
56 & Claude-3-Opus  & 0.7542 & 0.7588 & 0.8308 & 0.5406 & 0.7348 & - & 3 & FS \\
57 & Claude-3.5-Sonnet  & 0.7709 & 0.7752 & 0.8542 & 0.5654 & 0.7415 & - & 1 & FS \\
58 & Claude-3.5-Sonnet  & 0.7838 & 0.7873 & 0.8645 & 0.5781 & 0.7556 & - & 2 & FS \\
59 & Claude-3.5-Sonnet  & 0.7786 & 0.7824 & 0.8576 & 0.5671 & 0.7543 & - & 3 & FS \\
60 & Claude-3-Opus  & 0.7862 & 0.7919 & 0.8489 & 0.5673 & 0.7855 & - & 1 & FS+Spans \\
61 & Claude-3-Opus  & 0.7891 & 0.7944 & 0.8457 & 0.5964 & 0.7876 & - & 2 & FS+Spans \\
62 & Claude-3-Opus  & 0.7786 & 0.7856 & 0.8393 & 0.5575 & 0.7836 & - & 3 & FS+Spans \\
63 & Claude-3.5-Sonnet  & 0.7977 & 0.8012 & 0.8650 & 0.6032 & 0.7811 & - & 1 & FS+Spans \\
64 & Claude-3.5-Sonnet  & 0.8015 & 0.8048 & 0.8717 & 0.6049 & 0.7819 & - & 2 & FS+Spans \\
65 & Claude-3.5-Sonnet  & 0.8006 & 0.8040 & 0.8668 & 0.6056 & 0.7850 & - & 3 & FS+Spans \\
66 & Claude-3-Opus  & 0.7752 & 0.7762 & 0.8437 & 0.5900 & 0.7492 & - & 1 & FS+Spans+Hyp \\
67 & Claude-3-Opus  & 0.7834 & 0.7839 & 0.8486 & 0.6097 & 0.7571 & - & 2 & FS+Spans+Hyp \\
68 & Claude-3-Opus  & 0.7666 & 0.7682 & 0.8362 & 0.5842 & 0.7403 & - & 3 & FS+Spans+Hyp \\
69 & Claude-3.5-Sonnet  & 0.8192 & 0.8243 & 0.8819 & 0.6392 & 0.8076 & - & 1 & FS+Spans+Hyp \\
70 & Claude-3.5-Sonnet  & 0.8149 & 0.8197 & 0.8792 & 0.6346 & 0.8010 & - & 2 & FS+Spans+Hyp \\
71 & Claude-3.5-Sonnet  & 0.8197 & 0.8246 & 0.8817 & 0.6279 & 0.8112 & - & 3 & FS+Spans+Hyp \\
72 & gemma-3-27b-it  & 0.7886 & 0.7860 & 0.8316 & 0.6348 & 0.7739 & - & 1 & ZS \\
73 & qwen2.5-72b-instruct  & 0.7728 & 0.7699 & 0.8248 & 0.5776 & 0.7579 & - & 1 & ZS \\
74 & qwen2.5-72b-instruct  & 0.7810 & 0.7754 & 0.8374 & 0.5013 & 0.7757 & - & 1 & opt. ZS \\
75 & qwen2.5-72b-instruct  & 0.7351 & 0.7241 & 0.8094 & 0.4920 & 0.6892 & - & 1 & FS \\
76 & qwen2.5-72b-instruct  & 0.7484 & 0.7432 & 0.8196 & 0.4378 & 0.7357 & - & 1 & FS+Spans \\
77 & qwen2.5-72b-instruct  & 0.7604 & 0.7505 & 0.8239 & 0.6236 & 0.7028 & - & 1 & FS+Spans+Hyp \\
78 & deepseek-chat  & 0.7881 & 0.7869 & 0.8496 & 0.6087 & 0.7631 & - & 1 & ZS \\
79 & deepseek-chat  & 0.7886 & 0.7874 & 0.8487 & 0.6139 & 0.7640 & - & 2 & ZS \\
\bottomrule
\end{tabular}
% }
\caption{Baseline performance of models across multiple evaluation runs on the ContractNLI test set (Part 2 of 3).}
\label{tab:baseline_results_2}
\end{table*}

\begin{table*}[ht]
\centering \small
% \resizebox{\textwidth}{!}{%
\begin{tabular}{c|c|c|c|c|c|c|c|c|c}
\toprule
\textbf{No} & \textbf{Model ID} & \textbf{Acc.} & \textbf{F1[W]} & \textbf{F1[E]} & \textbf{F1[C]} & \textbf{F1[N]} & \textbf{Quant.} & \textbf{Try} & \textbf{Prompting} \\
\midrule
80 & deepseek-chat  & 0.7891 & 0.7883 & 0.8477 & 0.6126 & 0.7674 & - & 3 & ZS \\
81 & deepseek-chat  & 0.7872 & 0.7829 & 0.8503 & 0.6128 & 0.7521 & - & 1 & opt. ZS \\
82 & deepseek-chat  & 0.7886 & 0.7843 & 0.8535 & 0.6103 & 0.7526 & - & 2 & opt. ZS \\
83 & deepseek-chat  & 0.7891 & 0.7847 & 0.8519 & 0.6154 & 0.7539 & - & 3 & opt. ZS \\
84 & deepseek-chat  & 0.7681 & 0.7607 & 0.8346 & 0.6104 & 0.7182 & - & 1 & FS \\
85 & deepseek-chat  & 0.7743 & 0.7714 & 0.8377 & 0.5812 & 0.7465 & - & 1 & FS+Spans \\
86 & deepseek-chat  & 0.7963 & 0.7939 & 0.8612 & 0.6479 & 0.7573 & - & 1 & FS+Spans+Hyp \\
87 & deepseek-reasoner  & 0.7398 & 0.7455 & 0.8281 & 0.5229 & 0.7112 & - & 1 & ZS \\
88 & deepseek-reasoner  & 0.7394 & 0.7433 & 0.8305 & 0.5270 & 0.7024 & - & 2 & ZS \\
89 & claude-3.7-Sonnet  & 0.7704 & 0.7781 & 0.8633 & 0.5602 & 0.7398 & - & 1 & ZS \\
90 & Claude-3.7-Sonnet  & 0.7671 & 0.7733 & 0.8546 & 0.5586 & 0.7383 & - & 1 & opt. ZS \\
91 & Claude-3.7-Sonnet  & 0.7590 & 0.7602 & 0.8463 & 0.5607 & 0.7165 & - & 1 & FS \\
92 & Claude-3.7-Sonnet  & 0.7724 & 0.7766 & 0.8538 & 0.5805 & 0.7417 & - & 1 & FS+Spans \\
93 & Claude-3.7-Sonnet  & 0.8034 & 0.8068 & 0.8746 & 0.6309 & 0.7769 & - & 1 & FS+Spans+Hyp \\
94 & Llama3.1-70b-Instr  & 0.5758 & 0.5462 & 0.5014 & 0.2749 & 0.6604 & - & 1 & opt. ZS \\
95 & Llama3.1-70b-Instr  & 0.5849 & 0.5559 & 0.5208 & 0.2561 & 0.6667 & - & 2 & opt. ZS \\
96 & Llama3.1-70b-Instr  & 0.5725 & 0.5417 & 0.5046 & 0.2303 & 0.6574 & - & 3 & opt. ZS \\
97 & Llama3.1-70b-Instr  & 0.5729 & 0.5506 & 0.5421 & 0.2381 & 0.6358 & - & 1 & FS \\
98 & Llama3.3-70b-Instr  & 0.6767 & 0.6716 & 0.7366 & 0.5378 & 0.6346 & - & 1 & ZS \\
99 & Llama3.3-70b-Instr  & 0.7164 & 0.7123 & 0.7704 & 0.4745 & 0.7080 & - & 1 & opt. ZS \\
100 & Llama3.3-70b-Instr  & 0.6657 & 0.6565 & 0.7326 & 0.4431 & 0.6268 & - & 1 & FS \\
101 & Llama3.3-70b-Instr  & 0.6915 & 0.6879 & 0.7382 & 0.4244 & 0.6982 & - & 1 & FS+Spans \\
102 & Llama3.3-70b-Instr  & 0.7102 & 0.6974 & 0.7840 & 0.5294 & 0.6455 & - & 1 & FS+Spans+Hyp\\
103 & Mistral-8x7b-Instr  & 0.5423 & 0.5475 & 0.6445 & 0.4103 & 0.4770 & - & 1 & ZS \\
104 & Mistral-8x7b-Instr  & 0.6006 & 0.5931 & 0.6717 & 0.1592 & 0.6146 & - & 1 & opt. ZS \\
105 & Mistral-8x7b-Instr  & 0.6002 & 0.5804 & 0.6836 & 0.1931 & 0.5642 & - & 1 & FS \\
106 & Mistral-8x7b-Instr  & 0.6150 & 0.6017 & 0.6901 & 0.1951 & 0.6060 & - & 1 & FS+Spans \\
107 & Mistral-8x7b-Instr  & 0.5323 & 0.5302 & 0.6494 & 0.4152 & 0.4305 & - & 1 & FS+Spans+Hyp \\
108 & Mistral-7b-Instr  & 0.5364 & 0.5042 & 0.5279 & 0.0248 & 0.5951 & - & 1 & ZS \\
109 & Mistral-7b-Instr  & 0.5084 & 0.4821 & 0.6055 & 0.0000 & 0.4672 & - & 1 & opt. ZS \\
110 & Mistral-7b-Instr  & 0.5065 & 0.4702 & 0.6053 & 0.0082 & 0.4379 & - & 1 & FS \\
111 & Mistral-7b-Instr  & 0.4940 & 0.4576 & 0.6085 & 0.0076 & 0.4053 & - & 1 & FS+Spans \\
112 & Mistral-7b-Instr  & 0.4873 & 0.3660 & 0.6460 & 0.0090 & 0.1528 & - & 1 & FS+Spans+Hyp \\
\bottomrule
\end{tabular}
% }
\caption{Baseline performance of models across multiple evaluation runs on the ContractNLI test set (Part 3 of 3).}
\label{tab:baseline_results_3}
\end{table*}

\begin{table*}[ht]
\centering \small
% \resizebox{\textwidth}{!}{%
\begin{tabular}{c|c|c|c|c|c|c|c|c|c}
\toprule
\textbf{No} & \textbf{Model ID} & \textbf{Acc.} & \textbf{F1[W]} & \textbf{F1[E]} & \textbf{F1[C]} & \textbf{F1[N]} & \textbf{Quant.} & \textbf{Try} & \textbf{Prompting} \\
\midrule
0 & gpt-4o  & 0.7328  & 0.7427 & 0.7575 & 0.5936 & 0.7630  & - & 1 & ZS \\
1 & gpt-4o  & 0.7248  & 0.7367 & 0.7598 & 0.5760 & 0.7516  & - & 2 & ZS \\
2 & gpt-4o  & 0.7303  & 0.7420 & 0.7629 & 0.5799 & 0.7593  & - & 3 & ZS \\
3 & gpt-4o  & 0.7102  & 0.7059 & 0.6685 & 0.6364 & 0.7584  & - & 1 & opt. ZS \\
4 & gpt-4o  & 0.7047  & 0.7008 & 0.6653 & 0.6220 & 0.7537   & - & 2 & opt. ZS \\
5 & gpt-4o  & 0.7012  & 0.6968 & 0.6611 & 0.604 & 0.7532  & - & 3 & opt. ZS \\
6 & gpt-4o  & 0.7353  & 0.7363 & 0.7305 & 0.6333 & 0.7658  & - & 1 & FS \\
7 & gpt-4o  & 0.7283  & 0.7290 & 0.7195 & 0.6341 & 0.7602  & - & 2 & FS \\
8 & gpt-4o  & 0.7343  & 0.7347 & 0.7222 & 0.6562 & 0.7652  & - & 3 & FS \\
9 & gpt-4o  & 0.7233  & 0.7255 & 0.7183 & 0.6174 & 0.7575  & - & 1 & FS+Spans \\
10 & gpt-4o  & 0.7328  & 0.7337 & 0.7321 & 0.6324 & 0.7587  & - & 2 & FS+Spans \\
11 & gpt-4o  & 0.7378  & 0.7382 & 0.7268 & 0.6534 & 0.7689  & - & 3 & FS+Spans \\
12 & gpt-4o  & 0.7835  & 0.7842 & 0.7908 & 0.7339 & 0.7894  & - & 1 & CoT \\
13 & gpt-4o  & 0.7835  & 0.7841 & 0.7911 & 0.7403 & 0.7874   & - & 2 & CoT \\
14 & gpt-4o  & 0.7760  & 0.7766 & 0.7824 & 0.7273 & 0.7824   & - & 3 & CoT \\
15 & Claude-3-Opus  & 0.8177 & 0.8194 & 0.8398 & 0.7174 & 0.8232 & - & 1 & ZS \\
16 & Claude-3-Opus  & 0.8217 & 0.8231 & 0.8457 & 0.7221 & 0.8246  & - & 2 & ZS \\
17 & Claude-3-Opus  & 0.8177 & 0.8197 & 0.8458 & 0.7024 & 0.8215 & - & 3 & ZS \\
18 & Claude-3.5-Sonnet  & 0.8237 & 0.8253 & 0.8450 & 0.7302 & 0.8283 & - & 1 & ZS \\
19 & Claude-3.5-Sonnet  & 0.8217 & 0.8233 & 0.8420 & 0.7302 & 0.8269 & - & 2 & ZS \\
20 & Claude-3.5-Sonnet  & 0.8237 & 0.8253 & 0.8438 & 0.7343 & 0.8285 & - & 3 & ZS \\
21 & Claude-3-Opus  & 0.8282 & 0.8296 & 0.8534 & 0.7164 & 0.8327 & - & 1 & opt. ZS \\
22 & Claude-3-Opus  & 0.8242 & 0.8255 & 0.8502 & 0.7110 & 0.8281 & - & 2 & opt. ZS \\
23 & Claude-3-Opus  & 0.8262 & 0.8273 & 0.8500 & 0.72 & 0.8301 & - & 3 & opt. ZS \\
24 & Claude-3.5-Sonnet  & 0.8378 & 0.8385 & 0.8593 & 0.7420 & 0.8408 & - & 1 & opt. ZS \\
25 & Claude-3.5-Sonnet  & 0.8312 & 0.8319 & 0.8513 & 0.7420 & 0.8339 & - & 2 & opt. ZS \\
26 & Claude-3.5-Sonnet  & 0.8408 & 0.8413 & 0.8598 & 0.7607 & 0.8421 & - & 3 & opt. ZS \\
27 & Claude-3-Opus  & 0.8378 & 0.8382 & 0.8629 & 0.7489 & 0.8349 & - & 1 & FS \\
28 & Claude-3-Opus  & 0.8338 & 0.8346 & 0.8602 & 0.7311 & 0.8337 & - & 2 & FS \\
29 & Claude-3-Opus  & 0.8302 & 0.8311 & 0.8556 & 0.7338 & 0.8298  & - & 3 & FS \\
30 & Claude-3.5-Sonnet  & 0.8398 & 0.8405 & 0.8604 & 0.7511 & 0.8419 & - & 1 & FS \\
31 & Claude-3.5-Sonnet  & 0.8438 & 0.8441 & 0.8639 & 0.7699 & 0.8420 & - & 2 & FS \\
32 & Claude-3.5-Sonnet  & 0.8413 & 0.8420 & 0.8607 & 0.7527 & 0.8446 & - & 3 & FS \\
33 & Claude-3-Opus  & 0.8348 & 0.8352 & 0.8633 & 0.7373 & 0.8306  & - & 1 & FS+Spans \\
34 & Claude-3-Opus  & 0.8343 & 0.8344 & 0.8619 & 0.7380 & 0.8301  & - & 2 & FS+Spans \\
35 & Claude-3-Opus  & 0.8277 & 0.8285 & 0.8514 & 0.7342 & 0.8281 & - & 3 & FS+Spans \\
36 & Claude-3.5-Sonnet  & 0.8348 & 0.8356 & 0.8544 & 0.7420 & 0.8391 & - & 1 & FS+Spans \\
37 & Claude-3.5-Sonnet  & 0.8413 & 0.8417 & 0.8582 & 0.7696 & 0.8424 & - & 2 & FS+Spans \\
38 & Claude-3.5-Sonnet  & 0.8428 & 0.8432 & 0.8621 & 0.7625 & 0.8435 & - & 3 & FS+Spans \\
39 & Claude-3-Opus  & 0.8378 & 0.8373 & 0.8629 & 0.7385 & 0.8354 & - & 1 & FS+Spans+Hyp \\
40 & Claude-3-Opus  & 0.8368 & 0.8365 & 0.8605 & 0.7466 & 0.8340 & - & 2 & FS+Spans+Hyp \\
\bottomrule
\end{tabular}
% }
\caption{Detailed baseline performance of models across multiple evaluation runs. Evaluations are conducted on isolated spans—subsections of the contract text directly relevant to the hypothesis—. (Part 1 of 3)}
\label{tab:baseline_results_isolated_spans_1}
\end{table*}

\begin{table*}[ht]
\centering \small
% \resizebox{\textwidth}{!}{%
\begin{tabular}{l|c|c|c|c|c|c|c|c|c}
\toprule
\textbf{No} & \textbf{Model ID} & \textbf{Acc.} & \textbf{F1[W]} & \textbf{F1[E]} & \textbf{F1[C]} & \textbf{F1[N]} & \textbf{Quant.} & \textbf{Try} & \textbf{Prompting} \\
\midrule
41 & Claude-3-Opus  & 0.8358 & 0.8355 & 0.8588 & 0.7426 & 0.8344 & - & 3 & FS+Spans+Hyp \\
42 & Claude-3.5-Sonnet  & 0.8498 & 0.8499 & 0.8679 & 0.7919 & 0.8459 & - & 1 & FS+Spans+Hyp \\
43 & Claude-3.5-Sonnet  & 0.8508 & 0.8509 & 0.8644 & 0.8080 & 0.8477 & - & 2 & FS+Spans+Hyp \\
44 & Claude-3.5-Sonnet  & 0.8569 & 0.8569 & 0.8747 & 0.7891 & 0.8553 & - & 3 & FS+Spans+Hyp \\
45 & gemma-3-27b-it  & 0.8247 & 0.8238 & 0.8536 & 0.7378 & 0.8148 & - & 1 & ZS \\
46 & gemma-3-27b-it  & 0.8207 & 0.8183 & 0.8512 & 0.7037 & 0.8128 & - & 1 & opt. ZS \\
47 & gemma-3-27b-it  & 0.8127 & 0.8104 & 0.8462 & 0.6945 & 0.8025 & - & 1 & FS \\
48 & gemma-3-27b-it  & 0.8137 & 0.8116 & 0.8469 & 0.6939 & 0.8047 & - & 1 & FS+Spans \\
49 & gemma-3-27b-it  & 0.8182 & 0.8160 & 0.8472 & 0.7103 & 0.8104 & - & 1 & FS+Spans+Hyp \\
50 & qwen2.5-72b-instruct  & 0.8277 & 0.8263 & 0.8559 & 0.7345 & 0.8189 & - & 1 & ZS \\
51 & qwen2.5-72b-instruct  & 0.8217 & 0.8189 & 0.8599 & 0.6865 & 0.8098 & - & 1 & opt. ZS \\
52 & qwen2.5-72b-instruct  & 0.8212 & 0.8189 & 0.8579 & 0.6873 & 0.8116 & - & 1 & FS \\
53 & qwen2.5-72b-instruct  & 0.8227 & 0.8202 & 0.8593 & 0.6859 & 0.8134 & - & 1 & FS+Spans \\
54 & Qwen2.5-72b-Instr.  & 0.8418 & 0.8401 & 0.8651 & 0.7775 & 0.8304 & - & 1 & FS+Spans+Hyp \\
55 & deepseek-chat  & 0.8368 & 0.8365 & 0.8559 & 0.7692 & 0.8333 & - & 1 & ZS \\
56 & deepseek-chat  & 0.8373 & 0.8370 & 0.8574 & 0.7682 & 0.8330 & - & 2 & ZS \\
57 & deepseek-chat  & 0.8368 & 0.8366 & 0.8562 & 0.7675 & 0.8335  & - & 3 & ZS \\
58 & deepseek-chat  & 0.8353 & 0.8348 & 0.8564 & 0.7606 & 0.8311 & - & 1 & opt. ZS \\
59 & deepseek-chat  & 0.8348 & 0.8343 & 0.8568 & 0.7589 & 0.8299 & - & 2 & opt. ZS \\
60 & deepseek-chat  & 0.8353 & 0.8348 & 0.8573 & 0.7606 & 0.8301  & - & 3 & opt. ZS \\
61 & deepseek-chat  & 0.8307 & 0.8302 & 0.8547 & 0.7506 & 0.8250 & - & 1 & FS \\
62 & deepseek-chat  & 0.8302 & 0.8296 & 0.8593 & 0.7354 & 0.8227 & - & 2 & FS \\
63 & deepseek-chat  & 0.8368 & 0.8361 & 0.8632 & 0.7455 & 0.8309 & - & 3 & FS \\
64 & deepseek-chat  & 0.8317 & 0.8313 & 0.8571 & 0.7392 & 0.8276 & - & 1 & FS+Spans \\
65 & deepseek-chat  & 0.8368 & 0.8364 & 0.8570 & 0.7630 & 0.8335  & - & 2 & FS+Spans \\
66 & deepseek-chat  & 0.8383 & 0.8379 & 0.8637 & 0.7489 & 0.8335  & - & 3 & FS+Spans \\
67 & deepseek-chat  & 0.8473 & 0.8467 & 0.8689 & 0.7930 & 0.8375  & - & 1 & FS+Spans+Hyp \\
68 & deepseek-reasoner  & 0.7860 & 0.7903 & 0.8061 & 0.6528 & 0.8070 & - & 1 & ZS \\
69 & deepseek-reasoner  & 0.7810 & 0.7854 & 0.8011 & 0.6568 & 0.8000  & - & 2 & ZS \\
70 & deepseek-reasoner  & 0.7805 & 0.7852 & 0.8034 & 0.6458 & 0.8000  & - & 3 & ZS \\
71 & Claude-3.7-Sonnet  & 0.8061 & 0.8074 & 0.8180 & 0.7388 & 0.8131  & - & 1 & ZS \\
72 & Claude-3.7-Sonnet  & 0.8071 & 0.8084 & 0.8187 & 0.7403 & 0.8143  & - & 2 & ZS \\
73 & Claude-3.7-Sonnet  & 0.8051 & 0.8064 & 0.8162 & 0.7403 & 0.8122  & - & 3 & ZS \\
74 & Claude-3.7-Sonnet  & 0.8222 & 0.8230 & 0.8371 & 0.7609 & 0.8237 & - & 1 & opt. ZS \\
75 & Claude-3.7-Sonnet  & 0.8212 & 0.8220 & 0.8362 & 0.7593 & 0.8228  & - & 2 & opt. ZS \\
76 & Claude-3.7-Sonnet  & 0.8217 & 0.8225 & 0.8367 & 0.7609 & 0.8230  & - & 3 & opt. ZS \\
77 & Claude-3.7-Sonnet  & 0.8312 & 0.8326 & 0.8510 & 0.7439 & 0.8353  & - & 1 & FS \\
78 & Claude-3.7-Sonnet  & 0.8317 & 0.8327 & 0.8508 & 0.7556 & 0.8330  & - & 2 & FS \\
79 & Claude-3.7-Sonnet  & 0.8242 & 0.8252 & 0.8418 & 0.7531 & 0.8258  & - & 3 & FS \\
\bottomrule
\end{tabular}
% }
\caption{Detailed baseline performance of models across multiple evaluation runs. Evaluations are conducted on isolated spans—subsections of the contract text directly relevant to the hypothesis—. (Part 2 of 3)}
\label{tab:baseline_results_isolated_spans_2}
\end{table*}

\begin{table*}[ht]
\centering \small
% \resizebox{\textwidth}{!}{%
\begin{tabular}{l|c|c|c|c|c|c|c|c|c}
\toprule
\textbf{No} & \textbf{Model ID} & \textbf{Acc.} & \textbf{F1[W]} & \textbf{F1[E]} & \textbf{F1[C]} & \textbf{F1[N]} & \textbf{Quant.} & \textbf{Try} & \textbf{Prompting} \\
\midrule
80 & Claude-3.7-Sonnet  & 0.8277 & 0.8288 & 0.8429 & 0.7592 & 0.8313  & - & 1 & FS+Spans \\
81 & claude-3.7-Sonnet   & 0.8428 & 0.8431 & 0.8601 & 0.7617 & 0.8456 & - & 1 & FS+Spans+Hyp \\
82 & Llama3.1-70b-Instr  & 0.7328 & 0.7321 & 0.7321 & 0.6650 & 0.7477  & - & 1 & ZS \\
83 & Llama3.1-70b-Instr  & 0.7332 & 0.7325 & 0.7346 & 0.6505 & 0.7496  & - & 2 & ZS \\
84 & Llama3.1-70b-Instr  & 0.7411 & 0.7402 & 0.7400 & 0.6650 & 0.7578  & - & 3 & ZS \\
85 & Llama3.1-70b-Instr  & 0.7685 & 0.7655 & 0.8045 & 0.5699 & 0.7731 & - & 1 & opt. ZS \\
86 & Llama3.1-70b-Instr  & 0.7614 & 0.7585 & 0.7934 & 0.5675 & 0.7690  & - & 2 & opt. ZS \\
87 & Llama3.1-70b-Instr  & 0.7609 & 0.7588 & 0.7895 & 0.5898 & 0.7683  & - & 3 & opt. ZS \\
88 & Llama3.1-70b-Instr  & 0.7248 & 0.7226 & 0.7372 & 0.5707 & 0.7437  & - & 1 & FS \\
89 & Llama3.1-70b-Instr  & 0.7348 & 0.7325 & 0.7517 & 0.5660 & 0.7525 & - & 2 & FS \\
90 & Llama3.1-70b-Instr  & 0.7212 & 0.7196 & 0.7306 & 0.5885 & 0.7394 & - & 3 & FS \\
91 & Llama3.1-70b-Instr  & 0.7348 & 0.7321 & 0.7380 & 0.5962 & 0.7579  & - & 1 & FS+Spans \\
92 & Llama3.1-70b-Instr  & 0.7313 & 0.7288 & 0.7449 & 0.5532 & 0.7541 & - & 2 & FS+Spans \\
93 & Llama3.1-70b-Instr  & 0.7117 & 0.7087 & 0.7081 & 0.5789 & 0.7394  & - & 3 & FS+Spans \\
94 & Llama3.1-70b-Instr  & 0.8152 & 0.8132 & 0.8441 & 0.6818 & 0.8138 & - & 1 & FS+Spans+Hyp \\
95 & Llama3.1-70b-Instr  & 0.8157 & 0.8138 & 0.8437 & 0.6937 & 0.8126  & - & 2 & FS+Spans+Hyp\\
96 & Llama3.1-70b-Instr  & 0.8192 & 0.8175 & 0.8478 & 0.6948 & 0.8167 & - & 3 & FS+Spans+Hyp\\
97 & Llama3.3-70b-Instr  & 0.7845 & 0.7849 & 0.8092 & 0.7089 & 0.7790  & - & 1 & ZS \\
98 & Llama3.3-70b-Instr & 0.8117 & 0.8103 & 0.8389 & 0.7136 & 0.8051  & - & 1 & opt. ZS \\
99 & Llama3.3-70b-Instr  & 0.7941 & 0.7936 & 0.8177 & 0.6929 & 0.7935  & - & 1 & FS \\
100 & Llama3.3-70b-Instr  & 0.8021 & 0.8012 & 0.8303 & 0.6998 & 0.7966 & - & 1 & FS+Spans \\
101 & Llama3.3-70b-Instr  & 0.8403 & 0.8391 & 0.8651 & 0.7536 & 0.8337 & - & 1 & FS+Spans+Hyp \\
102 & Mistral-8x7b-Instr  & 0.7177 & 0.7192 & 0.7610 & 0.6478 & 0.6951 & - & 1 & ZS \\
103 & Mistral-8x7b-Instr  & 0.7515 & 0.7425 & 0.8117 & 0.5120 & 0.7290 & - & 1 & opt. ZS \\
104 & Mistral-8x7b-Instr  & 0.7574 & 0.7516 & 0.8065 & 0.5934 & 0.7350 & - & 1 & FS \\
105 & Mistral-8x7b-Instr  & 0.7438 & 0.7361 & 0.8008 & 0.5272 & 0.7217 & - & 1 & FS+Spans \\
106 & Mistral-8x7b-Instr  & 0.7514 & 0.7474 & 0.8106 & 0.6332 & 0.7125 & - & 1 & FS+Spans+Hyp \\
107 & Mistral-7b-Instr  & 0.6224 & 0.6052 & 0.6083 & 0.2867 & 0.6744 & - & 1 & ZS \\
108 & Mistral-7b-Instr  & 0.6217 & 0.5932 & 0.6828 & 0.0437 & 0.6339 & - & 1 & opt. ZS \\
\bottomrule
\end{tabular}
% }
\caption{Detailed baseline performance of models across multiple evaluation runs. Evaluations are conducted on isolated spans—subsections of the contract text directly relevant to the hypothesis—. (Part 3 of 3)}
\label{tab:baseline_results_isolated_spans_3}
\end{table*}

% \begin{table*}[ht]
% \centering
% % \small
% % \resizebox{\textwidth}{!}{%
% \begin{tabular}{|l|c|c|c|c|c|c|c|c|c|}
% \hline
% \textbf{No} & \textbf{Base Model ID} & \textbf{API} & \textbf{Acc.} & \textbf{F1[W]} & \textbf{F1[E]} & \textbf{F1[C]} & \textbf{F1[N]} \\
% \hline
% 0 & gemma-3-27b-it & GOOGLE & 0.8287 & 0.8283 & 0.8487 & 0.7546 & 0.8255 \\
% 1 & qwen2.5-72b-instruct & ALIBABA & 0.8192 & 0.8188 & 0.8353 & 0.7737 & 0.8132 \\
% 2 & gpt-4o & OPENAI & 0.7966 & 0.7972 & 0.7964 & 0.7592 & 0.8068 \\
% 3 & deepseek-chat & DEEPSEEK & 0.8192 & 0.8200 & 0.8315 & 0.7615 & 0.8224 \\
% 4 & Claude-3.7-Sonnet & BEDROCK & 0.8247 & 0.8254 & 0.8386 & 0.7495 & 0.8304 \\
% 5 & Llama3.3-70b-Instr & BEDROCK & 0.8217 & 0.8207 & 0.8422 & 0.7488 & 0.8165 \\
% 6 & Llama3.1-70b-Instr & BEDROCK & 0.7916 & 0.7903 & 0.8097 & 0.6846 & 0.7960 \\
% 7 & Claude-3.5-Sonnet & BEDROCK & 0.7991 & 0.8001 & 0.8157 & 0.7046 & 0.8072 \\
% 8 & Mistral-8x7b-Instr & BEDROCK & 0.7423 & 0.7429 & 0.7864 & 0.6655 & 0.7187 \\
% 9 & Mistral-7b-Instr & BEDROCK & 0.7032 & 0.6789 & 0.7782 & 0.2469 & 0.6828 \\
% \hline
% \end{tabular}
% % }
% \caption{Performance of PAKTON evaluated on the ContractNLI test set using different foundation models.}
% \label{tab:multiagent_results_reduced}
% \end{table*}

\setlength{\tabcolsep}{8pt} % horizontal padding
\renewcommand{\arraystretch}{1.25} % vertical padding
\subsection{Detailed Results}

All the experiments and the corresponding results are shown in detail in Tables ~\ref{tab:baseline_results_1}, ~\ref{tab:baseline_results_2}, ~\ref{tab:baseline_results_3}, ~\ref{tab:baseline_results_isolated_spans_1}, ~\ref{tab:baseline_results_isolated_spans_2} and ~\ref{tab:baseline_results_isolated_spans_3}.

\subsection{Execution Details of \textsc{PAKTON}}

To simulate a \textit{cross-document retrieval} setting, we indexed the training and validation splits of the ContractNLI dataset, preserving each example alongside its corresponding ground-truth label. Each contract chunk was embedded and stored in the most appropriate index, where grouping was determined by the combination of the hypothesis and its associated label. This approach emulates the core behavior of the \textit{Archivist} module, which supports organizing textual segments into logically distinct indices---such as by contract type or clause category.

For example, all instances associated with the hypothesis ``The Receiving Party shall not disclose the fact that the Agreement was agreed or negotiated'' and labeled as \textit{Neutral} were stored within a single index, while examples labeled as \textit{Entailment} or \textit{Contradiction} were assigned to their respective indices.

Each set of indices corresponding to the same hypothesis was interconnected into a composable graph using the \texttt{LlamaIndex} framework. These hypothesis-specific graphs were then integrated into a unified, higher-level composable graph. Every node---whether a graph or a leaf index---was annotated with a brief natural language description summarizing the content it encapsulated.

At inference time, this hierarchical structure was traversed recursively. At each level of the graph, a similarity comparison was conducted between the input query and the textual descriptions of child nodes to determine the most relevant subgraph to explore. This hierarchical traversal mechanism enables efficient prioritization of semantically aligned indices, thereby improving retrieval relevance.

The \textit{Researcher} module utilized this architecture as a cross-document retrieval system, returning the top-3 most relevant examples for a given query.

As for the \textit{Researcher} module, we utilized \textbf{Configuration 1} (see Section~\ref{subsubsection:configuration_1}), with \textit{No LLM filtering} and kept the top-10 reranked chunks to generate the response back to the \textit{Interrogator}.

Regarding the interrogation process, we capped the maximum number of turns to five in order to maintain efficiency and avoid excessively long interaction sequences.

% \begin{figure*}[ht]
%     \centering
%     \includegraphics[width=1\textwidth]{static/LOCAL_EXPERIMENTS.png}\\[1ex]
%     \caption{Experiments for baselines conducted in local deployment. Each row was conducted three times (except the deepseek row) and the average for each metric is calculated.}
%     \label{fig:local_experiments}
% \end{figure*}

% \begin{figure*}[ht]
%     \centering
%     \includegraphics[width=1\textwidth]{static/OPENAI_EXPERIMENTS.png}\\[1ex]
%     \caption{Experiments for baselines conducted in openai. Each row was conducted three times (except the row with the star) and the average for each metric is calculated.}
%     \label{fig:openai_experiments}
% \end{figure*}

% \begin{figure*}[ht]
%     \centering
%     \includegraphics[width=1\textwidth]{static/BEDROCK_EXPERIMENTS.png}\\[1ex]
%     \includegraphics[width=1\textwidth]{static/BEDROCK_EXPERIMENTS_2.png}\\[1ex]
%     \includegraphics[width=1\textwidth]{static/BEDROCK_EXPERIMENTS_3.png}\\[1ex]
%     \caption{Experiments for baselines conducted in bedrock. Each row was conducted three times (except the row with the star) and the average for each metric is calculated.}
%     \label{fig:bedrocks_experiments}
% \end{figure*}

\end{document}